\documentclass[10pt,journal,compsoc]{IEEEtran}

\usepackage{epsfig}
\usepackage{graphicx}
\usepackage{amsmath}
\usepackage{amssymb}

\usepackage{bm}
\usepackage{booktabs}
\usepackage{enumerate}
\usepackage{multirow}
\usepackage{subcaption}
\usepackage[dvipsnames,table]{xcolor}

\usepackage{array}
\newcolumntype{H}{>{\setbox0=\hbox\bgroup}c<{\egroup}@{}} 

\usepackage{tikz}

\usepackage[pagebackref=false,breaklinks=true,letterpaper=true,colorlinks,bookmarks=false]{hyperref}

\ifCLASSOPTIONcompsoc
  \usepackage[nocompress]{cite}
\else
  \usepackage{cite}
\fi

\newcommand{\argmin}{\operatornamewithlimits{argmin}}

\newcommand{\stufig}[5]                                       
{
        \begin{figure}[#5]
        \begin{center}
                \includegraphics[#1]{#2}
                \caption{#3}
                \label{#4}
        \end{center}
		\vspace{-\baselineskip}
        \end{figure}
}

\newcommand{\stufigstar}[5]                                   
{
        \begin{figure*}[#5]
        \begin{center}
                \includegraphics[#1]{#2}
                \caption{#3}
                \label{#4}
        \end{center}
		\vspace{-\baselineskip}
        \end{figure*}
}

\newenvironment{stusubfig}[1]
{
        \begin{figure}[#1]
        \begin{center}
}
{
        \end{center}
        \end{figure}
}

\newenvironment{stusubfig*}[1]
{
        \begin{figure*}[#1]
        \begin{center}
}
{
        \end{center}
        \end{figure*}
}

\begin{document}

\title{Real-Time RGB-D Camera Pose Estimation in \\ Novel Scenes using a Relocalisation Cascade}
\author{Tommaso Cavallari,
	    Stuart Golodetz,
	    Nicholas A.\ Lord,
	    Julien Valentin,\\
	    Victor A.\ Prisacariu,
	    Luigi Di Stefano
	    and Philip H.\ S.\ Torr%
\IEEEcompsocitemizethanks{\IEEEcompsocthanksitem TC, SG, NL and JV assert joint first authorship.
\IEEEcompsocthanksitem TC was first with the University of Bologna, then with the University of Oxford, and is now with FiveAI Ltd.
SG and NL were with the University of Oxford, and are now with FiveAI Ltd.
JV is with Google Inc.
VP and PT are with the University of Oxford.
LDS is with the University of Bologna.\protect\\
E-mail: \{tommaso.cavallari,stuart,nick\}@five.ai.
}%
}

%

\IEEEtitleabstractindextext{%
\begin{abstract}
Camera pose estimation is an important problem in computer vision, with applications as diverse as simultaneous localisation and mapping, virtual/augmented reality and navigation. Common techniques match the current image against keyframes with known poses coming from a tracker, directly regress the pose, or establish correspondences between keypoints in the current image and points in the scene in order to estimate the pose.
In recent years, regression forests have become a popular alternative to establish such correspondences.
They achieve accurate results, but have traditionally needed to be trained offline on the target scene, preventing relocalisation in new environments.
Recently, we showed how to circumvent this limitation by adapting a pre-trained forest to a new scene on the fly.
The adapted forests achieved relocalisation performance that was on par with that of offline forests, and our approach was able to estimate the camera pose in close to real time, which made it desirable for systems that require online relocalisation.
In this paper, we present an extension of this work that achieves significantly better relocalisation performance whilst running fully in real time.
To achieve this, we make several changes to the original approach: (i) instead of simply accepting the camera pose hypothesis produced by RANSAC without question, we make it possible to score the final few hypotheses it considers using a geometric approach and select the most promising one; (ii) we chain several instantiations of our relocaliser (with different parameter settings) together in a cascade, allowing us to try faster but less accurate relocalisation first, only falling back to slower, more accurate relocalisation as necessary; and (iii) we tune the parameters of our cascade, and the individual relocalisers it contains, to achieve effective overall performance.
Taken together, these changes allow us to significantly improve upon the performance our original state-of-the-art method was able to achieve on the well-known 7-Scenes and Stanford 4 Scenes benchmarks.
As additional contributions, we present a novel way of visualising the internal behaviour of our forests, and use the insights gleaned from this to show how to entirely circumvent the need to pre-train a forest on a generic scene.
\end{abstract}

}

\maketitle

\newcommand\copyrighttext{%
	\footnotesize \textcopyright{} 2019 IEEE. Personal use of this material is permitted.
	Permission from IEEE must be obtained for all other uses, in any current or future 
	media, including reprinting/republishing this material for advertising or promotional 
	purposes, creating new collective works, for resale or redistribution to servers or 
	lists, or reuse of any copyrighted component of this work in other works. 
	DOI: \href{https://doi.org/10.1109/TPAMI.2019.2915068}{10.1109/TPAMI.2019.2915068}}
\newcommand\copyrightnotice{%
	\begin{tikzpicture}[remember picture,overlay]
	\node[anchor=south,yshift=5pt] at (current page.south) {\fbox{\parbox{\dimexpr\textwidth-\fboxsep-\fboxrule\relax}{\copyrighttext}}};
	\end{tikzpicture}%
}

\copyrightnotice

\IEEEdisplaynontitleabstractindextext

%
\IEEEpeerreviewmaketitle

\IEEEraisesectionheading{\section{Introduction}\label{sec:introduction}}

\noindent Camera pose estimation is a key computer vision problem, with
applications in simultaneous localisation and mapping (SLAM)
\cite{Newcombe2011,MurArtal2014,Kaehler2015,Golodetz2018}, virtual and augmented reality
\cite{Bae2016,Castle2008,Golodetz2015SPDEMO,Paucher2010,Rodas2015,Valentin2015SP} and
navigation \cite{Lee2016}. In SLAM, the camera pose is commonly initialised
upon starting reconstruction and then tracked from frame to frame, but
tracking can easily be lost due to e.g.\ rapid movement or textureless regions
in the scene; when this happens, it is important to be able to relocalise the
camera with respect to the scene, rather than forcing the user to restart the
reconstruction. Camera relocalisation is also crucial for
loop closure when trying to build globally consistent maps
\cite{Fioraio2015,Kaehler2016,Whelan2015RSS}.

Approaches to camera relocalisation roughly fall into two main categories: (i) those that attempt to find the pose directly, e.g.\ by matching the input image against keyframes with known poses \cite{GalvezLopez2011,Gee2012,Glocker2015}, or by directly regressing the pose \cite{Kendall2015}, and (ii) those that establish correspondences between points in camera and world space, so as to deploy e.g.\ a Perspective-n-Point (PnP) algorithm \cite{Hartley2004} (on RGB data) or the Kabsch algorithm \cite{Kabsch1976} (on RGB-D data) to generate a number of camera pose hypotheses from which a single hypothesis can be selected, e.g.\ using RANSAC \cite{Fischler1981}. Hybrid approaches that first find pose candidates directly and then refine them geometrically also exist \cite{MurArtal2015,Valentin2016,Taira2018}.

Recently, Shotton et al.\ \cite{Shotton2013} proposed the use of a regression forest to directly predict corresponding 3D points in world space for all pixels in an RGB-D image (each pixel in the image effectively denotes a 3D point in camera space). By generating predictions for all pixels, their approach avoids the explicit detection, description and matching of keypoints typically required by more traditional correspondence-based methods, making it simpler and faster. Moreover, this also provides them with a significantly larger number of correspondences with which to verify or reject camera pose hypotheses. However, one major limitation they have is a need to train a regression forest on the scene of interest \emph{offline} (in advance), which prevents on-the-fly camera relocalisation in novel scenes.

Subsequent work has significantly improved upon the relocalisation performance of \cite{Shotton2013}. Guzman-Rivera et al.\ \cite{GuzmanRivera2014} rely on multiple regression forests to generate a number of camera pose hypotheses, then cluster them and use the mean pose of the cluster whose poses minimise the reconstruction error as the result. Valentin et al.\ \cite{Valentin2015RF} replace the modes \cite{Shotton2013} used in the leaves of the forests with mixtures of anisotropic 3D Gaussians in order to better model uncertainties in the 3D point predictions.
Brachmann et al.\ \cite{Brachmann2016} deploy a stacked classification-regression forest to achieve results of a quality similar to \cite{Valentin2015RF} for RGB-D relocalisation (their approach can also be used for pure RGB localisation, albeit with lower accuracy).
Meng et al.\ \cite{Meng2016} perform RGB relocalisation by estimating an initial camera pose using a regression forest, then querying a nearest neighbour keyframe image and refining the initial pose by sparse feature matching between the camera input image and the keyframe.
Massiceti et al.\ \cite{Massiceti2017} map between regression forests and neural networks to leverage the performance benefits of neural networks for dense regression while retaining the efficiency of random forests for evaluation.
Meng et al.\ \cite{Meng2017IROS} store a priority queue of non-visited branches whilst passing a feature vector down the forest during testing, and then backtrack to see whether some of those branches might have been better than the one chosen.
Meng et al.\ \cite{Meng2017arXiv} make use of both point and line segment features to achieve more robust relocalisation in poorly textured areas and/or in the face of motion blur.
Brachmann et al.~\cite{Brachmann2017CVPR} show how to replace the RANSAC stage of the conventional pipeline with a probabilistic approach to hypothesis selection that can be differentiated, allowing end-to-end training of the full system.
Li et al.\ \cite{Li2018RSS} use a fully-convolutional encoder-decoder network to predict scene coordinates for the whole image at once, to take global context into account. This obviates \cite{Brachmann2017CVPR}'s need for patch sampling, but needs significant data augmentation to avoid overfitting.
Brachmann and Rother~\cite{Brachmann2018CVPR} significantly improve on the results of \cite{Brachmann2017CVPR}, whilst also showing how to avoid the need for a 3D model at training time (albeit at a cost in performance).
Very recently, Li et al.\ \cite{Li2018arXiv} have shown how to use an angle-based reprojection loss to remove \cite{Brachmann2018CVPR}'s need to initialise the scene coordinates with a heuristic when training without a model.
However, despite all of these advances, none of these papers remove the need to train on the scene of interest in advance.

Recently, we showed \cite{Cavallari2017} that this need for \emph{offline} training on the scene of interest can be overcome through \emph{online} adaptation to a new scene of a regression forest that has been pre-trained on a generic scene. We achieve genuine on-the-fly relocalisation similar to that which can be obtained using keyframe-based approaches (e.g.\ \cite{Glocker2015}), but with significantly higher relocalisation performance. Moreover, unlike such approaches, which can struggle to relocalise from novel poses because of their reliance on matching the input image to a database of keyframes, our approach performs well even at some distance from the training trajectory.

Our initial implementation of this approach \cite{Cavallari2017} achieved relocalisation performance that was competitive with offline-trained forests, whilst requiring no pre-training on the scene of interest and relocalising in close to real time. This made it a practical and high-quality alternative to keyframe-based methods for online relocalisation in novel scenes.
In this paper, we present an extension of \cite{Cavallari2017} that achieves significantly better relocalisation performance whilst running fully in real time.
To achieve this, we make several novel improvements to the original approach:
\begin{enumerate}
	\item Instead of simply accepting the camera pose produced by RANSAC without question, we make it possible to select the most promising of the final few hypotheses it considers using a geometric approach (see \S\ref{subsubsec::hypothesisranking}).
	\item We chain several instances of our relocaliser (with different parameters) into a cascade, letting us try fast, less accurate relocalisation first, and only fall back to slow, more accurate relocalisation if needed (see \S\ref{subsubsec::cascade}).
	\item We tune the parameters of our cascade, and the individual relocalisers it contains, to achieve the most effective overall performance (see \S\ref{sec:parametertuning}).
\end{enumerate}
These changes allow us to achieve state-of-the-art results on the well-known 7-Scenes \cite{Shotton2013} and Stanford 4 Scenes \cite{Valentin2016} benchmarks. We then make two further contributions:
\begin{enumerate}
	\item[4)] We present a novel way of visualising the internal behaviour of SCoRe forests, and use this to explain in detail why adapting a forest pre-trained on one scene to a new scene makes sense (see \S\ref{subsec:forestvisualisation}).
	\item[5)] Based on the insights gleaned from this visualisation, we show that it is feasible to avoid pre-training the forest altogether by making use of a random decision function in each branch node (see \S\ref{subsec:nopretraining}).
	Whilst this approach leads to a small loss in performance with respect to a pre-trained forest, we show that it can still achieve better performance than existing methods.
\end{enumerate}

\noindent \textbf{This paper is organised as follows}: in \S\ref{sec:relatedwork}, we survey existing camera relocalisation approaches; in \S\ref{sec:method}, we describe our approach; in \S\ref{sec:experiments}, we evaluate our approach extensively on the 7-Scenes \cite{Shotton2013} and Stanford 4 Scenes \cite{Valentin2016} benchmarks; finally, in \S\ref{sec:conclusion}, we conclude. Source code for our approach can be found online at \url{https://github.com/torrvision/spaint}.

\section{Related Work}
\label{sec:relatedwork}

Due to the importance of camera relocalisation, many approaches have been proposed to tackle it over the years \cite{Piasco2018}:

\emph{(i) Straight-to-pose} methods try to determine the pose directly from the input image. Within these, matching methods try to match the input image against keyframes stored in an image database (potentially interpolating between keyframes where necessary), and direct regression methods train a decision forest or neural network to directly predict the pose. For example, Gee and Mayol-Cuevas \cite{Gee2012} estimate the pose by matching the input image against synthetic views of the scene.
Other methods match descriptors computed from the input image against a database, e.g.\ Galvez-Lopez et al.\ \cite{GalvezLopez2011} compute a bag of binary words based on BRIEF descriptors for the current image and compare it with bags of words for keyframes in the database using an L1 score. Glocker et al.\ \cite{Glocker2015} encode frames using Randomised Ferns, which when evaluated on images yield binary codes that can be matched quickly by their Hamming distance.
In terms of direct regression, Kendall et al.~\cite{Kendall2015}'s PoseNet uses a convolutional neural network to directly regress the 6D camera pose from the current image. Their later works build on this to model the uncertainty in the result \cite{Kendall2016} and explore different loss functions to achieve better results \cite{Kendall2017}.
Melekhov et al.\ \cite{Melekhov2017} train an hourglass network, using skip connections between their encoder and decoder, to directly regress the camera pose.
Kacete et al.\ \cite{Kacete2017} directly regress the camera pose using a sparse decision forest.
Clark et al.\ \cite{Clark2017} and Walch et al.\ \cite{Walch2017} directly regress the pose using LSTMs.
Valada et al.\ \cite{Valada2018} train a multi-task network to predict both 6D global pose and the relative 6D poses between consecutive frames, and report dramatic improvements over earlier neural network-based approaches on 7-Scenes \cite{Shotton2013} and Cambridge Landmarks \cite{Kendall2015}, although their best results rely on using the estimated pose from the previous frame. Radwan et al.\ \cite{Radwan2018} add semantics to this approach.

\emph{(ii) Correspondence-based} methods (e.g.\ the regression forest approaches we mention in \S\ref{sec:introduction}) find correspondences between camera and world space points and use them to estimate the pose. A common approach is to find 2D-to-3D correspondences between keypoints in the current image and 3D scene points, so as to deploy e.g.\ a Perspective-n-Point (PnP) algorithm \cite{Hartley2004} (on RGB data) or the Kabsch algorithm \cite{Kabsch1976} (on RGB-D data) to generate a number of pose hypotheses that can be pruned to a single hypothesis using RANSAC \cite{Fischler1981}.
Williams et al.\ \cite{Williams2011} recognise/match keypoints using an ensemble of randomised lists, and exclude unreliable or ambiguous matches when generating hypotheses.
Li et al.\ \cite{Li2015} use graph matching to disambiguate visually-similar keypoints.
Sattler et al.\ \cite{Sattler2015} use a fine visual vocabulary and a visibility graph to find locally unique 2D-to-3D matches. Sets of consistent matches are then used to compute hypotheses.
Sattler et al.\ \cite{Sattler2017} find correspondences in both the 2D-to-3D and 3D-to-2D directions and apply a 6-point DLT algorithm to compute hypotheses.
Schmidt et al.\ \cite{Schmidt2017} train a fully-convolutional network using a contrastive loss to output robust descriptors that can be used to establish dense correspondences.

Some hybrid methods use both paradigms. Mur-Artal et al.\ \cite{MurArtal2015} describe a relocalisation approach that initially finds pose candidates using bag of words recognition \cite{GalvezLopez2012}, which they
incorporate into their ORB-SLAM system.
They then refine these candidates using PnP and RANSAC.
Valentin et al.\ \cite{Valentin2016} find pose candidates using a retrieval forest and a multiscale navigation graph, before refining them using continuous pose optimisation.
Taira et al.\ \cite{Taira2018} use NetVLAD \cite{Arandjelovic2016} to find the $N$ closest database images to the query image, before using dense feature matching and P3P-RANSAC to generate candidate poses. They then render a synthetic view of the scene from each candidate pose, and yield the pose whose view is most similar to the query image.

Several less traditional approaches have also been tried.
Deng et al.~\cite{Deng2016} match a 3D point cloud representing the scene to a local 3D point cloud constructed from a set of query images that can be incrementally extended by the user to achieve a successful match.
Lu et al.~\cite{Lu2015} perform 3D-to-3D localisation that reconstructs a 3D model from a short video using structure-from-motion and matches that against the scene within a multi-task point retrieval framework.
Laskar et al.\ \cite{Laskar2017} train a Siamese network to predict the relative pose between two images. At test time, they find the $N$ nearest neighbours to the query image in a database, predict their poses relative to the query image, and use these in conjunction with the known poses of the neighbours to estimate the query pose.
Balntas et al.\ \cite{Balntas2018} train a Siamese network to generate global feature descriptors using a continuous metric learning loss based on camera frustum overlap. Like \cite{Laskar2017}, they then predict the relative poses between the query image and nearest neighbours in a database, and use these to determine the query pose.
Sch{\"o}nberger et al.\ \cite{Schoenberger2018} train a variational encoder-decoder network to hallucinate complete, denoised semantic 3D subvolumes from incomplete ones. At test time, they match query subvolumes against ones in a database using the network's embedding space, and use the matches to generate pose hypotheses.

\section{Method}
\label{sec:method}

\stufigstar{width=.8\linewidth}{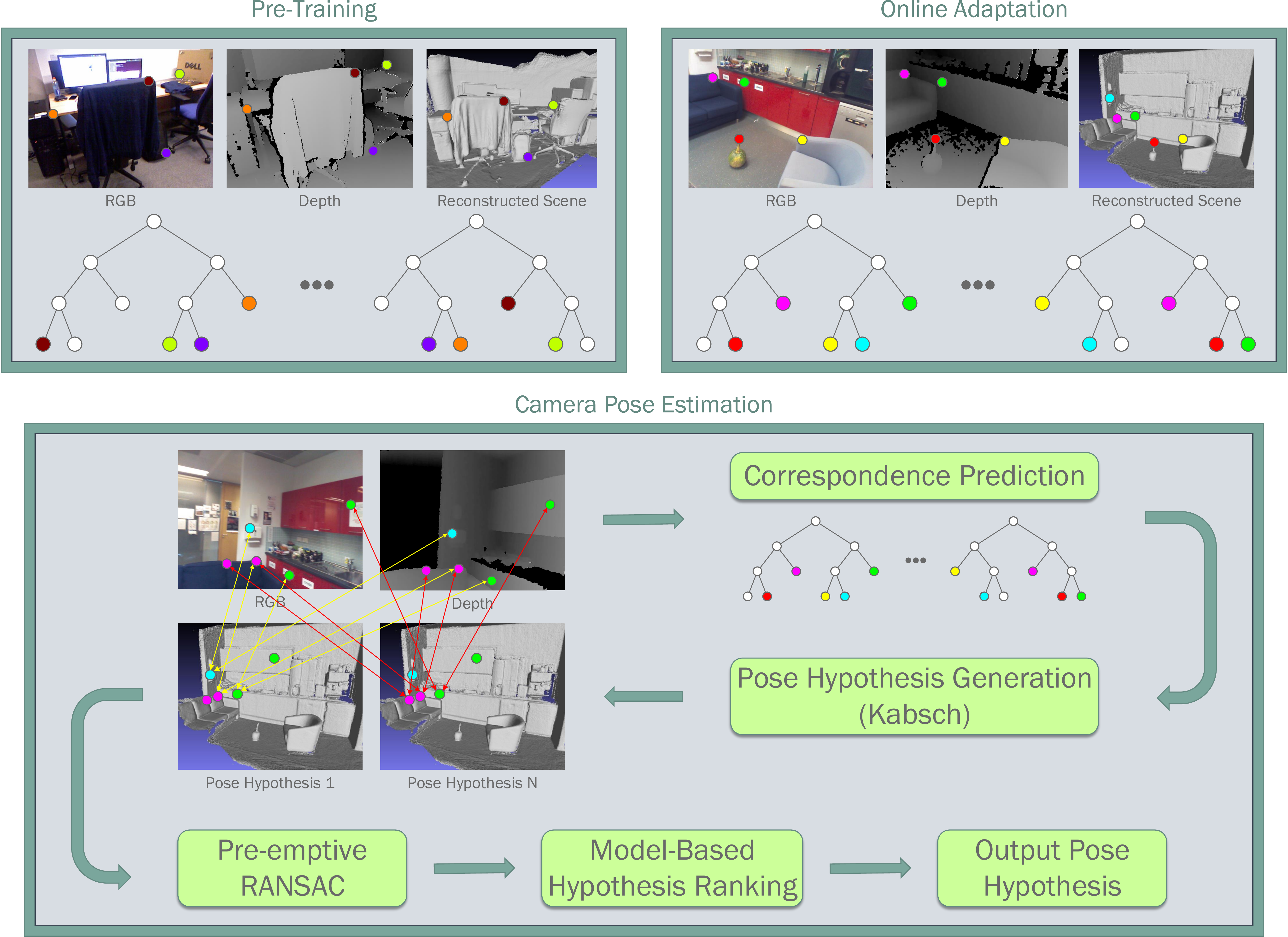}{
	\textbf{Overview of our approach (without the cascade)}. First, we train a regression forest \emph{offline} to predict 2D-to-3D correspondences for a generic scene.
	To adapt this forest to a new scene, we remove the scene-specific information in the forest's leaves while retaining the branching structure (with learned split parameters) of the trees; we then refill the leaves \emph{online} using training examples from the new scene.
	The adapted forest can be deployed to predict correspondences for the new scene, triples of which are then fed to the Kabsch \cite{Kabsch1976} algorithm to generate a large number of pose hypotheses.
	We then reduce these hypotheses to a much smaller number of refined hypotheses using RANSAC \cite{Fischler1981} (in this paper, we modify our RANSAC module from \cite{Cavallari2017} to return multiple hypotheses rather than just a single one).
	Finally, we score and rank the final few hypotheses using a model-based approach, yielding a single resulting output pose.
}{fig:pipeline}{!t}

\subsection{Overview}
\label{subsec::methodoverview}

Our approach is shown in Figure~\ref{fig:pipeline}.
Initially, we train a regression forest \emph{offline} to predict 2D-to-3D
correspondences for a \emph{generic} scene using the approach described in
\cite{Valentin2015RF}. To adapt this forest to a new scene, we remove the
contents of the leaf nodes in the forest (i.e.\ GMM modes and associated covariance matrices)
whilst retaining the branching structure of the trees (including learned split parameters). We then adapt the forest
\emph{online} to the new scene by feeding training examples down the forest to
refill the empty leaves, dynamically learning a set of leaf distributions specific
to that scene. Thus adapted, the forest can then be used to predict
correspondences for the new scene that can be used for camera pose estimation.
Reusing the tree structures spares us from expensive offline learning on
deployment in a novel scene, allowing for relocalisation on the fly.

To estimate the camera pose, we extend the pipeline we proposed in \cite{Cavallari2017},
which fed triples of correspondences to the Kabsch \cite{Kabsch1976} algorithm to
generate pose hypotheses, and then refined them down to a single output pose
using pre-emptive RANSAC. As highlighted in \cite{Cavallari2017}, returning a
single pose from RANSAC has the disadvantage of sometimes yielding the wrong pose
when the energies of the last few candidates considered by RANSAC are relatively
similar (e.g.\ when different parts of the scene look nearly the same). To address this,
we thus add in an additional pipeline step that scores and ranks the last few RANSAC
candidates using an independent, model-based approach (see \S\ref{subsubsec::hypothesisranking}).
This significantly improves the performance of the relocaliser in situations exhibiting
serious appearance aliasing (see \S\ref{sec:experiments}), but at a cost in speed.
To mitigate this, we introduce the concept of a relocalisation cascade (see \S\ref{subsubsec::cascade}),
which runs multiple variants of our relocaliser in sequence, starting with a fast variant
that is less likely to succeed, and progressively falling back to slower, better relocalisers
as the earlier ones fail. This leads to fast average-case relocalisation performance
without significantly compromising on quality.


\subsection{Details}
\label{subsec::methoddetails}

\subsubsection{Offline Forest Training}
\label{subsubsec::forestpretraining}

Training is done as in \cite{Valentin2015RF}, greedily optimising a standard
reduction-in-spatial-variance objective over the randomised parameters of simple
threshold functions. Like \cite{Valentin2015RF}, we make use of `Depth' and
`Depth-Adaptive RGB' (`DA-RGB') features, centred at a pixel $\textbf{p}$, as
follows:
%
\begin{equation}
\textstyle f^{\text{Depth}}_\Omega = D(\mathbf{p}) - D\left(\mathbf{p} + \frac{\boldsymbol{\delta}}{D(\mathbf{p})}\right)
\end{equation}
\begin{equation}
\textstyle f^{\text{DA-RGB}}_\Omega = C(\mathbf{p},c) - C\left(\mathbf{p} + \frac{\boldsymbol{\delta}}{D(\mathbf{p})}, c\right)
\end{equation}
%
In this, $D(\mathbf{p})$ is the depth at $\mathbf{p}$, $C(\mathbf{p},c)$ is the
value of the $c^{\text{th}}$ colour channel at $\mathbf{p}$, and $\Omega$ is a
vector of randomly sampled feature parameters. For `Depth', the only parameter
is the 2D image-space offset $\boldsymbol{\delta}$, whereas `DA-RGB' adds the
colour channel selection parameter $c \in \{R,G,B\}$. We randomly generate 128
values of $\Omega$ for `Depth' and 128 for `DA-RGB'. We concatenate the
evaluations of these functions at each pixel of interest to yield 256D feature
vectors.

At training time, a set $S$ of training examples, each consisting of such a
feature vector $\mathbf{f} \in \mathbb{R}^{256}$, its corresponding 3D location
in the scene and its colour, is assembled via sampling from a ground truth RGB-D
video with known camera poses for each frame (obtained by tracking from depth camera
input). A random subset of these training examples is selected to train each tree in
the forest.

Starting from the root of each tree, we recursively partition the training examples in
the current node into two using a binary threshold function. To decide how to split each
node $n$, we randomly generate a set $\Theta_n$ of $512$ candidate split parameter pairs,
where each $\theta = (\phi,\tau) \in \Theta_n$ denotes the binary threshold function
\begin{equation}
\textstyle \theta(\mathbf{f}) = \mathbf{f}[\phi] \ge \tau.
\end{equation}
In this, $\phi \in [0,256)$ is a randomly-chosen feature index, and $\tau \in \mathbb{R}$
is a threshold, chosen to be the value of feature $\phi$ in a randomly chosen training example.
Examples that pass the test are routed to the right subtree of $n$; the remainder are routed
to the left.
To pick a suitable split function for $n$, we use exhaustive search to find a $\theta^{*} \in \Theta_n$
whose corresponding split function maximises the information gain that can be achieved by splitting the
training examples that reach $n$. Formally, the information gain corresponding to split parameters
$\theta \in \Theta_n$ is
\begin{equation}
\textstyle V(S_n) - \sum_{i\in\{\text{L,R}\}} \frac{|S^i_n(\theta)|}{|S_n|} \; V(S^i_n(\theta)),
\end{equation}
in which $V(X)$ denotes the spatial variance of set $X$, and $S^L_n(\theta)$ and
$S^R_n(\theta)$ denote the left and right subsets into which the set $S_n
\subseteq S$ of training examples reaching $n$ is partitioned by the split
function denoted by $\theta$. Spatial variance is defined in terms of the
log of the determinant of the covariance of a fitted 3D Gaussian \cite{Valentin2015RF}.

For a given tree, the above process is simply recursed to a maximum depth of 15.
We train 5 trees per forest.
The (approximate, empirical) distributions in the leaves are discarded at the end of this process (we replace them during online forest adaptation, as discussed in the next section).

\subsubsection{Online Forest Adaptation}
\label{subsubsec::forestadaptation}

\begin{stusubfig}{!t}
	\begin{subfigure}{.48\linewidth}
		\centering
		\includegraphics[width=\linewidth]{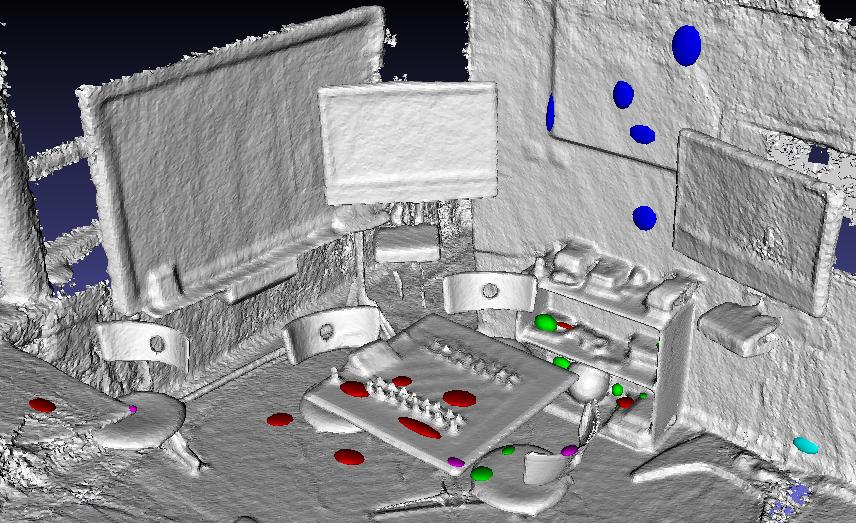}
	\end{subfigure}%
	\hspace{3mm}%
	\begin{subfigure}{.48\linewidth}
		\centering
		\includegraphics[width=\linewidth]{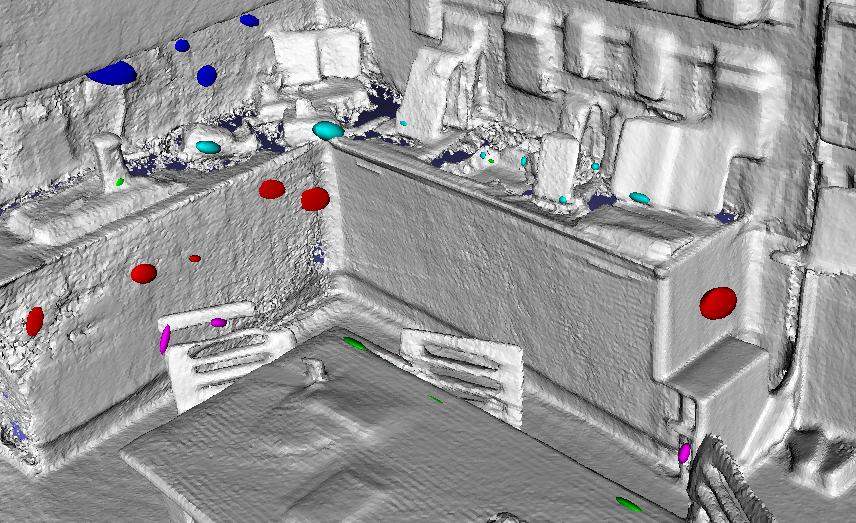}
	\end{subfigure}%
	\caption{An example of the effect that online adaptation has on a
		pre-trained forest: (L) the modal clusters present in a small number of
		randomly selected leaves of a forest pre-trained on the \emph{Chess} scene from
		the 7-Scenes dataset \cite{Shotton2013} (the colour of each mode indicates its
		containing leaf); (R) the modal clusters that are added to the same leaves
		during the process of adapting the forest to the \emph{Kitchen} scene.}
	\label{fig:modes}
	\vspace{-1.5\baselineskip}
\end{stusubfig}

To adapt a forest to a new scene, we replace the distributions discarded from its leaves at the end of pre-training with dynamically updated ones drawn entirely from the new scene.
Here, we detail how the new leaf distributions used by the relocaliser are computed and updated online.

We draw inspiration from the use of reservoir sampling \cite{Vitter1985} in SemanticPaint \cite{Valentin2015SP}, which makes it possible to store an unbiased subset of an empirical distribution in a bounded amount of memory.
On initialisation, we allocate (on the GPU) a fixed-size sample reservoir for each leaf of the existing forest.
Our reservoirs contain up to $\kappa$ entries, each of which stores a 3D location (in world coordinates) and an associated colour.
At runtime, we pass training examples (of the form described in \S\ref{subsubsec::forestpretraining}) down the forest and identify the leaves to which each example is mapped.
We then add the 3D location and colour of each example to the reservoirs associated with its leaves.

To obtain the 3D locations of the training examples, we need to know the transformation that maps points from camera space to world space.
When training on sequences from a dataset, this is trivially available as the ground truth camera pose, but in a live scenario, it will generally be obtained as the output of a fallible tracker.\footnote{We are largely agnostic to the camera tracker used, but in keeping with our scenario of relocalising in a known scene, at least some frames must be tracked reliably to allow the relocaliser to be trained.}
To avoid corrupting our forest's reservoirs, we avoid passing new examples down the forest when tracking is unreliable.
We measure tracker reliability using the support vector machine (SVM) approach described in \cite{Kaehler2016}.
For frames for which a reliable camera pose \emph{is} available, we proceed as follows:
\begin{enumerate}
	\item First, we compute feature vectors for a subset of the pixels in the image, as detailed in \S\ref{subsubsec::forestpretraining}. We empirically choose our subset by subsampling densely on a regular grid with $4$-pixel spacing, i.e.\ we choose pixels $\{(4i,4j) \in [0,w) \times [0,h) : i,j \in \mathbb{N}^+ \}$, where $w$ and $h$ are respectively the width and height of the image.
	\item Next, we pass each feature vector down the forest, adding the 3D position and colour of the corresponding scene point to the reservoir of the leaf reached in each tree. Our CUDA-based random forest implementation uses the node indexing described in~\cite{Sharp2008}.
	\item Finally, for each leaf reservoir, we cluster the contained points using a CUDA implementation of Really Quick Shift (RQS) \cite{Fulkerson2010} to find a set of modal 3D locations. We sort the clusters in each leaf in decreasing size order, and keep at most $M_{\max}$ modal clusters per leaf. For each cluster we keep, we compute 3D and colour centroids, and a covariance matrix. The cluster distributions are used when estimating the likelihood of a camera pose, and also during continuous pose optimisation (see \S\ref{subsubsec:poseestimation}).
	Since running RQS over all the leaves in the forest would take too long if run in a single frame, we amortise the cost over multiple frames by updating $256$ leaves in parallel each frame in round-robin fashion. A typical forest contains around $42,000$ leaves, so each leaf is updated roughly once every $6$s.
\end{enumerate}
Figure~\ref{fig:modes} illustrates the effect that
online adaptation has on a pre-trained forest: (a) shows the modal clusters
present in a small number of randomly selected leaves of a forest pre-trained on
the \emph{Chess} scene from the 7-Scenes dataset \cite{Shotton2013}; (b) shows
the modal clusters that are added to the same leaves during the process of
adapting the forest to the \emph{Kitchen} scene. Note that whilst the positions
of the predicted modes have (unsurprisingly) completely changed, the split
functions in the forest's branch nodes (which we preserve) still do a good
job of routing similar parts of the scene into the same leaves, enabling
effective sampling of 2D-to-3D correspondences for camera pose estimation.

\subsubsection{Camera Pose Estimation}
\label{subsubsec:poseestimation}

As in \cite{Valentin2015RF}, camera pose estimation is based on the preemptive, locally-optimised RANSAC of \cite{Chum2003}.
We begin by randomly generating an initial set of up to $N_{\max}$ pose hypotheses.
A pose hypothesis $H \in \mathbf{SE}(3)$ is a transform that maps points in camera space to world space.
To generate each pose hypothesis, we apply the Kabsch algorithm \cite{Kabsch1976} to $3$ point pairs of the form $(\mathbf{x}_i^\mathcal{C}, \mathbf{x}_i^\mathcal{W})$, where $\mathbf{x}_i^\mathcal{C} = D(\mathbf{u}_i) K^{-1} \dot{\mathbf{u}}_i$ is obtained by back-projecting a randomly chosen point $\mathbf{u}_i$ in the live depth image $D$ into camera space, and $\mathbf{x}_i^\mathcal{W}$ is a corresponding scene point in world space, randomly sampled from $M(\mathbf{u}_i)$, the modes of the leaves to which the forest maps $\mathbf{u}_i$.
In this, $K$ denotes the depth camera intrinsics.
We subject each hypothesis to three checks:
\begin{enumerate}
	\item First, we randomly choose one of the three point pairs $(\mathbf{x}_i^\mathcal{C},\mathbf{x}_i^\mathcal{W})$ and compare the RGB colour of the corresponding pixel $\mathbf{u}_i$ in the colour input image to the colour centroid of the mode (see \S\ref{subsubsec::forestadaptation}) from which we sampled $\mathbf{x}_i^\mathcal{W}$. We reject the hypothesis iff the L$^\infty$ distance between the two exceeds a threshold.
	\item Next, we check that the three hypothesised scene points are sufficiently far from each other. We reject the hypothesis iff the minimum distance between any pair of points is below a threshold (tuned, but generally $30$cm).
	\item Finally, we check that the distances between all scene point pairs and their corresponding back-projected depth point pairs are sufficiently similar, i.e.\ that the hypothesised transform is `rigid enough'. We reject the hypothesis iff this is not the case.
\end{enumerate}
If a hypothesis gets rejected by one of the checks, we try to generate an alternative hypothesis to replace it.
In practice, we use $N_{\max}$ dedicated threads, each of which attempts to generate a single hypothesis.
Each thread continues generating hypotheses until either (a) it finds a hypothesis that passes all of the checks, or (b) a maximum number of iterations is reached.
We proceed with however many hypotheses we obtain by the end of this process.

Having generated our large initial set of hypotheses, we next aggressively prune it by scoring each hypothesis and keeping the $N_\textup{cull}$ lowest-energy transforms (if there are fewer than $N_\textup{cull}$ hypotheses, we keep all of them).
To score the hypotheses, we first select an initial set $I = \{i\}$ of pixel indices in $D$ (of size $\eta$), and back-project the denoted pixels $\mathbf{u}_i$ to corresponding points $\mathbf{x}_i^\mathcal{C}$ in camera space as described above.
We then score each hypothesis $H$ by summing the Mahalanobis distances between the transformations of each $\mathbf{x}_i^\mathcal{C}$ under $H$ and their nearest modes:
\begin{equation}
\textstyle E(H) = \sum_{i \in I} \left( \min_{(\boldsymbol{\mu},\Sigma) \in M(\mathbf{u}_i)} \left\| \Sigma^{-\frac{1}{2}} (H\mathbf{x}_i^\mathcal{C} - \boldsymbol{\mu}) \right\| \right)
\end{equation}
After this initial cull, we use pre-emptive RANSAC to prune the remaining $\le N_\textup{cull}$ hypotheses to a much smaller set of chosen hypotheses, which will then be scored and ranked (see \S\ref{subsubsec::hypothesisranking}). (This differs from \cite{Cavallari2017}, where our RANSAC module was designed to output a single hypothesis, and no subsequent scoring was performed.)
We iteratively (i) expand the sample set $I$ (by adding $\eta$ new pixels each time), (ii) refine the pose candidates via Levenberg-Marquardt optimisation \cite{Levenberg1944,Marquardt1963} of the energy function $E$, (iii) re-evaluate and re-score the hypotheses, and (iv) discard the worse half. We stop when the number of hypotheses remaining reaches a desired threshold.
The actual optimisation is performed not in $\mathbf{SE}(3)$, where it would be hard to do, but in the corresponding Lie algebra, $\mathfrak{se}(3)$.
See \cite{Valentin2015RF} for details of this process, and \cite{Strasdat2012} for a longer explanation of Lie algebras.

In \cite{Cavallari2017}, this process yielded a single pose hypothesis, which it was possible to either return directly, or, if the relocaliser was integrated into a 3D reconstruction framework such as InfiniTAM \cite{Prisacariu2017}, return after first refining it using ICP \cite{Besl1992}. Here, the process instead yields a set of chosen hypotheses, from which we then need to select a single, final output pose. To do this, we assume the presence of a 3D scene model, since 3D reconstruction is a key application of our approach, and propose a model-based way of ranking the hypotheses to choose a best pose (if a 3D model is not present, one of the hypotheses can be returned as-is, or the hypotheses can be scored and ranked in a different way).

\subsubsection{Model-Based Hypothesis Ranking}
\label{subsubsec::hypothesisranking}

\stufig{width=.8\linewidth}{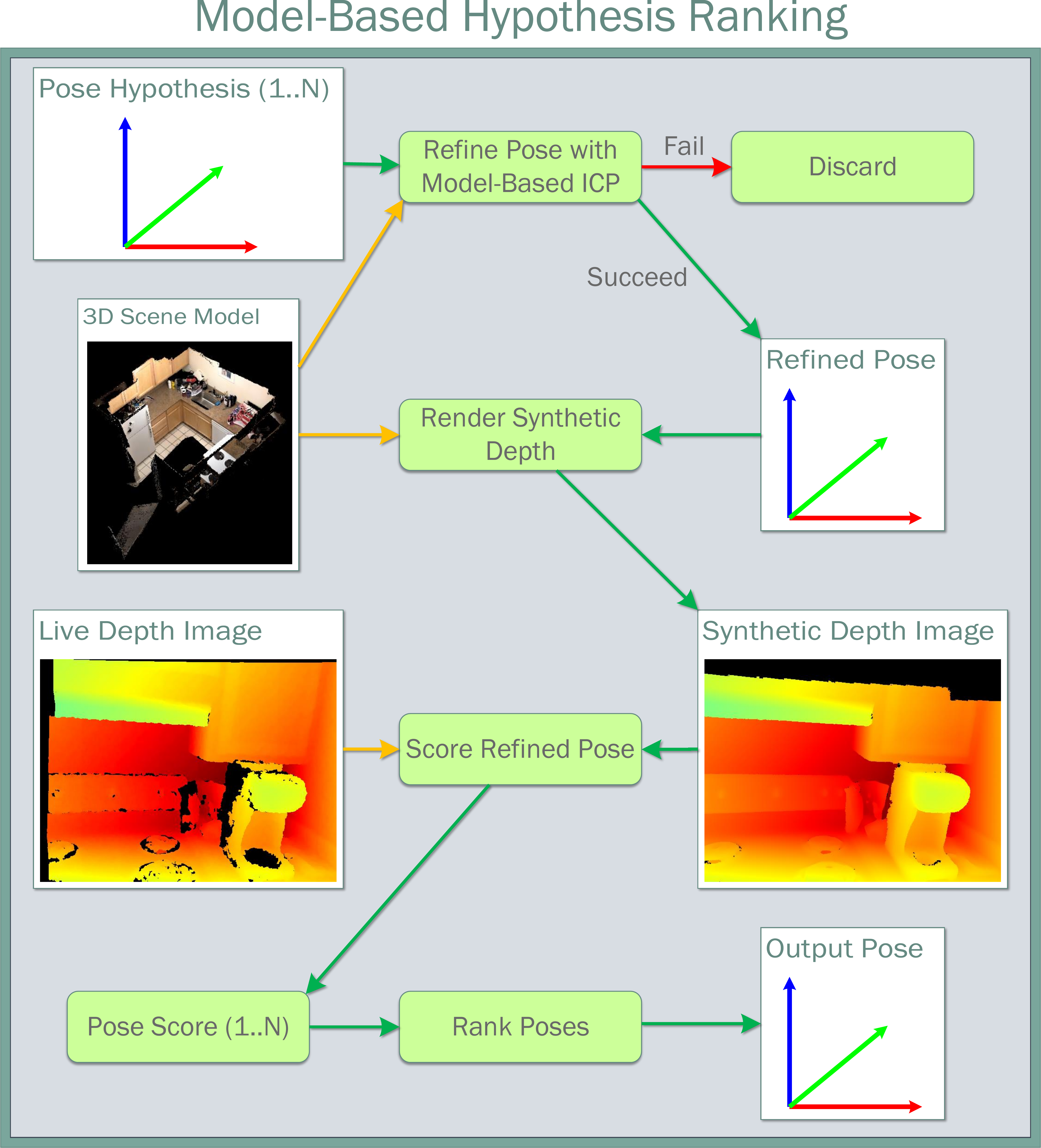}{Our model-based approach to ranking the camera pose hypotheses that survive the RANSAC stage (see \S\ref{subsubsec::hypothesisranking}). For each hypothesis, we first refine the pose by performing ICP \cite{Besl1992} with respect to the 3D scene model. If this succeeds, we score the hypothesis by comparing a synthetic depth raycast from the refined pose to the live depth image from the camera. Once all hypotheses have been scored, we rank them and return the one with the lowest score.}{fig:hypothesisranking}{!t}

To score a hypothesis chosen by RANSAC, we first refine it using ICP \cite{Besl1992} with respect to the 3D scene model (see Figure~\ref{fig:hypothesisranking}). If this fails, we discard the pose, since regardless of its correctness, it was not a pose that would have been good enough to allow tracking to resume. If this succeeds, we further verify correctness by rendering a synthetic depth raycast of the scene model from the refined pose, and comparing it to the live depth image from the camera.\footnote{Comparing a colour raycast to the live colour image is also possible, but we found depth-based ranking to be more effective in practice.} To do this, we draw inspiration from \cite{Golodetz2018}, which compared synthetic depth raycasts to verify a relative transform estimate between two different sub-scenes. By contrast, we use comparisons between the live depth image and synthetic raycasts of a single scene to compute scores that can be used to rank the different pose hypotheses against each other.

Formally, let $D_\ell$ be the live depth image, $\Xi = \{\xi_1,...\xi_k\}$ be the set of chosen pose hypotheses, and $\tilde{D}_\xi$ be a synthetic depth raycast of the 3D model from pose $\xi$. Moreover, for any depth image $D$, let $\Omega(D)$ denote the domain of $D$ and $\Omega^v(D)$ denote the range of pixels $\mathbf{x}$ for which $D(\mathbf{x})$ is valid, and let $\Omega^v_{\ell,\xi} = \Omega^v(D_\ell) \cap \Omega^v(\tilde{D}_\xi)$. To assign a score to hypothesis $\xi$, we first compute a mean (masked) absolute depth difference between $D_\ell$ and $\tilde{D}_\xi$ via
\begin{equation}
\textstyle \mu(\xi) = \left( \sum_{\mathbf{x} \in \Omega^v_{\ell,\xi}} \left| D_\ell(\mathbf{x}) - \tilde{D}_\xi(\mathbf{x}) \right| \right) / |\Omega^v_{\ell,\xi}|,
\end{equation}
similar to Equation (4) in \cite{Golodetz2018}. We then compute a final score $s(\xi)$ for $\xi$ via
\begin{equation}
\textstyle s(\xi) = \begin{cases}
\mu(\xi) & \mbox{if } |\Omega^v(\tilde{D}_\xi)| / |\Omega(\tilde{D}_\xi)| \ge 0.1 \\
\infty & \mbox{otherwise.}
\end{cases}
\end{equation}
In this, the purpose of the (empirically chosen) threshold is to ensure that the hypothesised pose points sufficiently towards the 3D scene model to allow it to be verified: we found $0.1$ to be effective in practice. Having scored all of the hypotheses in this way, we can then simply pick the pose $\xi^*$ with the lowest score (i.e.\ the one whose synthetic depth raycast was closest to the live depth) and return it:
\begin{equation}
\textstyle \xi^* = \argmin_{\xi \in \Xi} s(\xi)
\end{equation}

\subsubsection{Relocalisation Cascade}
\label{subsubsec::cascade}

\stufig{width=\linewidth}{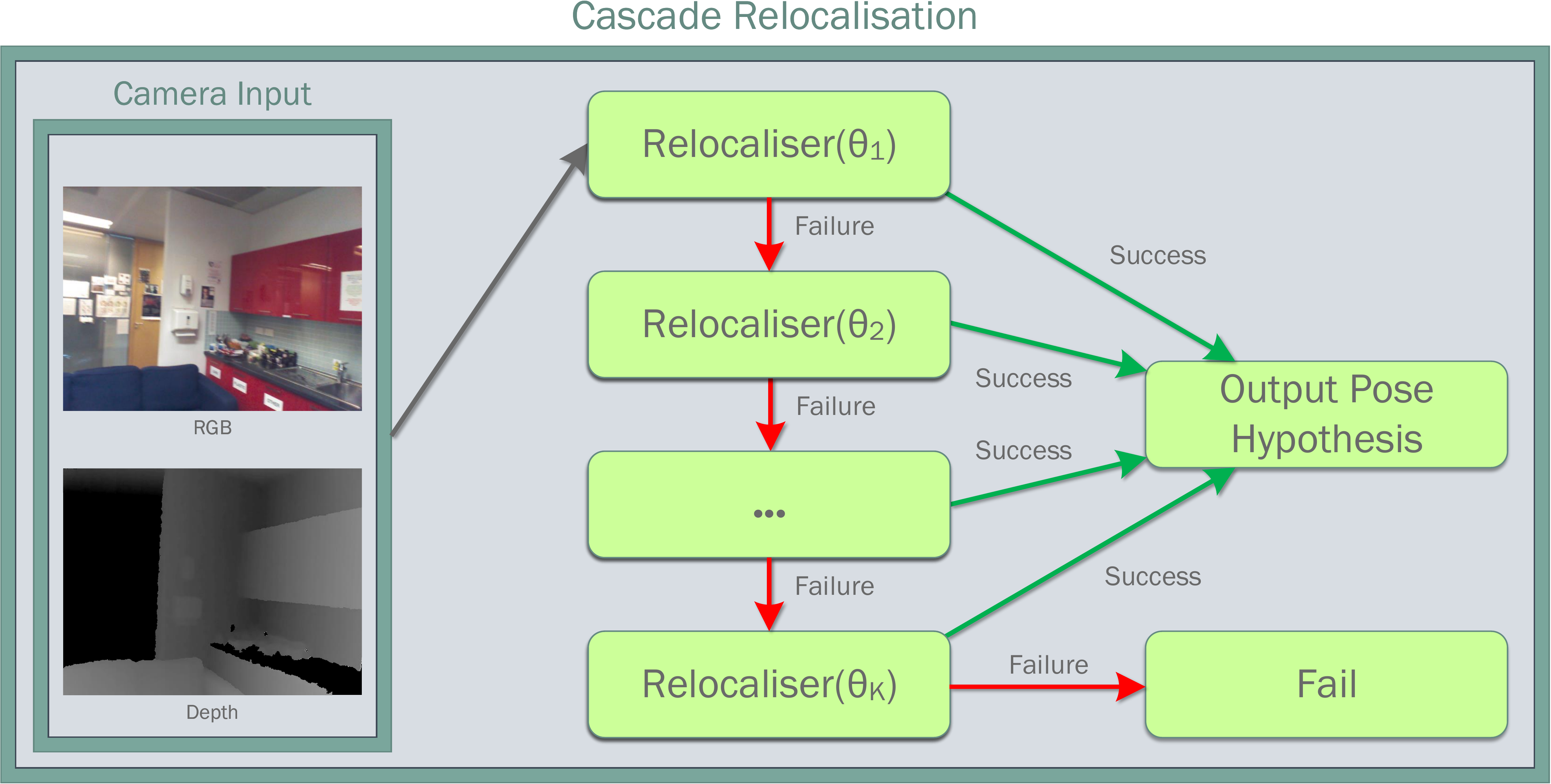}{Our relocalisation cascade (see \S\ref{subsubsec::cascade}): we instantiate multiple instances of our relocaliser, backed by the same regression forest, but with different hypothesis generation and RANSAC parameters, and run them one at a time on the camera input until one of them produces an acceptable pose (or we reach the end of the cascade). The idea is to gradually fall back from fast but less effective relocalisers to slower but more effective ones, with the aim of yielding an effective overall relocaliser that is fast on average.}{fig:cascade}{!t}

Ranking the last few RANSAC candidates as described in \S\ref{subsubsec::hypothesisranking} significantly improves relocalisation performance in scenarios exhibiting serious appearance aliasing (see \S\ref{sec:experiments}), but can in practice be quite expensive (clearly undesirable in an interactive SLAM context) because of the need to perform ICP for each candidate. In practice, however, many scenarios do not exhibit such aliasing, and in those contexts, hypothesis ranking provides little benefit. As an additional contribution, we thus propose a novel \emph{relocalisation cascade} approach that uses hypothesis ranking only when necessary.

Figure~\ref{fig:cascade} shows how this works: we instantiate multiple instances of our relocaliser, backed by the same regression forest, but with different parameters for the hypothesis generation and RANSAC steps, and run them one at a time on the camera input until one of them produces an acceptable pose (or we reach the end of the cascade). The idea is to put `faster' relocalisers (i.e.\ ones that have been tuned more for speed than performance) towards the start of the cascade, and `slower' ones (i.e.\ ones that have been tuned more for performance than speed) towards the end of it, yielding an effective cascade that is fast on average. The key to making this work well is in how to decide whether or not a given relocaliser in the cascade has produced an acceptable pose. If we accept poses produced by early-stage relocalisers too readily, the cascade's relocalisation quality will suffer; conversely, if we are too draconian in rejecting such poses, then the cascade will be slow (since we will be running the slower, late-stage relocalisers far too often).

In practice, the depth-difference scores computed during hypothesis ranking (see \S\ref{subsubsec::hypothesisranking}) provide us with an effective basis on which to make these decisions: in particular, it suffices to fall back from one relocaliser in the cascade to the next iff the score associated with the final output pose of the relocaliser is above a threshold (reflecting a high likelihood of an incorrect pose). The thresholds used are important parameters of the cascade: the way in which we tune both them and the parameters of the relocalisers in the cascade is described in \S\ref{sec:parametertuning}.

\section{Experiments}
\label{sec:experiments}

We perform both quantitative and qualitative experiments to evaluate our approach.
In \S\ref{subsec:headlineperformance}, we compare several variants of our approach,
trained on the \emph{Office} sequence from the 7-Scenes dataset \cite{Shotton2013}
and then adapted to each target scene, to state-of-the-art \emph{offline} relocalisers
trained directly on the target scene. We show that our adapted forests achieve
state-of-the-art relocalisation performance despite being trained on a different
scene, allowing them to be used for \emph{online} relocalisation. Further results,
showing that very similar performances can be obtained with forests trained on the other
7-Scenes sequences, can be found in \S\ref{sec:additionalexperiments}, along with detailed
timings for each stage of our pipeline.
In \S\ref{subsec:trackinglossrecovery}, we show our ability
to adapt forests on-the-fly from live sequences, allowing us to
support tracking loss recovery in interactive scenarios.
In \S\ref{subsec:novelposes}, we evaluate how well our approach
generalises to novel poses in comparison to a keyframe-based random fern
relocaliser based on \cite{Glocker2015}.
Our results show that we are able to relocalise well even from poses that are quite far
away from the training trajectory.
In \S\ref{subsec:forestvisualisation},
we visualise the internal behaviour of SCoRe forests, and use this to explain why
the behaviour of a forest is relatively independent of the specific scene on which
it was trained. Finally, in \S\ref{subsec:nopretraining}, we use the insights gained
from this visualisation to show that pre-training can be avoided entirely by replacing
the pre-trained forest with a generated one that plays the same role.

\subsection{Headline Performance}
\label{subsec:headlineperformance}

\newcommand{\best}[1]{\textbf{\textcolor{red}{#1}}}
\newcommand{\secondbest}[1]{\textcolor{blue}{#1}}

\iftrue
\begin{table*}[!t]
	\centering
	\scriptsize
	\begin{tabular}{lcccccccccc}
		\toprule
		& \textbf{Chess}       & \textbf{Fire}        & \textbf{Heads} & \textbf{Office} & \textbf{Pumpkin} & \textbf{Kitchen} & \textbf{Stairs} & \textbf{Average} & \textbf{Avg.\ Med.\ Error} & \textbf{Frame Time (ms)} \\
		\midrule
		\multicolumn{5}{l}{\textbf{RGB-D, online, single-frame}} \\
		\midrule
		Ours (Default) & 99.75\%              & 97.35\%              & \best{100\%}         & \secondbest{99.80\%} & 82.25\%              & \best{95.64\%} & 79.10\%              & 93.41\%              & \best{0.011m}/1.16$^\circ$       & 128 \\
		+ ICP          & 99.85\%              & 99.15\%              & \best{100\%}         & \best{99.85\%}       & 90.00\%              & 91.52\%        & 80.00\%              & 94.34\%              & 0.013m/1.16$^\circ$ & 133 \\
		+ Ranking      & \best{99.95\%}       & \best{99.70\%}       & \best{100\%}         & 99.48\%              & 90.85\%              & 90.68\%        & \best{94.20\%}       & \best{96.41\%}       & 0.013m/1.17$^\circ$ & 257 \\
		\midrule
		Ours (Fast) w/ICP
		& 99.75\%              & 97.10\%              & 98.40\%              & 99.55\%              & 89.35\%              & 89.26\%        & 62.40\%              & 90.83\%              & 0.014m/1.17$^\circ$              & \best{30} \\
		Ours (Cascade F$\stackrel{7.5\textup{cm}}{\rightarrow}$S)
		& \secondbest{99.90\%} & 98.95\%              & \secondbest{99.90\%} & 99.48\%              & \secondbest{90.95\%} & 89.34\%        & 86.10\%              & 94.95\%              & 0.013m/1.17$^\circ$ & \secondbest{52} \\
		Ours (Cascade F$\stackrel{5\textup{cm}}{\rightarrow}$I$\stackrel{7.5\textup{cm}}{\rightarrow}$S)
		& 99.85\%              & \secondbest{99.40\%} & \secondbest{99.90\%} & 99.40\%              & 90.85\%              & 89.64\%        & \secondbest{89.80\%} & \secondbest{95.55\%} & 0.013m/1.17$^\circ$ & 66 \\
		\midrule
		Ours (Random)  & 99.80\% & 96.90\% & \best{100\%} & 98.48\% & 78.65\% & 91.98\% & 71.60\% & 91.06\% & \secondbest{0.012m}/1.18$^\circ$ & 121 \\
		+ ICP          & 99.85\% & 98.50\% & \best{100\%} & 99.10\% & 89.50\% & 90.32\% & 77.80\% & 93.58\% & 0.013m/1.16$^\circ$ & 126 \\
		\midrule
		Ours (2017) \cite{Cavallari2017} & 99.2\% & 96.5\% & 99.7\% & 97.6\% & 84.0\% & 81.7\% & 33.6\% & 84.6\% & -- & -- \\
		+ ICP                            & 99.4\% & 99.0\% & \best{100\%} & 98.2\% & \best{91.2\%} & 87.0\% & 35.0\% & 87.1\% & -- & 141 \\
		\midrule
		\multicolumn{5}{l}{\textbf{RGB-D, offline, single-frame}} \\
		\midrule
		Shotton \emph{et al.} \cite{Shotton2013}            & 92.6\% & 82.9\% & 49.4\% & 74.9\% & 73.7\% & 71.8\% & 27.8\% & 67.6\% & -- & -- \\
		Guzman-Rivera \emph{et al.} \cite{GuzmanRivera2014} & 96\% & 90\% & 56\% & 92\% & 80\% & 86\% & 55\% & 79.3\% & -- & -- \\
		Valentin \emph{et al.} \cite{Valentin2015RF}        & 99.4\% & 94.6\% & 95.9\% & 97.0\% & 85.1\% & 89.3\% & 63.4\% & 89.5\% & -- & -- \\
		Brachmann \emph{et al.} \cite{Brachmann2016}        & 99.6\% & 94.0\% & 89.3\% & 93.4\% & 77.6\% & 91.1\% & 71.7\% & 88.1\% & 0.061m/2.7$^\circ$ & -- \\
		Meng \emph{et al.} \cite{Meng2017arXiv}             & 99.5\% & 97.6\% & 95.5\% & 96.2\% & 81.4\% & 89.3\% & 72.2\% & 90.3\% & 0.017m/\best{0.70$^\circ$} & -- \\
		Schmidt \emph{et al.} \cite{Schmidt2017}            & 97.75\% & 96.55\% & 99.8\% & 97.2\% & 81.4\% & \secondbest{93.4\%} & 77.7\% & 92.0\% & -- & -- \\
		Brachmann and Rother \cite{Brachmann2018CVPR}       & 97.1\% & 89.6\% & 92.4\% & 86.6\% & 59.0\% & 66.6\% & 29.3\% & 76.1\% & 0.036m/\secondbest{1.1$^\circ$} & -- \\
		\midrule
		\multicolumn{5}{l}{\emph{\textbf{RGB-only, offline, single-frame}}} \\
		\midrule
		\emph{Brachmann and Rother \cite{Brachmann2018CVPR}} & \emph{93.8\%} & \emph{75.6\%} & \emph{18.4\%} & \emph{75.4\%} & \emph{55.9\%} & \emph{50.7\%} & \emph{2.0\%}  & \emph{60.4\%} & \emph{0.084m/2.4$^\circ$}   & -- \\
		\emph{Li \emph{et al.} \cite{Li2018arXiv}}           & \emph{96.1\%} & \emph{88.6\%} & \emph{86.9\%} & \emph{80.6\%} & \emph{60.3\%} & \emph{61.9\%} & \emph{11.3\%} & \emph{71.8\%} & \emph{0.043m/1.3$^\circ$}   & -- \\
		\midrule
		\multicolumn{5}{l}{\emph{\textbf{RGB-only, offline, sequential}}} \\
		\midrule
		\emph{Valada \emph{et al.} \cite{Valada2018}}       & --     & --     & --     & --     & --     & --     & --     & \emph{59.1\%} & \emph{0.048m/3.801$^\circ$} & -- \\
		\emph{Radwan \emph{et al.} \cite{Radwan2018}}       & --     & --     & --     & --     & --     & --     & --     & \emph{99.2\%} & \emph{0.013m/0.77$^\circ$}  & \emph{79} \\
		\bottomrule
	\end{tabular}
	\caption{Comparing our \emph{adaptive} approach to state-of-the-art \emph{offline} methods on the 7-Scenes dataset \cite{Shotton2013} (the percentages denote proportions of test frames with $\le 5$cm translation error and $\le 5^\circ$ angular error; red and blue colours denote respectively the best and second-best results in each column). Italics denote results that are not directly comparable (see \S\ref{subsec:headlineperformance}). `+~Ranking' means ranking the last $16$ candidates produced by RANSAC, as per \S\ref{subsubsec::hypothesisranking}. For all versions of our method except \emph{Ours (Random)}, we report the results obtained by adapting a forest pre-trained on the \emph{Office} sequence from 7-Scenes. We achieve results that are superior to the offline methods, without needing to pre-train on the test scene. Moreover, our F$\stackrel{7.5\textup{cm}}{\rightarrow}$S cascade achieves state-of-the-art results whilst running at nearly $20$ FPS. Finally, \emph{Ours (Random)} uses a randomly generated forest (see \S\ref{subsec:nopretraining}) and achieves state-of-the-art results without any offline training \emph{on any scene}.}
	\label{tbl:comparativeperformance7}
\end{table*}
\fi

\iftrue
\begin{table*}[!t]
	\centering
	\scriptsize
	\begin{tabular}{lccccccccc}
		\toprule
		\textbf{Sequence} & \textbf{LTN} \cite{Valentin2016} & \textbf{BTBRF} \cite{Meng2017IROS} & \textbf{PLForests} \cite{Meng2017arXiv} & \textbf{Ours} & \textbf{+ ICP} & \textbf{+ Ranking} & \textbf{Ours (F$\stackrel{7.5\textup{cm}}{\rightarrow}$S)} & \textbf{Ours (Random)} & \textbf{+ ICP} \\
		\midrule
		Kitchen  & 85.7\% & 92.7\%              & 98.9\%              & \best{100\%}         & \best{100\%}         & \best{100\%}   &   \best{100\%} & \secondbest{99.72\%} &   \best{100\%} \\
		Living   & 71.6\% & 95.1\%              & \best{100\%}        & \secondbest{99.80\%} & \best{100\%}         & \best{100\%}   &   \best{100\%} & 99.59\% &   \best{100\%} \\
		\midrule
		Bed      & 66.4\% & 82.8\%              & \secondbest{99.0\%} & \best{100\%}         & \best{100\%}         & \best{100\%}   &   \best{100\%} &   \best{100\%} &   \best{100\%} \\
		Kitchen  & 76.7\% & 86.2\%              & 99.0\% & \best{100\%}         & \best{100\%}         & \best{100\%}   & \secondbest{99.52\%} &   \best{100\%} &   \best{100\%} \\
		Living   & 66.6\% & \secondbest{99.7\%} & \best{100\%}        & \best{100\%}         & \best{100\%}         & \best{100\%}   &   \best{100\%} &   \best{100\%} &   \best{100\%} \\
		Luke     & 83.3\% & 84.6\%              & \secondbest{98.9\%} & 97.92\%              & \best{99.20\%}       & \best{99.20\%} & \best{99.20\%} & 96.31\% & \best{99.20\%} \\
		\midrule
		Floor5a  & 66.2\% & 89.9\%              & 98.8\%              & 99.20\% & \best{100\%}         & \best{100\%}   & \secondbest{99.60\%} & 98.59\% &   \best{100\%} \\
		Floor5b  & 71.1\% & 98.9\%              & 99.0\%              & \secondbest{99.75\%} & 99.01\%              & \best{100\%}   & 99.01\% & 99.26\% & 99.01\% \\
		\midrule
		Gates362 & 51.8\% & \secondbest{96.7\%} & \best{100\%}        & \best{100\%}         & \best{100\%}         & \best{100\%}   &   \best{100\%} &   \best{100\%} &   \best{100\%} \\
		Gates381 & 52.3\% & 92.9\%              & 98.8\%              & 99.24\% & \best{100\%}         & \best{100\%}   & \secondbest{99.91\%} & 98.10\% & \secondbest{99.91\%} \\
		Lounge   & 64.2\% & 94.8\%              & \secondbest{99.1\%} & \best{100\%}         & \best{100\%}         & \best{100\%}   &   \best{100\%} &   \best{100\%} &   \best{100\%} \\
		Manolis  & 76.0\% & \secondbest{98.0\%} & \best{100\%}        & \best{100\%}         & \best{100\%}         & \best{100\%}   &   \best{100\%} &   \best{100\%} &   \best{100\%} \\
		\midrule
		\textbf{Average} & 67.4\% & 92.7\% & 99.3\% & 99.66\% & \secondbest{99.85\%} & \best{99.93\%} & 99.77\% & 99.30\% & 99.84\% \\
		\textbf{Avg.\ Med.\ Trans.}  & -- & -- & -- & 0.009m & \best{0.006m} & \secondbest{0.007m} & \secondbest{0.007m} & 0.010m & \secondbest{0.007m} \\
		\textbf{Avg.\ Med.\ Rot.}    & -- & -- & -- & \secondbest{0.51$^\circ$} & \best{0.26$^\circ$} & \best{0.26$^\circ$} & \best{0.26$^\circ$} & 0.54$^\circ$ & \best{0.26$^\circ$} \\
		\textbf{Frame Time (ms)}     & -- & -- & -- & 122 & 127 & 240 & \best{33} & \secondbest{119} & 123 \\
		\bottomrule
	\end{tabular}
	\caption{Comparing our \emph{adaptive} approach to state-of-the-art \emph{offline} methods on the Stanford 4 Scenes dataset \cite{Valentin2016}. See Table~\ref{tbl:comparativeperformance7} for an explanation of the different variants of our approach. All variants achieve state-of-the-art results, equalling or exceeding even the near-perfect results of \cite{Meng2017arXiv}. Moreover, our F$\stackrel{7.5\textup{cm}}{\rightarrow}$S cascade can achieve such results at $30$ FPS.}
	\label{tbl:comparativeperformance12}
	\vspace{-\baselineskip}
\end{table*}
\fi

To evaluate the headline performance of our approach, we compare our relocaliser to a variety of state-of-the-art \emph{offline} relocalisers on the 7-Scenes \cite{Shotton2013} and Stanford 4 Scenes \cite{Valentin2016} benchmarks (see Tables~\ref{tbl:comparativeperformance7} and \ref{tbl:comparativeperformance12}).\footnote{For completeness, we include \cite{Brachmann2018CVPR,Li2018arXiv,Valada2018} and \cite{Radwan2018} in Table~\ref{tbl:comparativeperformance7} alongside the other results. However, it is important to note that they are not directly comparable to the other approaches: on the one hand, they are not allowed to use depth, which puts them at a disadvantage; on the other hand, \cite{Valada2018} and \cite{Radwan2018} make use of the estimated pose from the previous frame, which is unavailable under the standard evaluation protocol and gives them a significant advantage. For these reasons, we italicise all four sets of results to make it clear that they cannot be directly compared to the other methods in the table.}

We test several variants of our approach, using the following testing procedure. For all variants of our approach except \emph{Ours (Random)}, we first pre-train a forest on a generic scene (\emph{Office} from 7-Scenes \cite{Shotton2013}) and remove the contents of its leaves, as described in \S\ref{sec:method}: this process runs \emph{offline} over a number of hours (but we only need to do it once). Next, we adapt the forest by feeding it new examples from a training sequence captured on the scene
of interest: this runs \emph{online} at frame rate (in a real system, this allows us to start relocalising almost immediately whilst training carries on in the background, as we show in \S\ref{subsec:trackinglossrecovery}).
Finally, we test the adapted forest by using it to relocalise from every frame of a separate testing sequence captured on the scene of interest.

\emph{Ours (Default)} is an improved variant of \cite{Cavallari2017}, in which each leaf can now store up to $50$ modes. With hypothesis ranking of the last $16$ candidates enabled, this approach achieves an average of $96.41\%$ of frames within $5$cm/$5^\circ$ of the ground truth on 7-Scenes, beating the previous state-of-the-art \cite{Schmidt2017} by over $4$\%. However, as mentioned in \S\ref{subsubsec::cascade}, hypothesis ranking significantly slows down the speed of the relocaliser: we thus present several faster variants of our approach. \emph{Ours (Fast)} is a variant tuned for maximum speed (see Table~\ref{tbl:parameters} for the tuned parameters). With ICP enabled, it is able to achieve an average of over $90$\% on 7-Scenes in under $30$ms; however, it achieves slightly lower performance on the difficult \emph{Stairs} sequence. To achieve a better trade-off between performance and speed, we present two relocalisation cascades (see \S\ref{subsubsec::cascade}), F$\stackrel{7.5\textup{cm}}{\rightarrow}$S and F$\stackrel{5\textup{cm}}{\rightarrow}$I$\stackrel{7.5\textup{cm}}{\rightarrow}$S, each of which starts with our \emph{Fast} relocaliser and gradually falls back to slower, more effective relocalisers if the depth difference for the currently predicted pose exceeds the specified thresholds. We describe the tuning of the parameters of the individual relocalisers in the cascades and the depth-difference thresholds used to fall back between them in \S\ref{sec:parametertuning}. Both cascades presented achieve state-of-the-art results in under $70$ms, offering high-quality performance at frame rate (nearly $20$ FPS in the case of F$\stackrel{7.5\textup{cm}}{\rightarrow}$S).

Finally, \emph{Ours (Random)} denotes a variant of our approach that uses a randomly generated forest (see \S\ref{subsec:nopretraining}). By achieving an average of $93.58$\% on 7-Scenes after ICP, it shows that it is possible to achieve state-of-the-art performance without any prior offline training on a generic scene.


\subsection{Tracking Loss Recovery}
\label{subsec:trackinglossrecovery}

\stufig{width=.9\linewidth}{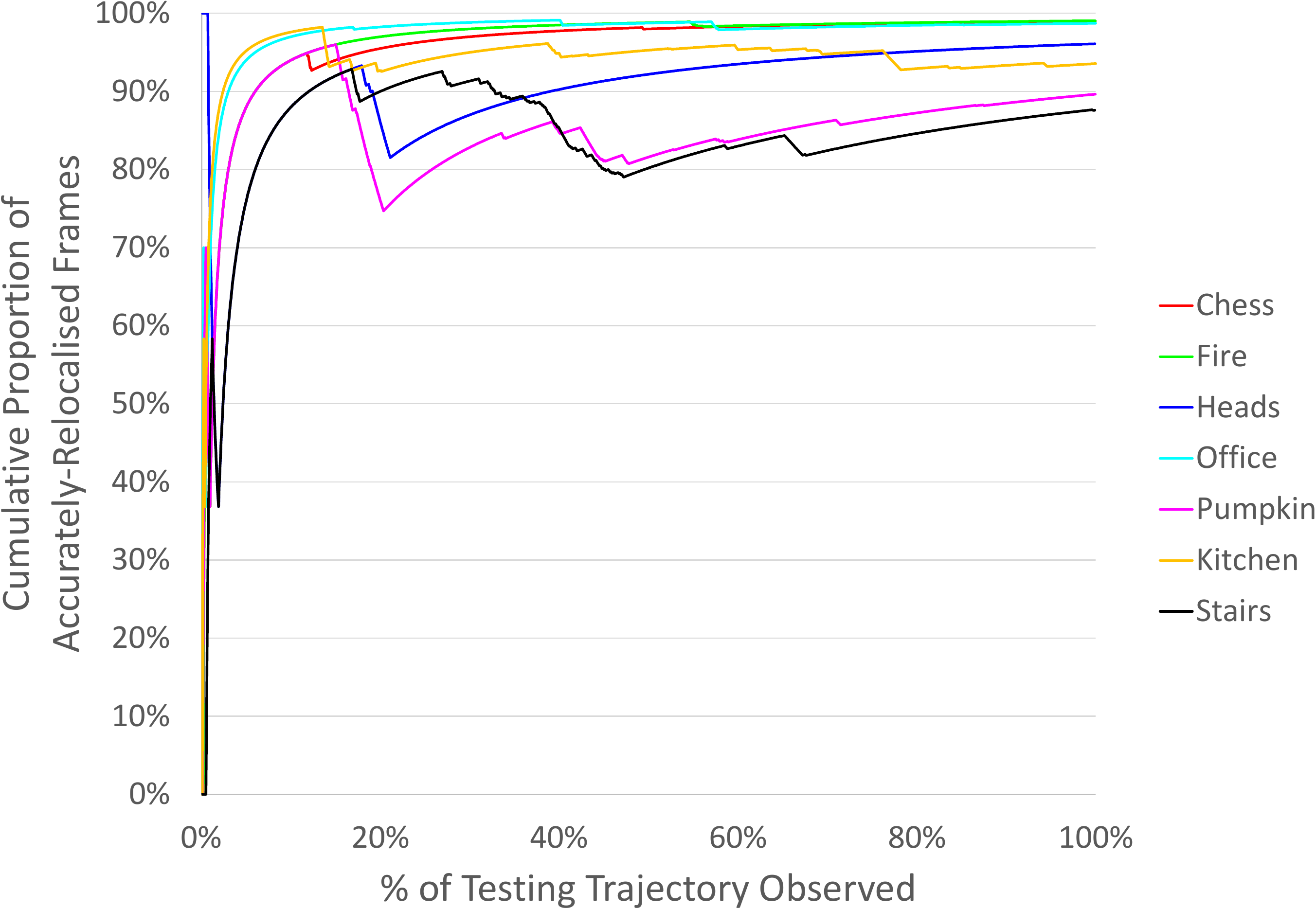}{Our approach's performance for tracking loss recovery (\S\ref{subsec:trackinglossrecovery}). Filling the leaves of a forest pre-trained on \emph{Office} frame-by-frame \emph{directly} from the \emph{testing} sequence, we are able to start relocalising almost immediately in new scenes.}{fig:trackinglossrecovery}{!t}

In \S\ref{subsec:headlineperformance}, we investigated our ability to adapt a forest to a new scene by filling its leaves with data from a training sequence for that scene, before testing the adapted forest on a separate testing sequence shot on the same scene.
Here, we quantify our approach's ability to perform this adaptation \emph{on the fly} by filling the leaves frame-by-frame from the testing sequence: this allows recovery from tracking loss in an interactive scenario without the need for prior online training on anything other than the live sequence, making our approach extremely convenient for tasks such as interactive 3D reconstruction.

Our testing procedure is as follows: at each new frame (except the first), we assume that tracking has failed, and try to relocalise using the forest we have available at that point; we record whether or not this succeeds. Regardless, we then restore the ground truth camera pose (or the tracked camera pose, in a live sequence) and, provided tracking hasn't actually failed, use examples from the current frame to continue training the forest. As Figure~\ref{fig:trackinglossrecovery} shows, we are able to start relocalising almost immediately in a live sequence (in a matter of frames, typically 4--6 are enough). Subsequent performance then varies based on the difficulty of the sequence, but rarely drops below $80\%$. This makes our approach highly practical for interactive relocalisation.

\subsection{Generalisation to Novel Poses}
\label{subsec:novelposes}

\stufig{width=\linewidth}{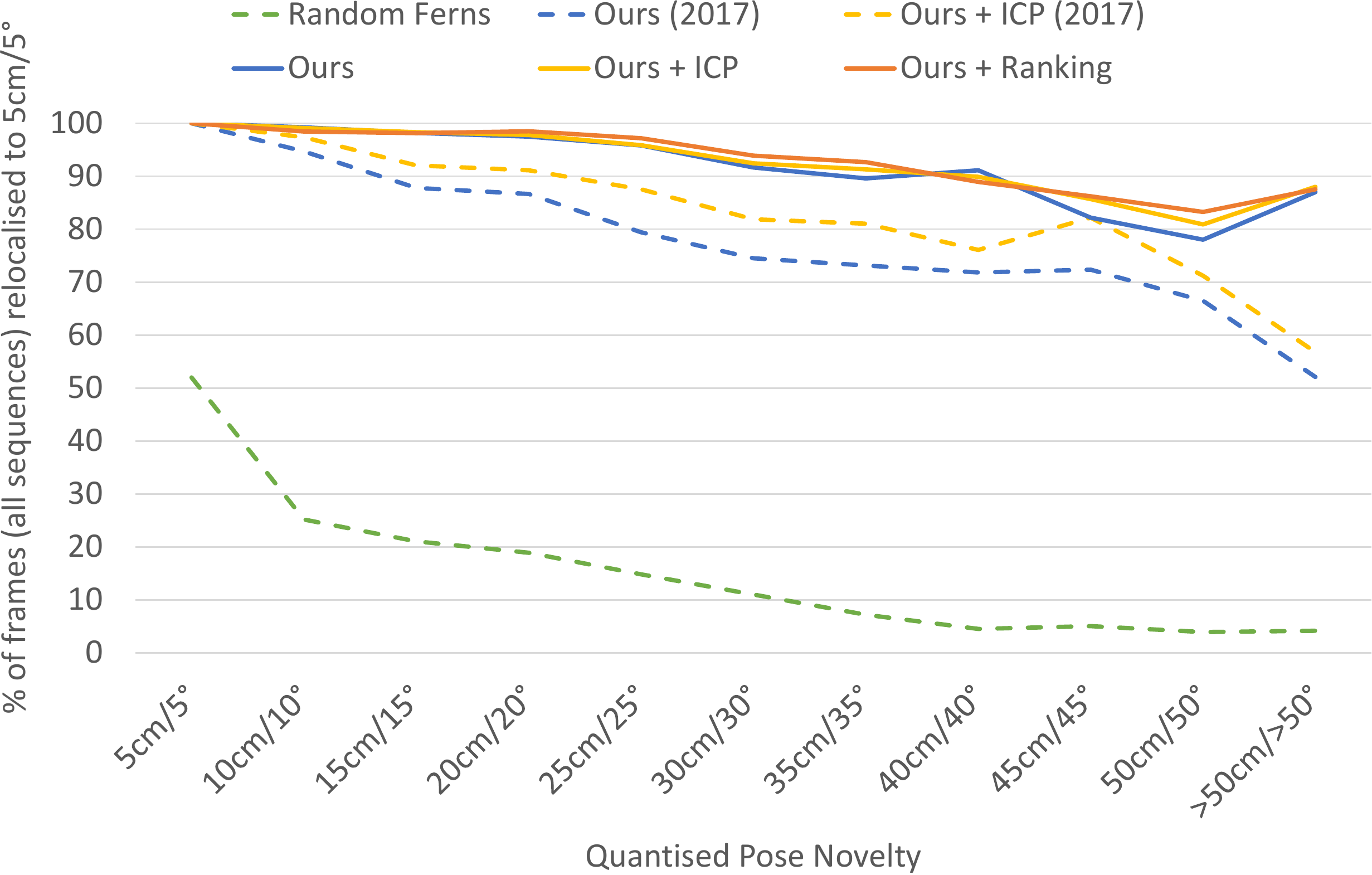}{Evaluating how well our approach generalises to novel poses in comparison to a keyframe-based random fern relocaliser based on \cite{Glocker2015}. The performance decay experienced as test poses get further from the training trajectory is much less severe with our approach than with ferns.}{fig:novelposes-graph}{!t}

\begin{figure}[!t]
	\includegraphics[width=\linewidth]{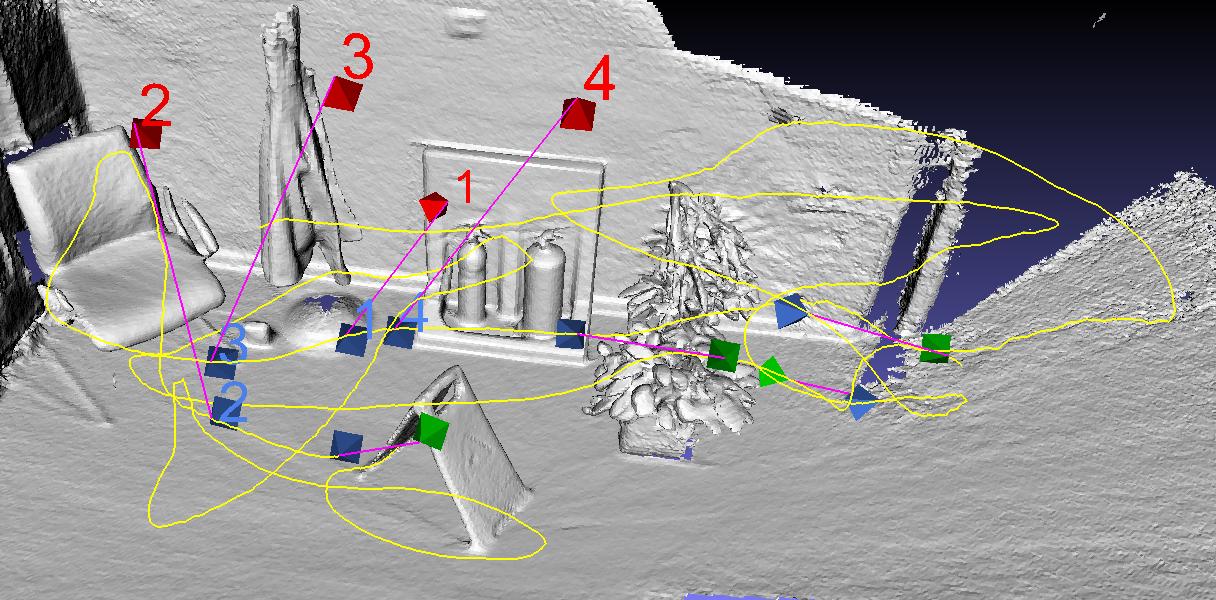}
	\includegraphics[width=\linewidth]{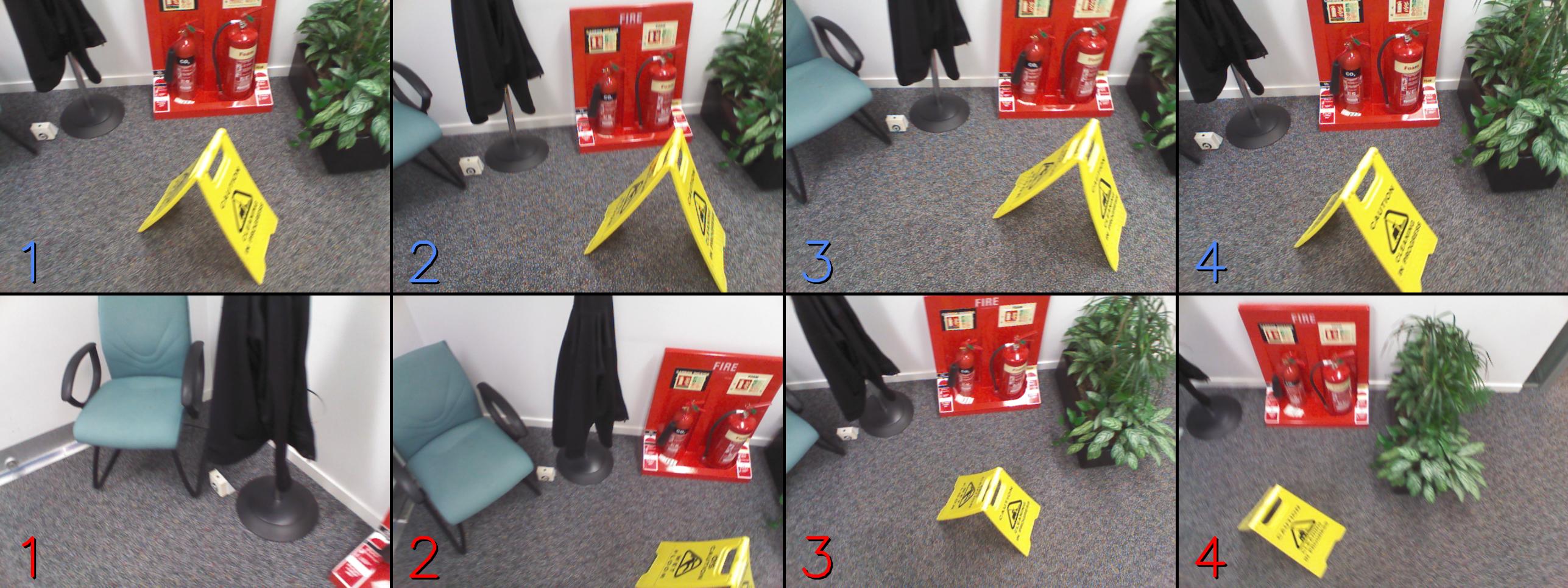}
	\caption{Novel poses from which we are able to
		relocalise to within 5cm/5$^\circ$ on the \emph{Fire} sequence from 7-Scenes
		\cite{Shotton2013}. Pose novelty measures the distance of a test pose from a
		nearby pose (blue) on the training trajectory (yellow). We can relocalise from
		both easy poses (up to 35cm/35$^\circ$ from the training trajectory, green) and
		hard poses ($>$ 35cm/35$^\circ$, red). The images below the main figure show
		views of the scene from the training poses and testing poses indicated.}
	\label{fig:novelposes-example}
	\vspace{-\baselineskip}
\end{figure}

To evaluate how well our approach generalises to novel poses, we examine how the
proportion of frames we can relocalise decreases as the distance of the (ground
truth) test poses from the training trajectory increases. We compare our
approach with the keyframe-based relocaliser in InfiniTAM \cite{Kaehler2016},
which is based on the random fern approach of Glocker et al.\
\cite{Glocker2015}, and with the original version of our own approach \cite{Cavallari2017}.
Relocalisation from novel poses is a well-known failure case
of keyframe-based methods, so we would expect the random fern approach to
perform poorly away from the training trajectory; by contrast, it is interesting
to see the extent to which both incarnations of our approach can relocalise from
a wide range of novel poses.

We perform the comparison separately for each 7-Scenes sequence, and then
aggregate the results. For each sequence, we first group the test poses into
bins by pose novelty. Each bin is specified in terms of a maximum translation
and rotation difference of a test pose with respect to the training trajectory
(for example, poses that are within 5cm and 5$^\circ$ of any training pose are
assigned to the first bin, remaining poses that are within 10cm and 10$^\circ$
are assigned to the second bin, etc.). We then determine the proportion of the
test poses in each bin for which it is possible to relocalise to within $5$cm
translational error and $5^\circ$ angular error using (a) the random fern
approach, (b) our approach (both versions) without ICP and (c) our approach (both versions)
with ICP. As shown in Figure~\ref{fig:novelposes-graph}, the decay in performance experienced
as the test poses get further from the training trajectory is much less severe with
our approach than with ferns. Moreover, our improvements to our original approach
\cite{Cavallari2017} have notably improved its ability to successfully relocalise test frames
that are more than 50cm/50$^\circ$ from the training trajectory (from less than $60$\% to more than $85$\%).

A qualitative example of our ability to relocalise from novel poses is shown in
Figure~\ref{fig:novelposes-example}. In the main figure, we show a range of test
poses from which we can relocalise in the \emph{Fire} scene, linking them to
nearby poses on the training trajectory so as to illustrate their novelty in
comparison to poses on which we have trained. The most difficult of these test
poses are also shown in the images below alongside their nearby training poses,
visually illustrating the significant differences between the two.

\subsection{Visualising the Forest's Behaviour}
\label{subsec:forestvisualisation}

Ultimately, we are able to adapt a forest to a new scene because the split functions that we preserve in the forest's branch nodes are able to route similar parts of the new scene into the same leaves, regardless of the scene on which the forest was originally trained. In effect, we exploit the observation that the forest is ultimately a way of clustering points in a scene together based on their appearance, in a way that is broadly independent of the scene on which it was pre-trained (we would prefer to cluster points based on their spatial location, but since that information is not available at test time, we rely on appearance as a proxy). Such appearance clustering is very common in the literature, e.g.\ \cite{Wang2016} uses a decision forest to perform patch matching for optical flow and stereo, and \cite{Sattler2015} uses a visual vocabulary to match features with points for relocalisation.

\begin{stusubfig}{!t}
	\begin{subfigure}{.32\linewidth}
		\centering
		\includegraphics[width=\linewidth]{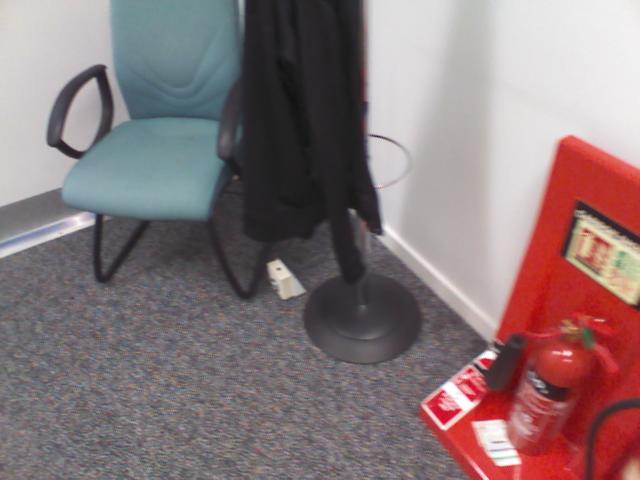}
	\end{subfigure}%
	\hspace{1mm}%
	\begin{subfigure}{.32\linewidth}
		\centering
		\includegraphics[width=\linewidth]{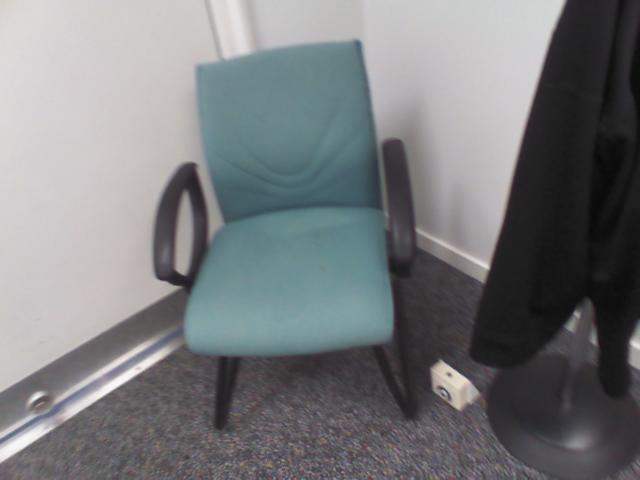}
	\end{subfigure}%
	\hspace{1mm}%
	\begin{subfigure}{.32\linewidth}
		\centering
		\includegraphics[width=\linewidth]{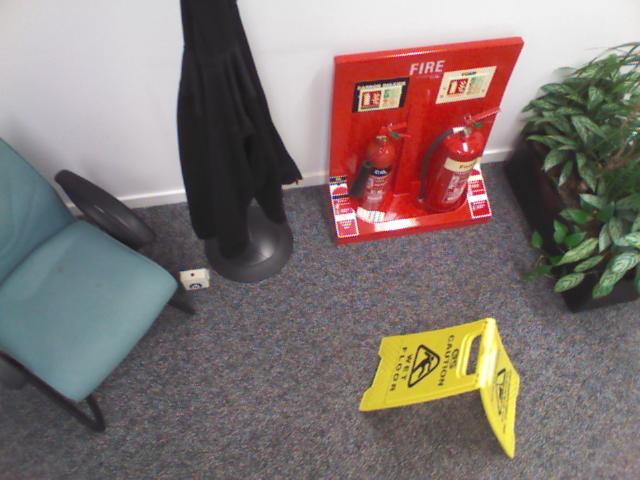}
	\end{subfigure}%
	\\[1mm]
	\begin{subfigure}{.32\linewidth}
		\centering
		\includegraphics[width=\linewidth]{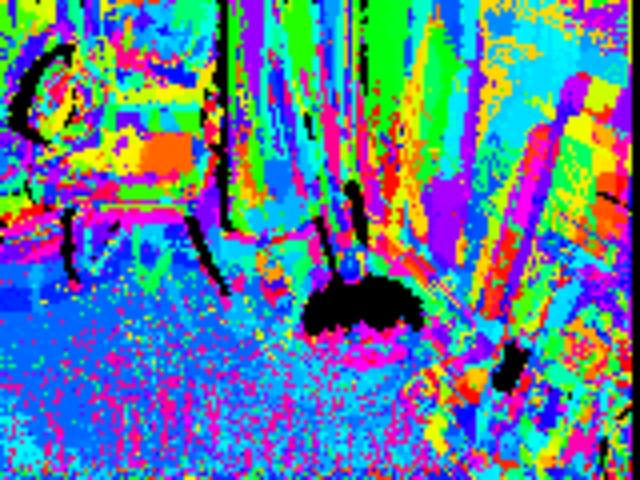}
	\end{subfigure}%
	\hspace{1mm}%
	\begin{subfigure}{.32\linewidth}
		\centering
		\includegraphics[width=\linewidth]{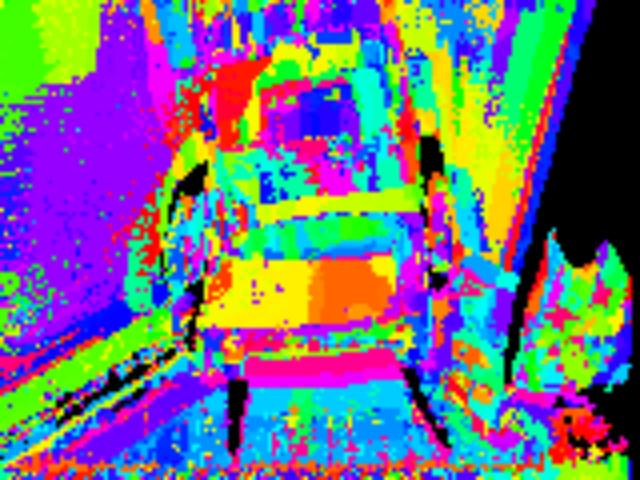}
	\end{subfigure}%
	\hspace{1mm}%
	\begin{subfigure}{.32\linewidth}
		\centering
		\includegraphics[width=\linewidth]{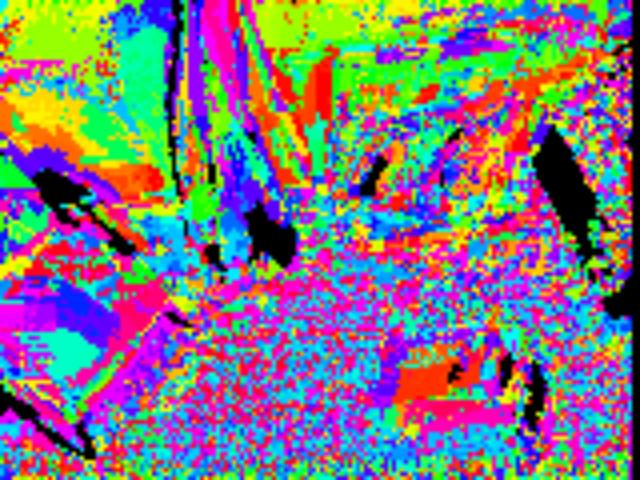}
	\end{subfigure}%
	\\[1mm]
	\begin{subfigure}{.32\linewidth}
		\centering
		\includegraphics[width=\linewidth]{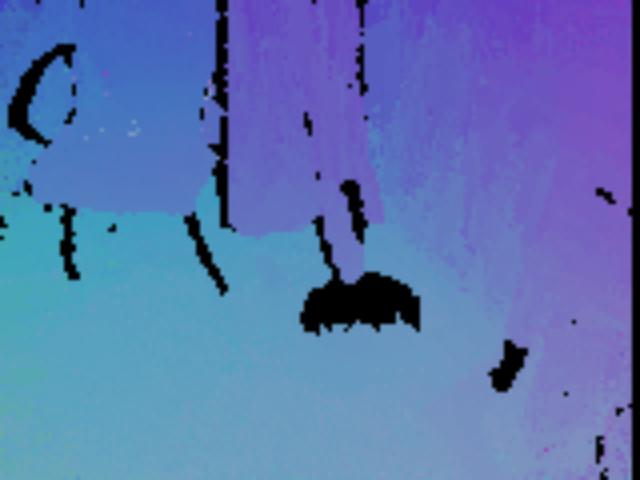}
	\end{subfigure}%
	\hspace{1mm}%
	\begin{subfigure}{.32\linewidth}
		\centering
		\includegraphics[width=\linewidth]{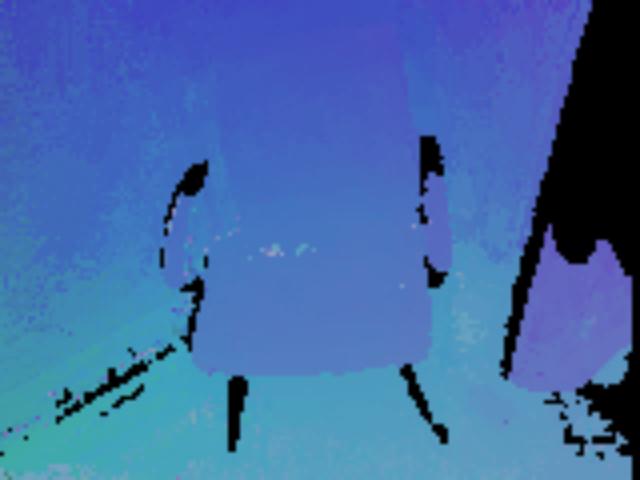}
	\end{subfigure}%
	\hspace{1mm}%
	\begin{subfigure}{.32\linewidth}
		\centering
		\includegraphics[width=\linewidth]{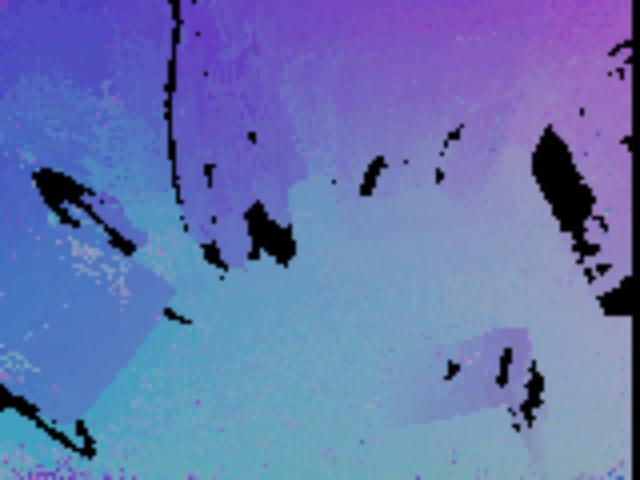}
	\end{subfigure}%
	\\[1mm]
	\begin{subfigure}{.32\linewidth}
		\centering
		\includegraphics[width=\linewidth]{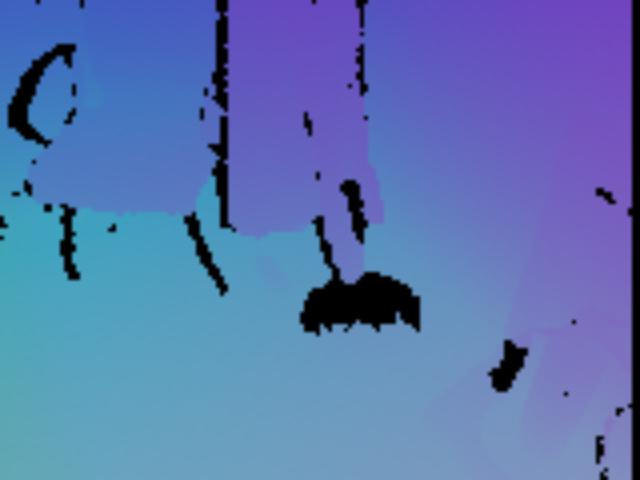}
	\end{subfigure}%
	\hspace{1mm}%
	\begin{subfigure}{.32\linewidth}
		\centering
		\includegraphics[width=\linewidth]{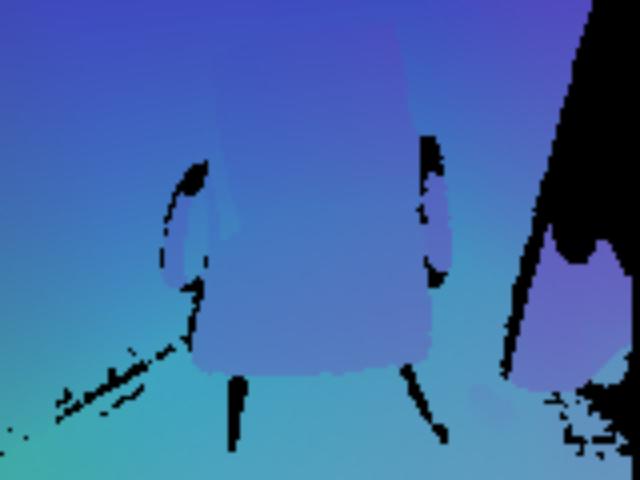}
	\end{subfigure}%
	\hspace{1mm}%
	\begin{subfigure}{.32\linewidth}
		\centering
		\includegraphics[width=\linewidth]{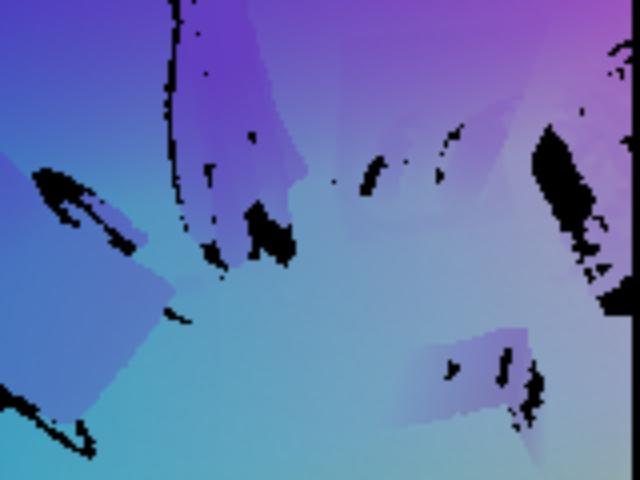}
	\end{subfigure}%
	\caption{Visualising the leaves (from the first tree) and world-space points that the forest predicts for three images of \emph{Fire} from 7-Scenes \cite{Shotton2013}. Top-to-bottom: input images, predicted leaves (randomly colourised), predicted world-space points (mapped to an RGB cube), ground truth world-space points.}
	\label{fig:forestvisualisation}
	\vspace{-1.5\baselineskip}
\end{stusubfig}

To illustrate this, we visualise both the leaves and world-space points that the forest predicts for the pixels in three images of the \emph{Fire} sequence from 7-Scenes \cite{Shotton2013} in Figure~\ref{fig:forestvisualisation}. As the pixel-to-leaf mapping (second row) shows, the forest (as expected) clusters points with similar appearances together, as can be seen from the fact that many similar-looking points that are adjacent to each other fall into the same leaves. Notably, the forest also manages, to some extent, to predict the same leaves for similar-looking points from different frames, as long as they view the scene from roughly similar viewpoints -- e.g.\ the pixels on the seat of the chair in the first two columns are mostly predicted to fall into the same leaves. Its ability to do so clearly depends on the viewpoint-invariance or otherwise of the features we use (see \S\ref{subsubsec::forestpretraining}), and indeed when the viewpoint changes more significantly, as in the third column, different leaves can be predicted. However, in practice this is not a problem: there is no need for the forest to predict the same leaves for points in the scene when viewed from different angles, only to predict leaves that represent roughly the same locations in space. In reality, many leaves can occupy the same part of space, and as long as the forest is able to predict one of them, we can still produce suitable correspondences (see third row of Figure~\ref{fig:forestvisualisation}). This makes our approach highly robust in practice, and explains why good results can be achieved even with the relatively simple features we use.

\subsection{Is Pre-Training Necessary?}
\label{subsec:nopretraining}

Since the purpose of the forest is purely to cluster scene points based on their appearance (see \S\ref{subsec:forestvisualisation}), this raises the interesting question of whether pre-training on an actual scene is really necessary in the first place. To explore this, we performed an additional experiment in which we tried replacing the pre-trained forest used for our previous experiments with ones that were entirely randomly generated. We considered forests consisting of $5$ complete binary trees of equal height. As in \S\ref{subsubsec::forestpretraining}, we generated a binary threshold function for each branch node consisting of a feature index $\phi \in [0,256)$ and a threshold $\tau \in \mathbb{R}$. In this case, instead of randomly choosing a feature index, we first randomly decided whether to use a Depth or DA-RGB feature (with a probability $p$ of choosing a depth feature), and then chose a relevant feature index randomly. We empirically fixed the threshold $\tau$ to $0$ (we also tried replicating the distribution of thresholds from a pre-trained forest, but found that it made no difference in practice: this makes sense, since the features we use are signed depth/colour differences, which we would naturally expect to be distributed around $0$). We tuned the height of the forest and the probability $p$ offline using coordinate descent (see \S\ref{subsec:tuning-single}), with $t_{\max} = 200$ms, keeping all other parameters as the defaults (see Table~\ref{tbl:parameters}). This tuning process found an optimal height of $14$ and a probability $p$ of $0.4$.

Our results on both the 7-Scenes \cite{Shotton2013} and Stanford 4 Scenes \cite{Valentin2016} benchmarks are shown as \emph{Ours (Random)} in Tables~\ref{tbl:comparativeperformance7} and \ref{tbl:comparativeperformance12}. In both cases, we achieve similar-quality results to those of our default relocaliser, at similar speeds. This indicates that pre-training on a real scene is not strictly necessary, and that the appearance-clustering role played by a pre-trained forest can be replaced by alternative approaches without compromising performance.

\section{Conclusion}
\label{sec:conclusion}

In recent years, offline approaches based on using regression to predict
2D-to-3D correspondences
\cite{Shotton2013,GuzmanRivera2014,Valentin2015RF,Brachmann2016,Massiceti2017,Meng2017arXiv}
have been shown to achieve state-of-the-art camera relocalisation results, but
their adoption for online relocalisation in practical systems such as InfiniTAM
\cite{Kaehler2015,Kaehler2016} has been hindered by the need to train
extensively on the target scene ahead of time.
In \cite{Cavallari2017}, we showed that it was possible to circumvent this limitation
by adapting offline-trained regression forests to novel scenes online.
Our adapted forests achieved relocalisation performance on 7-Scenes
\cite{Shotton2013} that was competitive with the offline-trained forests of
existing methods, and our approach ran in under $150$ms, making it competitive
for practical purposes with fast keyframe-based approaches such as random ferns
\cite{Glocker2015,Kaehler2016}. Unlike such approaches, we were also much better
able to relocalise from novel poses, removing much of the need for the user to
move the camera around when relocalising.

In this paper, we have extended this approach to achieve results that comfortably exceed
the existing state-of-the-art. In particular, our F$\stackrel{7.5\textup{cm}}{\rightarrow}$S
cascade simultaneously beats the current top performer \cite{Schmidt2017} on 7-Scenes by
around $3$\% and runs at nearly $20$ FPS. We also achieve near-perfect results on Stanford
4 Scenes \cite{Valentin2016}, beating even the existing high-performing state-of-the-art
\cite{Meng2017arXiv}. Finally, we have shown that it is possible to obtain state-of-the-art
results on both datasets without any need at all for offline pre-training on a generic scene,
whilst clarifying the role that regression forests play in scene coordinate regression pipelines.

\appendices

\section{Additional Experiments}
\label{sec:additionalexperiments}

\subsection{Timing Breakdown}
\label{subsec:timingbreakdown}

To evaluate the usefulness of our approach for on-the-fly relocalisation in new
scenes, we compare several variants of it to the keyframe-based random fern
relocaliser implemented in InfiniTAM \cite{Glocker2015,Kaehler2016}. To be practical
in a real-time system, a relocaliser needs to perform in real time during normal operation
(i.e.\ for online training whilst successfully tracking the scene), and ideally
take no more than $200$ms for relocalisation itself (when the system has lost track).

As shown in Table~\ref{tbl:timings}, the random fern relocaliser is fast both
for online training and relocalisation, taking only $1.2$ms per frame to update
the keyframe database, and $6.8$ms to relocalise when tracking is lost. However,
speed aside, the range of poses from which it is able to relocalise is quite
limited. By contrast, the variants of our approach, whilst taking longer both
for online training and for actual relocalisation, can relocalise from a much
broader range of poses, whilst still running at more than acceptable speeds.
Indeed, the \emph{Fast} variant of our approach is able to relocalise in under
$30$ms, making it competitive with the $6.8$ms taken by random ferns, whilst
dramatically outperforming it on relocalisation quality. Our cascade approaches
are slower, but achieve even better relocalisation performance, and at $< 70$ms
are still more than fast enough for practical use.

A timing breakdown for the default variant of our relocaliser, \emph{Ours (Default)},
is shown in Figure~\ref{fig:timings}. Notably, a significant proportion of the time
it takes is spent on optimising poses during the RANSAC stage of the pipeline
(a fact that is later exploited when we tune the parameters for our cascades
in \S\ref{sec:parametertuning} and \S\ref{sec:cascadedesign}). With hypothesis ranking
disabled, the second largest amount of time is spent on generating candidate hypotheses:
at least some of the time this takes is likely due to warp divergence on the GPU, since
some candidate generation threads are likely to generate acceptable hypotheses before
others. With ranking enabled, the ranking itself dominates: this is unsurprising, since
it is linear in the number of hypotheses considered.

\begin{table}[!t]
	\centering
	\scriptsize
	\begin{tabular}{lcc}
		\toprule
		& \textbf{Per-Frame Training (ms)} & \textbf{Relocalisation (ms)} \\
		\midrule
		Ours (Default) & 5.8 & 128.0 \\
		+ ICP          & 5.8 & 132.7 \\
		+ Ranking      & 5.8 & 256.8 \\
		\midrule
		Ours (Fast)  & 11.0 & 25.1 \\
		+ ICP & 11.0 & 29.9 \\
		\midrule
		Ours (Cascade F$\stackrel{7.5\textup{cm}}{\rightarrow}$S) & 11.0 & 52.4 \\
		Ours (Cascade F$\stackrel{5\textup{cm}}{\rightarrow}$I$\stackrel{7.5\textup{cm}}{\rightarrow}$S) & 11.1 & 66.1 \\
		\midrule
		Random Ferns \cite{Glocker2015,Kaehler2016} & 1.2 & 6.8 \\
		\bottomrule
	\end{tabular}
	\vspace{1mm}
	\caption{Comparing the typical timings of our approach vs.\ random ferns during
		both normal operation and relocalisation. Our approach is slower than random
		ferns, but achieves dramatically higher relocalisation performance, especially
		from novel poses. All of our experiments are run on a machine with an Intel Core
		i7-7820X CPU and an NVIDIA GeForce GTX 1080Ti GPU.}
	\label{tbl:timings}
\end{table}

\begin{stusubfig}{!t}
	\begin{subfigure}{.49\linewidth}
		\centering
		\includegraphics[height=4cm]{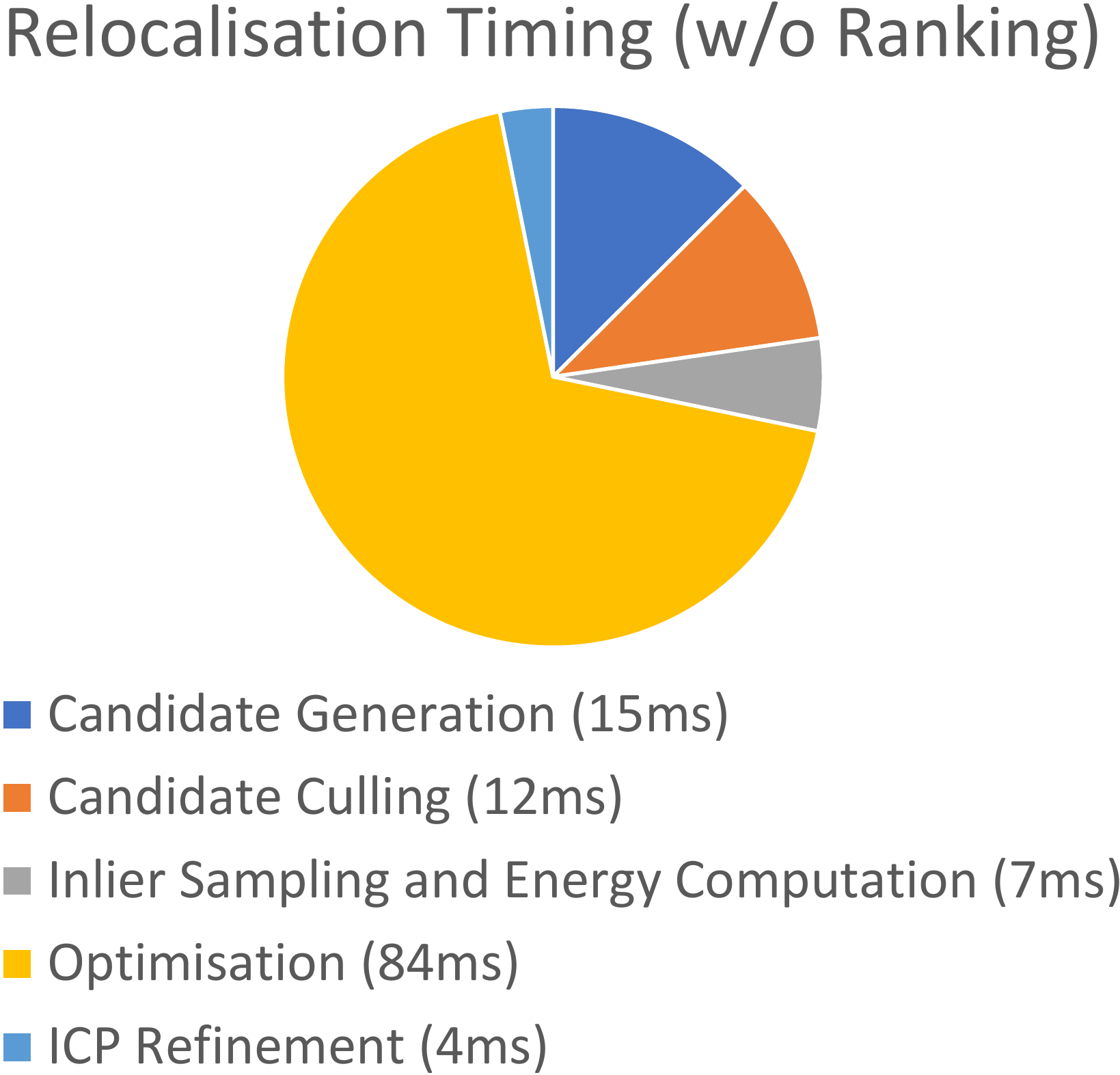}
	\end{subfigure}%
	\hspace{1mm}%
	\begin{subfigure}{.49\linewidth}
		\centering
		\includegraphics[height=4cm]{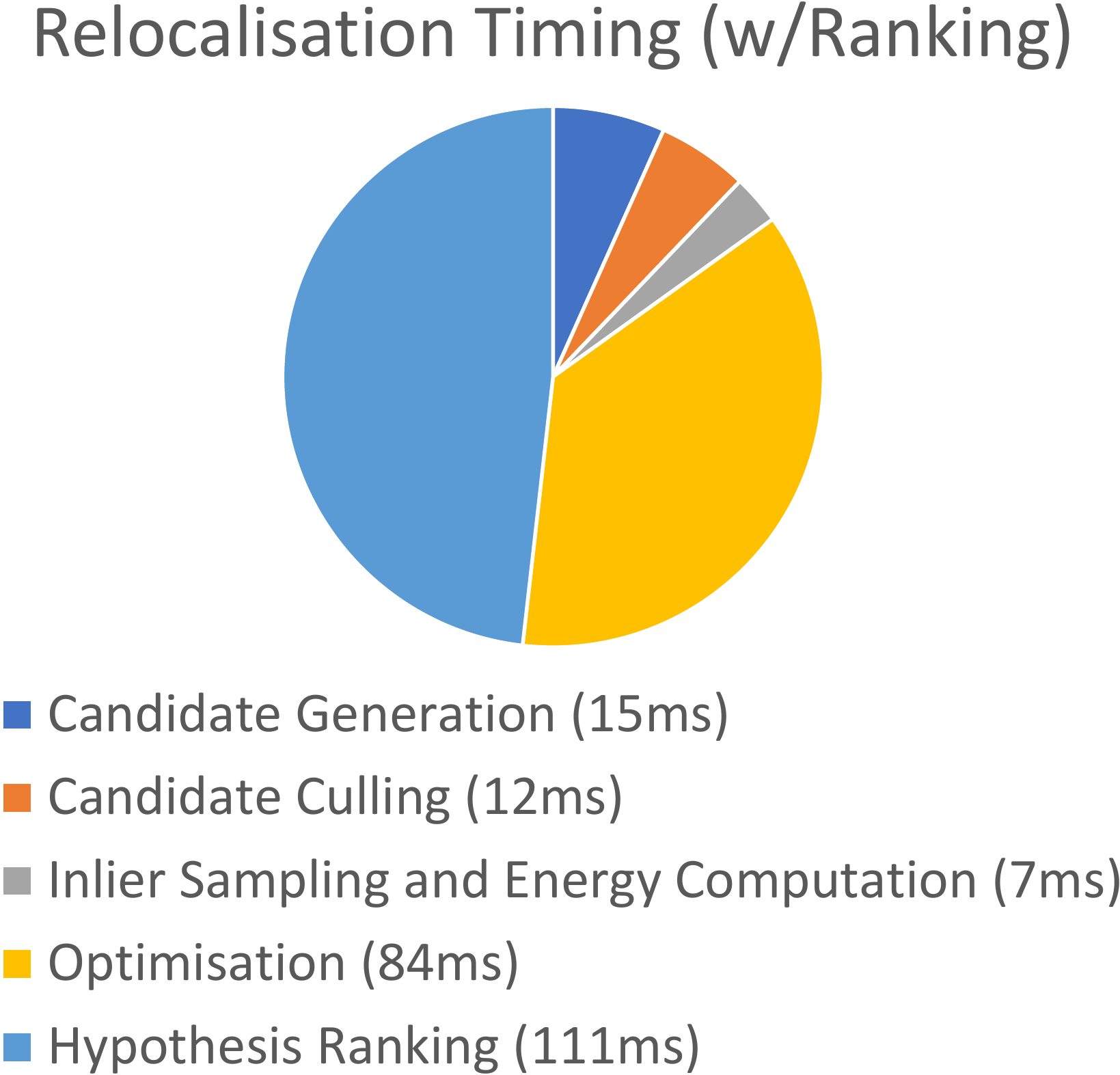}
	\end{subfigure}%
	\caption{A timing breakdown for the default variant of our relocaliser (see \S\ref{subsec:timingbreakdown}). Pose optimisation during the RANSAC stage of the pipeline is the dominant cost when hypothesis ranking is disabled. Ranking becomes the dominant cost when it is enabled, since it is linear in the number of hypotheses considered.}
	\label{fig:timings}
	\vspace{-\baselineskip}
\end{stusubfig}

\begin{table*}[!t]
	\centering
	\scriptsize
	\begin{tabular}{clcccccccc}
		\toprule
		~ & & \multicolumn{7}{c}{\textbf{Relocalisation Performance on Test Scene}} \\
		\rowcolor{white}
		\multirow{-2}{*}{\textbf{Training Scene}} & & \textbf{Chess} & \textbf{Fire} & \textbf{Heads} & \textbf{Office} & \textbf{Pumpkin} & \textbf{Kitchen} & \textbf{Stairs} & \textbf{Average (all scenes)} \\
		\midrule
		\midrule
		\cellcolor{white}                                       & Reloc     & 99.55\% & 97.00\% & 100\% & 98.85\% & 80.70\% & 95.56\% & 75.60\% & 92.47\% \\
		\cellcolor{white}                                       & + ICP     & 99.85\% & 98.75\% & 100\% & 99.13\% & 88.75\% & 91.32\% & 78.90\% & 93.81\% \\
		\cellcolor{white}\multirow{-3}{*}{Chess}                & + Ranking & 99.75\% & 99.15\% & 100\% & 99.15\% & 90.25\% & 90.02\% & 95.80\% & 96.30\% \\
		\midrule
		\cellcolor{white}                                       & Reloc     & 99.55\% & 98.20\% & 99.80\% & 98.68\% & 77.65\% & 95.14\% & 75.40\% & 92.06\% \\
		\cellcolor{white}                                       & + ICP     & 99.85\% & 99.80\% & 100\% & 98.95\% & 87.80\% & 91.36\% & 76.50\% & 93.47\% \\
		\cellcolor{white}\multirow{-3}{*}{Fire}                 & + Ranking & 99.80\% & 100\% & 100\% & 99.08\% & 90.25\% & 90.18\% & 94.40\% & 96.24\% \\
		\midrule
		\cellcolor{white}                                       & Reloc     & 99.50\% & 96.15\% & 100\% & 96.80\% & 76.95\% & 93.00\% & 53.30\% & 87.96\% \\
		\cellcolor{white}                                       & + ICP     & 99.85\% & 98.60\% & 100\% & 98.78\% & 89.05\% & 91.26\% & 59.00\% & 90.93\% \\
		\cellcolor{white}\multirow{-3}{*}{Heads}                & + Ranking & 99.80\% & 99.00\% & 100\% & 98.20\% & 90.75\% & 89.72\% & 86.80\% & 94.90\% \\
		\midrule
		\cellcolor{white}                                       & Reloc     & 99.75\% & 97.35\% & 100\% & 99.80\% & 82.25\% & 95.64\% & 79.10\% & 93.41\% \\
		\cellcolor{white}                                       & + ICP     & 99.85\% & 99.15\% & 100\% & 99.85\% & 90.00\% & 91.52\% & 80.00\% & 94.34\% \\
		\cellcolor{white}\multirow{-3}{*}{Office}               & + Ranking & 99.95\% & 99.70\% & 100\% & 99.48\% & 90.85\% & 90.68\% & 94.20\% & 96.41\% \\
		\midrule
		\cellcolor{white}                                       & Reloc     & 99.40\% & 97.00\% & 99.90\% & 99.05\% & 81.95\% & 94.62\% & 75.50\% & 92.49\% \\
		\cellcolor{white}                                       & + ICP     & 99.85\% & 98.65\% &   100\% & 99.83\% & 90.00\% & 91.50\% & 76.00\% & 93.69\% \\
		\cellcolor{white}\multirow{-3}{*}{Pumpkin}              & + Ranking & 99.85\% & 99.40\% &   100\% & 99.28\% & 91.10\% & 90.36\% & 93.50\% & 96.21\% \\
		\midrule
		\cellcolor{white}                                       & Reloc     & 99.95\% & 97.60\% & 100\% & 99.55\% & 80.65\% & 95.20\% & 76.30\% & 92.75\% \\
		\cellcolor{white}                                       & + ICP     & 99.85\% & 98.95\% & 100\% & 99.80\% & 89.25\% & 91.42\% & 78.40\% & 93.95\% \\
		\cellcolor{white}\multirow{-3}{*}{Kitchen}              & + Ranking & 99.90\% & 99.60\% & 100\% & 99.25\% & 90.20\% & 90.74\% & 94.50\% & 96.31\% \\
		\midrule
		\cellcolor{white}                                       & Reloc     & 99.55\% & 97.05\% & 99.90\% & 99.00\% & 80.15\% & 94.70\% & 82.70\% & 93.29\% \\
		\cellcolor{white}                                       & + ICP     & 99.85\% & 98.60\% &   100\% & 99.28\% & 88.55\% & 91.10\% & 84.20\% & 94.51\% \\
		\cellcolor{white}\multirow{-3}{*}{Stairs}               & + Ranking & 99.85\% & 99.15\% &   100\% & 99.23\% & 90.60\% & 89.94\% & 96.90\% & 96.52\% \\
		\midrule
		\midrule
		\cellcolor{white}                                       & Reloc     & 99.61\% & 97.19\% & 99.94\% & 98.82\% & 80.04\% & 94.84\% & 73.99\% & 92.06\% \\
		\cellcolor{white}                                       & + ICP     & 99.85\% & 98.93\% &   100\% & 99.37\% & 89.06\% & 91.35\% & 76.14\% & 93.53\% \\
		\cellcolor{white}\multirow{-3}{*}{Average}              & + Ranking & 99.84\% & 99.43\% &   100\% & 99.10\% & 90.57\% & 90.23\% & 93.73\% & 96.13\% \\
		\bottomrule
	\end{tabular}
	\vspace{1mm}
	\caption{The performance of our \emph{adaptive} approach after pre-training on various scenes
		of the 7-Scenes dataset \cite{Shotton2013}. We show the scene used to pre-train the forest in each version of our approach in the left column. The pre-trained forests are adapted \emph{online} for the test scene, as described in the main text. Note that `+ ICP' and `+ Ranking' are cumulative, i.e.\ the third set of results for each training scene refer to a version of our approach in which both ICP and model-based hypothesis ranking are used.
		The percentages denote proportions of test frames with $\le 5$cm translational error and $\le 5^\circ$ angular error.
	}
	\label{tbl:adaptationperformance}
	\vspace{-\baselineskip}
\end{table*}

\begin{table}[!t]
	\centering
	\scriptsize
	\begin{tabular}{lccc}
		\toprule
		& \multicolumn{3}{c}{\textbf{Average Frame Time (ms)}} \\
		& Successful & Failed & All \\
		\midrule
		Default             & 125.0 & 107.9 & 124.0 \\
		Default (w/ICP)     & 129.8 & 114.0 & 128.9 \\
		Default (w/Ranking) & 254.5 & 255.9 & 254.6 \\
		\midrule
		Fast         & 24.9 & 24.8 & 24.9 \\
		Fast (w/ICP) & 29.8 & 30.1 & 30.0 \\
		\midrule
		Intermediate         & 74.6 & 71.7 & 73.9 \\
		Intermediate (w/ICP) & 79.4 & 76.7 & 78.8 \\
		\midrule
		Slow             & 77.3 & 73.5 & 76.4 \\
		Slow (w/Ranking) & 202.5 & 202.3 & 202.4 \\
		\midrule
		Cascade F$\stackrel{5\textup{cm}}{\rightarrow}$I                                          & 51.0 & 104.5 & 54.8 \\
		Cascade F$\stackrel{5\textup{cm}}{\rightarrow}$S                                          & 73.0 & 202.8 & 81.9 \\
		Cascade F$\stackrel{7.5\textup{cm}}{\rightarrow}$S                                        & 44.6 & 148.3 & 51.4 \\
		Cascade F$\stackrel{5\textup{cm}}{\rightarrow}$I$\stackrel{7.5\textup{cm}}{\rightarrow}$S & 61.0 & 180.4 & 69.3 \\
		\bottomrule
	\end{tabular}
	\caption{The average times taken to relocalise successful/failed/all frames from the 7-Scenes dataset \cite{Shotton2013}. `Successful' frames are defined as those frames whose relocalised poses are within $5$cm/$5^\circ$ of the ground truth. Note that unlike the average numbers elsewhere in the paper, which as per common practice were computed by averaging the averages for the different sequences in the dataset, these averages were computed by averaging over the individual frames (this is equivalent to weighting the average of averages by the number of frames in each sequence).}
	\label{tbl:successfulfailedtimings}
\end{table}

\subsection{Successful/Failed Frame Timings}

To better understand how the time taken by our approach to try to relocalise a frame varies depending on whether or not that frame can be successfully relocalised, we timed several different variants of our relocaliser on 7-Scenes \cite{Shotton2013}, and compared the average times taken for just the successful/failed frames to the timing results for all frames (see Table~\ref{tbl:successfulfailedtimings}). The results indicate that for the non-cascade variants, there is little difference between the times taken for successful/failed frames, which is what we would expect. By contrast, for the cascade variants, the average time taken for failed frames is significantly higher than that for successful ones: this makes sense, since the way in which our cascades work is to try the relocalisers in order, and for failed frames, we end up running the full cascade. For successful frames, we are often able to avoid running the slower relocalisers towards the end of the cascade, as indicated by the fact that the average times for the successful frames with each cascade are much lower than the corresponding average times for the slower relocalisers in the cascade. Moreover, it is notable that the average times for all frames are quite close to those for successful frames, indicating that in practice, most frames are successfully relocalised.

\begin{stusubfig*}{!t}
	\begin{subfigure}{.18\linewidth}
		\centering
		\includegraphics[height=2.5cm]{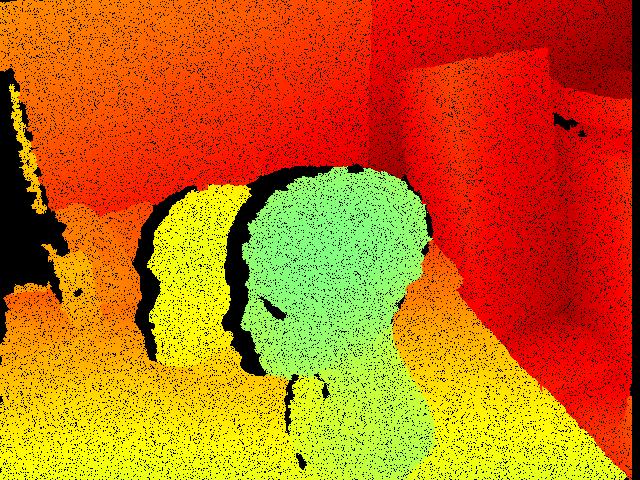}
		\caption{10\%}
	\end{subfigure}%
	\hspace{4mm}%
	\begin{subfigure}{.18\linewidth}
		\centering
		\includegraphics[height=2.5cm]{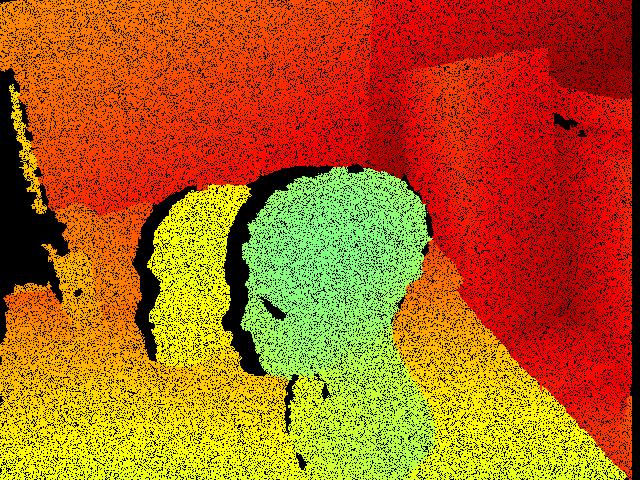}
		\caption{20\%}
	\end{subfigure}%
	\hspace{4mm}%
	\begin{subfigure}{.18\linewidth}
		\centering
		\includegraphics[height=2.5cm]{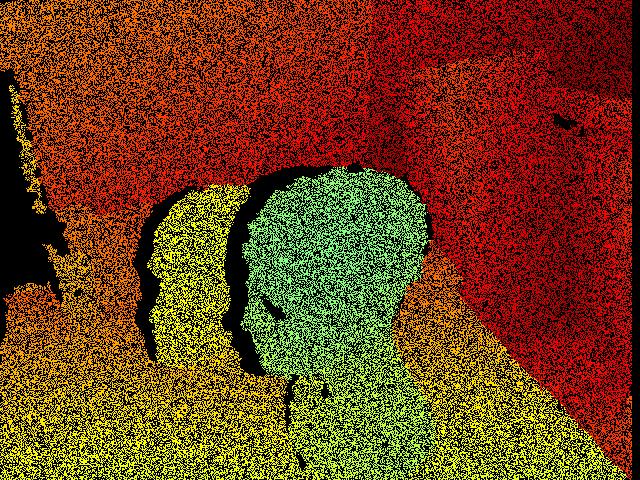}
		\caption{50\%}
	\end{subfigure}%
	\hspace{4mm}%
	\begin{subfigure}{.18\linewidth}
		\centering
		\includegraphics[height=2.5cm]{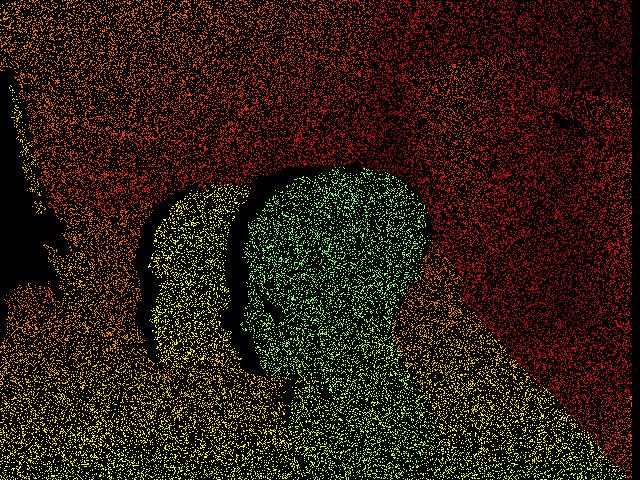}
		\caption{80\%}
	\end{subfigure}%
	\hspace{4mm}%
	\begin{subfigure}{.18\linewidth}
		\centering
		\includegraphics[height=2.5cm]{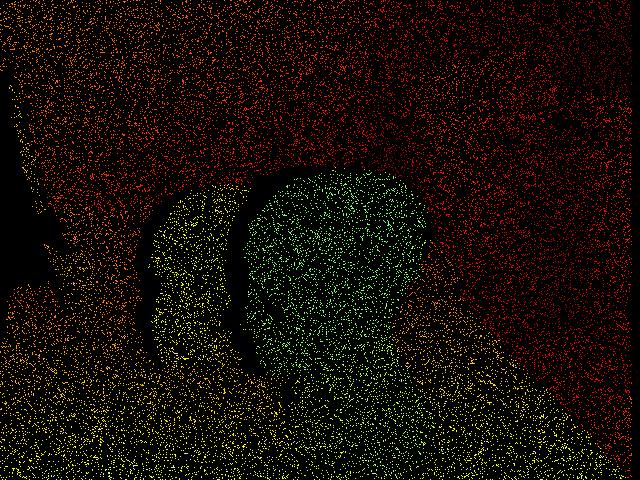}
		\caption{90\%}
	\end{subfigure}%
	\caption{Examples of the different levels of missing depth used in our experiment (see \S\ref{subsubsec:missingdepth}).}
	\label{fig:missingdepth-examples}
	\vspace{-\baselineskip}
\end{stusubfig*}

\begin{stusubfig*}{!t}
	\begin{subfigure}{.18\linewidth}
		\centering
		\includegraphics[height=2.5cm]{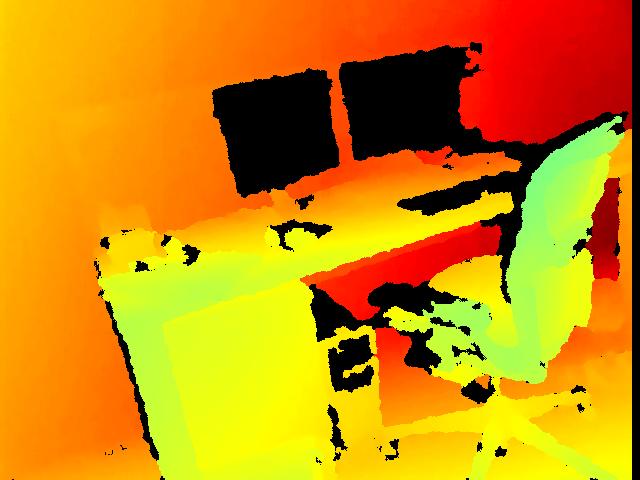}
		\\[2mm]
		\includegraphics[height=2.5cm]{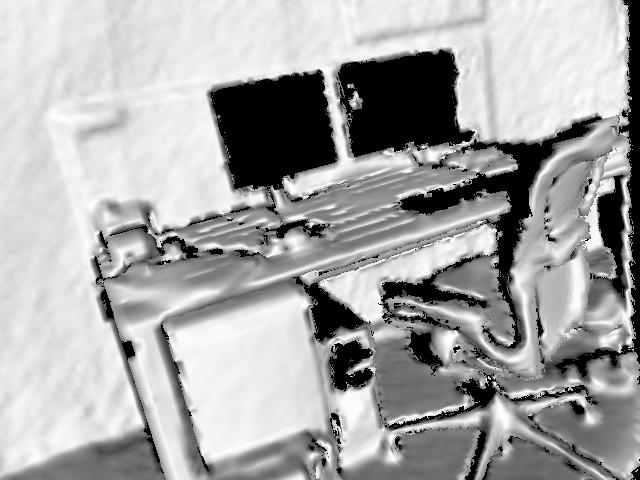}
		\caption{$\sigma = 0$}
	\end{subfigure}%
	\hspace{4mm}%
	\begin{subfigure}{.18\linewidth}
		\centering
		\includegraphics[height=2.5cm]{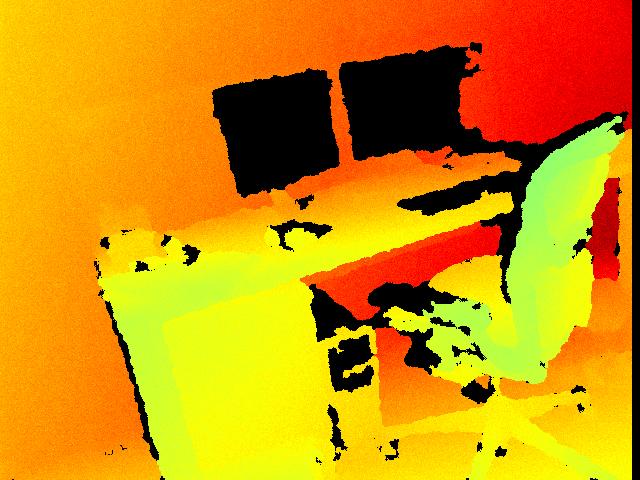}
		\\[2mm]
		\includegraphics[height=2.5cm]{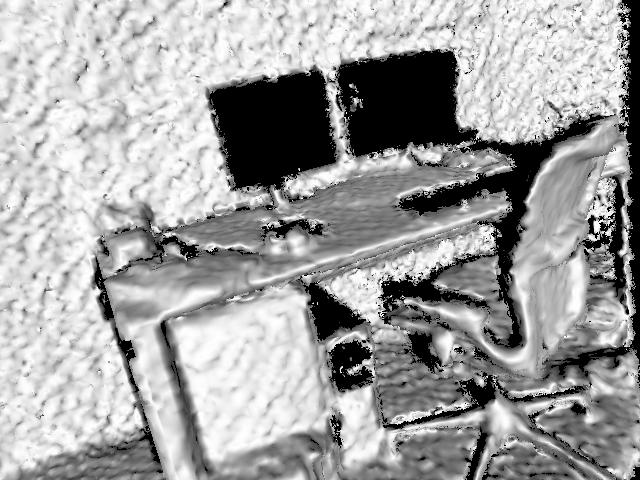}
		\caption{$\sigma = 0.025$}
	\end{subfigure}%
	\hspace{4mm}%
	\begin{subfigure}{.18\linewidth}
		\centering
		\includegraphics[height=2.5cm]{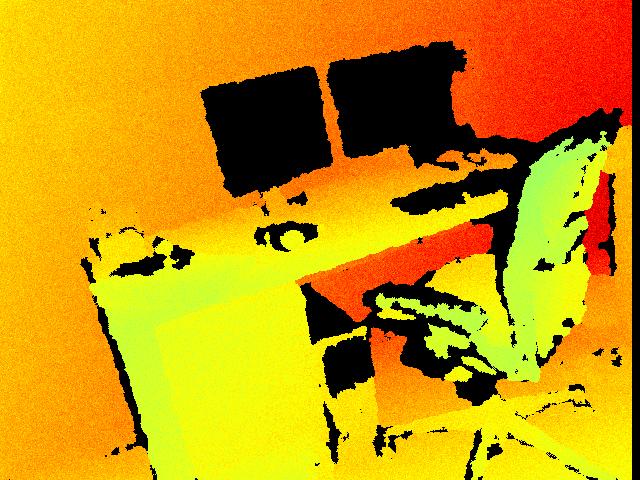}
		\\[2mm]
		\includegraphics[height=2.5cm]{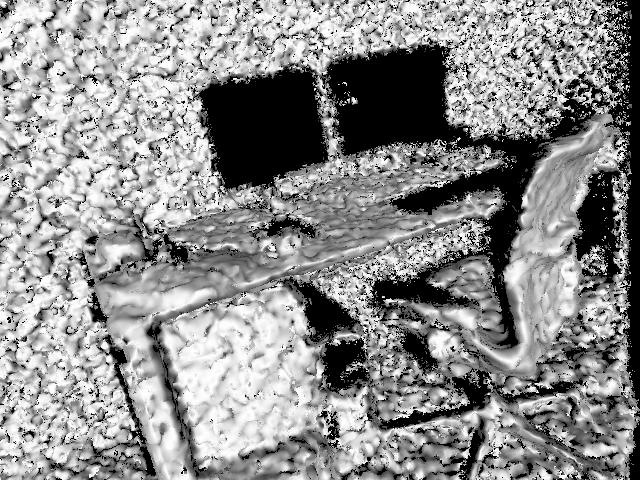}
		\caption{$\sigma = 0.05$}
	\end{subfigure}%
	\hspace{4mm}%
	\begin{subfigure}{.18\linewidth}
		\centering
		\includegraphics[height=2.5cm]{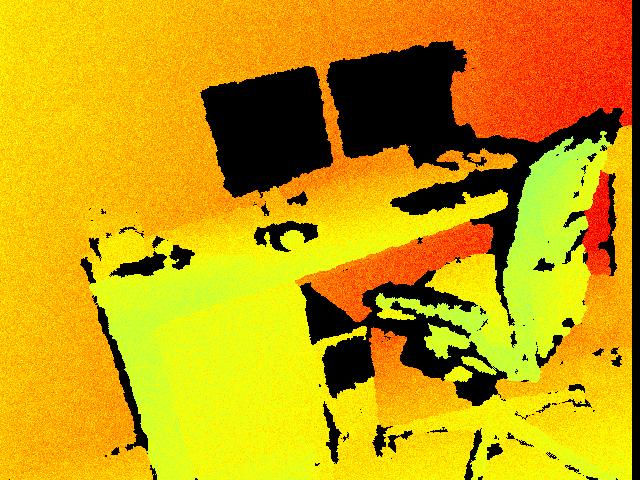}
		\\[2mm]
		\includegraphics[height=2.5cm]{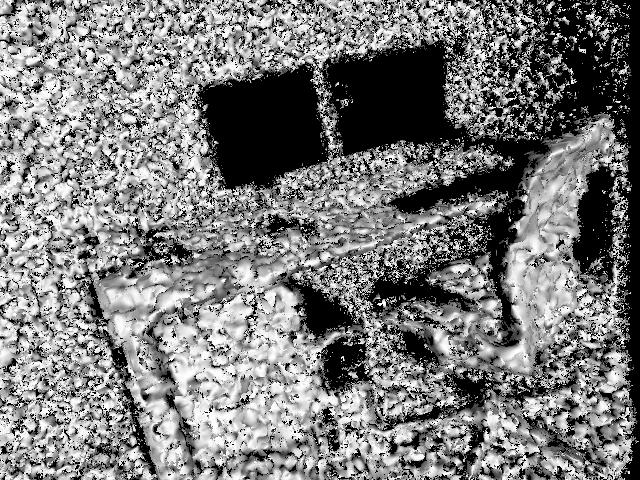}
		\caption{$\sigma = 0.075$}
	\end{subfigure}%
	\hspace{4mm}%
	\begin{subfigure}{.18\linewidth}
		\centering
		\includegraphics[height=2.5cm]{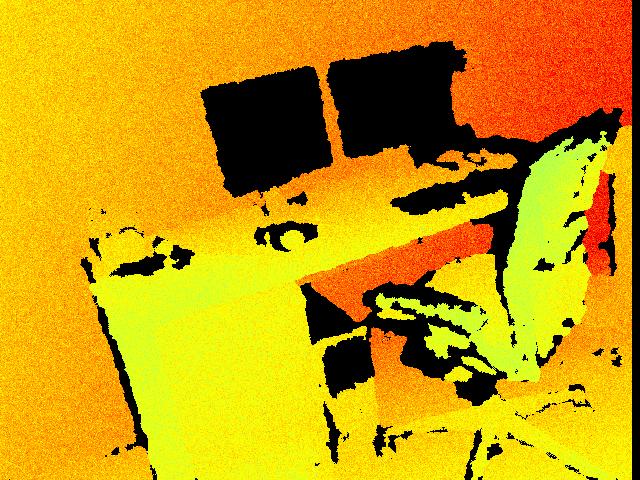}
		\\[2mm]
		\includegraphics[height=2.5cm]{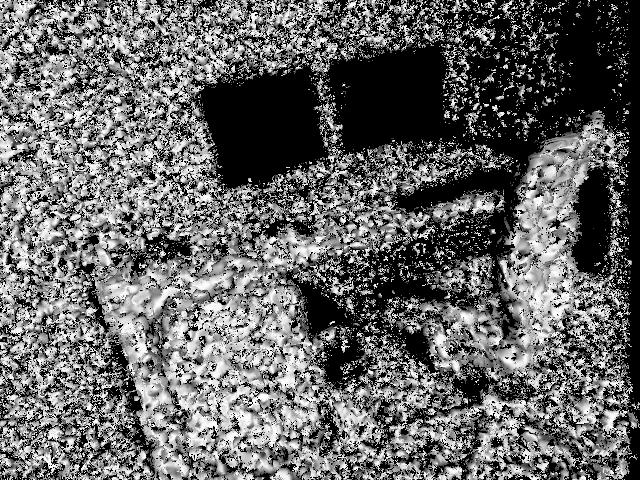}
		\caption{$\sigma = 0.1$}
	\end{subfigure}%
	\caption{Examples of the different levels of noisy depth used in our experiment (see \S\ref{subsubsec:noisydepth}), and their effects on the 3D reconstructions fused by InfiniTAM (using a voxel size of $2$cm). The depth images with added noise are shown on the top row. Rendered images of the 3D models fused based on these levels of noise are shown on the bottom row.}
	\label{fig:noisydepth-examples}
	\vspace{-\baselineskip}
\end{stusubfig*}

\subsection{Adaptation Performance}
\label{subsec:adaptationperformance}

In \S\ref{subsec:headlineperformance}, we evaluated how the performance of forests that had been pre-trained offline on the \emph{Office} sequence from 7-Scenes \cite{Shotton2013} and then adapted to the target scene compared to that of offline methods that had been trained offline directly on the target scene. Here, we show that very similar results can be obtained by pre-training on \emph{any} of the sequences from 7-Scenes, thus demonstrating that there is nothing specific to \emph{Office} that makes it particularly suitable for pre-training (this is to be expected, since we show in \S\ref{subsec:nopretraining} that similar results can be obtained without pre-training offline at all).

As mentioned in \S\ref{subsec:headlineperformance}, we use the following testing procedure.
First, we pre-train a forest on a generic scene \emph{offline} and remove the contents of its leaves.
Next, we adapt the forest by feeding it new examples from a training sequence captured on the scene of interest: this runs \emph{online} at frame rate.
Finally, we test the adapted forest by using it to relocalise from every frame of a separate testing sequence captured on the scene of interest.

As shown in Table~\ref{tbl:adaptationperformance}, in all cases the results are very accurate.
Whilst there are certainly some variations in the performance achieved by
adapted forests pre-trained on different scenes (in particular, the forest trained
on the \emph{Heads} sequence from the dataset, which is very simple, is slightly
worse), the differences are not profound: in particular, relocalisation
performance seems to be more tightly coupled to the difficulty of the scene of
interest than to the scene on which the forest was pre-trained. Notably, all of
our adapted forests achieve results that are comparable (and in many cases superior)
to those of state-of-the-art \emph{offline} methods (see Table~\ref{tbl:comparativeperformance7}).

\subsection{Robustness to Missing/Noisy Depth}

To evaluate our approach's robustness to missing/noisy depth, we performed two sets of experiments, one in which we randomly masked out various percentages of the depth images, and another in which we corrupted all of the depth values in the images with zero-mean, depth-dependent Gaussian noise, with various standard deviations.\footnote{For both experiments, we disabled the built-in SVM in InfiniTAM that measures tracker reliability (see \S\ref{subsubsec::forestadaptation}), in order to better isolate how our relocaliser performs for missing/noisy depth. In this case, all of the poses are ground truth poses from 7-Scenes \cite{Shotton2013} that are already assumed to be sufficiently reliable.} Our expectation was that our relocaliser would be relatively robust to missing depth, since only a small number of reliable correspondences are needed to accurately estimate the pose, but that it might be more sensitive to noisy depth, since we rely on having reasonably accurate world space points in the leaves of the forest in order to correctly relocalise.

\subsubsection{Missing Depth}
\label{subsubsec:missingdepth}

For the missing depth experiment, we evaluated how the performance of our \emph{Default} relocaliser on 7-Scenes \cite{Shotton2013} varied for different levels of missing depth, ranging from 10\% up to 95\%. To mask out a percentage $p \in [0,1]$ of the pixels in a depth image, we uniformly sampled a real number $r_i \sim [0,1]$ for each pixel in the image, and set the pixel to zero iff $r_i \le p$. Examples of the different levels of missing depth involved can be seen in Figure~\ref{fig:missingdepth-examples}.

The results are shown in Figure~\ref{fig:missingdepth-results}. As expected, they demonstrate that our relocaliser is relatively robust to missing depth: the average pre-ICP relocalisation performance remains above $85\%$ even with $70\%$ of the depth values missing, and the average post-ICP performance is even more robust, remaining over $85\%$ even with $90\%$ of the depth values missing (this is to be expected, since even if the pre-ICP relocaliser itself performs a bit worse, it only has to return an initial pose that falls within the ICP convergence basin to allow ICP to succeed). The performance does start to decrease more significantly when $95$\% of the depth values are missing: this is most likely because by that stage, there are far fewer remaining points, and it becomes harder to find the reliable correspondences needed. Nevertheless, even at that stage, our post-ICP performance remains over $80\%$.

\subsubsection{Noisy Depth}
\label{subsubsec:noisydepth}

\stufig{width=\linewidth}{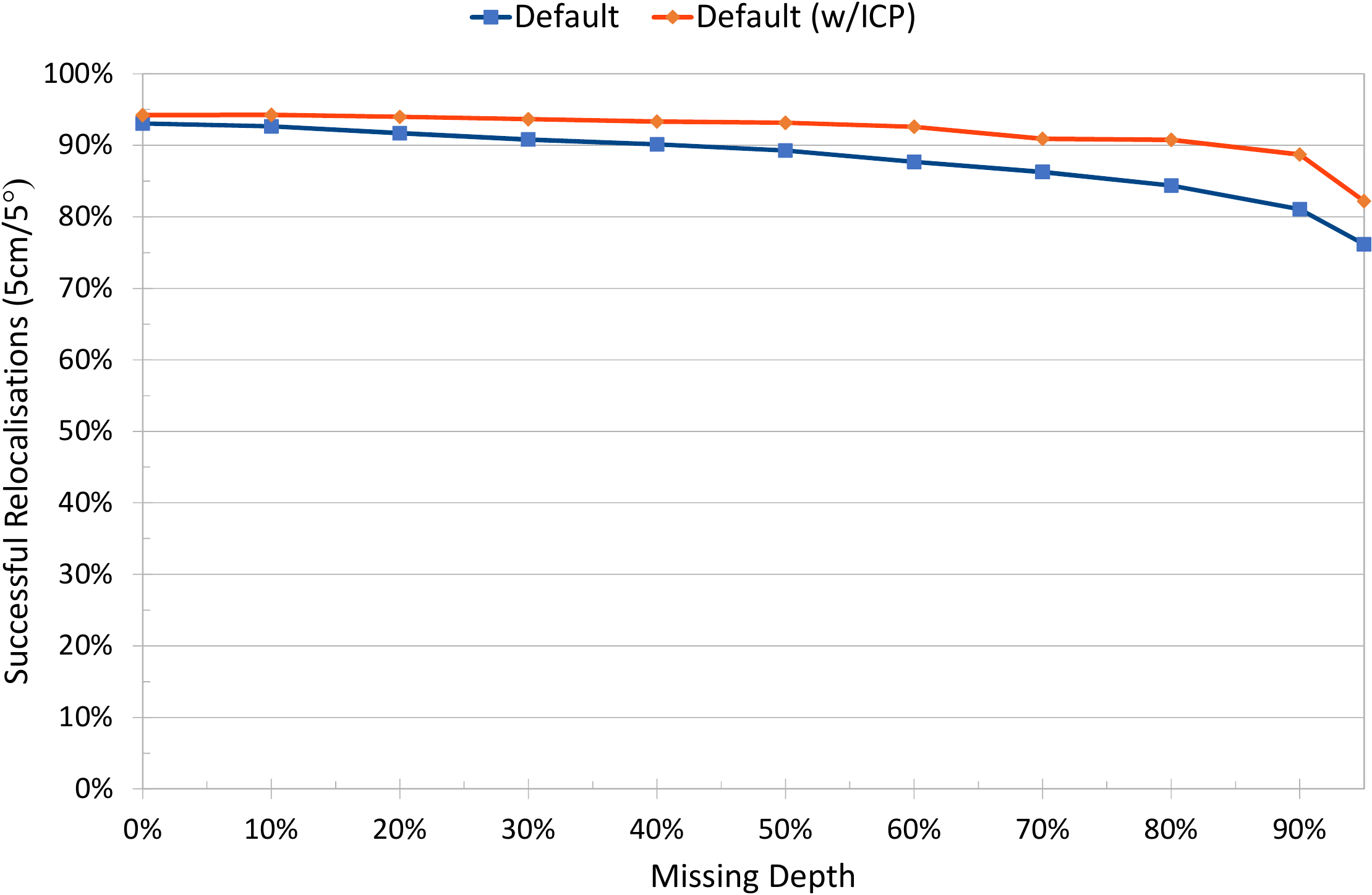}{The performance of our \emph{Default} relocaliser on 7-Scenes \cite{Shotton2013} for different levels of missing depth (see \S\ref{subsubsec:missingdepth}).}{fig:missingdepth-results}{!t}

\stufig{width=\linewidth}{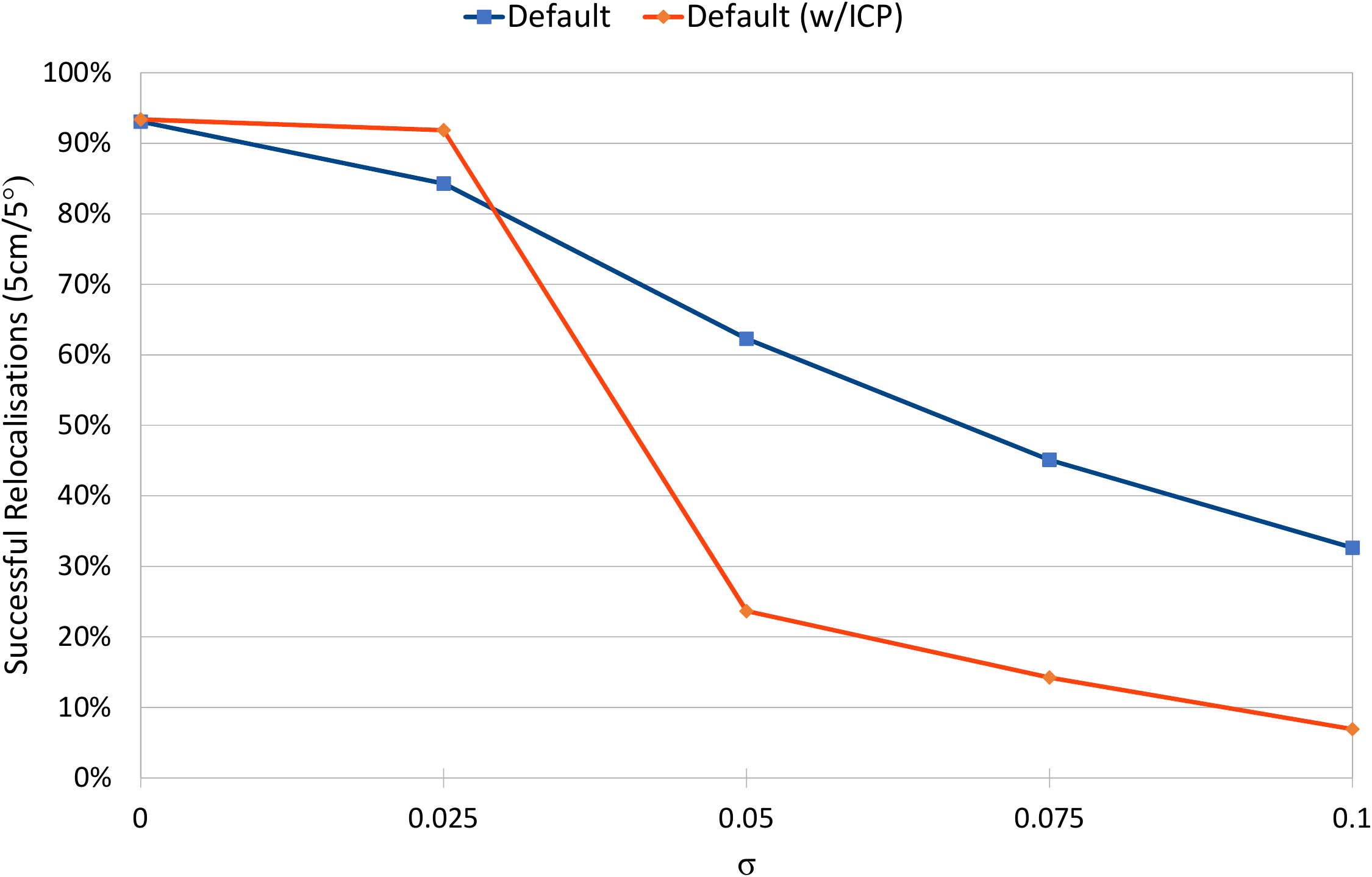}{The performance of our \emph{Default} relocaliser on 7-Scenes \cite{Shotton2013} for different levels of zero-mean, depth-dependent Gaussian noise (see \S\ref{subsubsec:noisydepth}).}{fig:noisydepth-results}{!t}

For the noisy depth experiment, we evaluated how the performance of our \emph{Default} relocaliser on 7-Scenes \cite{Shotton2013} varied when we added different levels of zero-mean, depth-dependent Gaussian noise to the depth images. We considered Gaussians with several different $\sigma$ values, as shown in Figure~\ref{fig:noisydepth-examples}. To add depth-dependent noise to a depth image for a given $\sigma$ value, we uniformly sampled a value $n_i \sim \mathcal{N}(0,\sigma^2)$ for each pixel in the image, and then replaced the pixel's depth value $d_i$ with $d_i + n_i \times d_i$. Examples of the effect this has for different $\sigma$ values can be seen in Figure~\ref{fig:noisydepth-examples}.

As expected, the results in Figure~\ref{fig:noisydepth-results} indicate that our method is rather more sensitive to noisy depth than it was to missing depth (see \S\ref{subsubsec:missingdepth}). In particular, it seems reasonably tolerant of a small amount of depth noise ($\sigma = 0.025$, i.e.\ a standard deviation of $2.5$cm at $1$m), but its performance degrades much more significantly for larger amounts of noise. This makes sense, since our method needs to find reasonably accurate correspondences between points in camera space and world space in order to relocalise, and thus the world space points we add to the leaves of the forest at adaptation time need to be reasonably accurate. If the depth is too noisy, these points are likely to be inaccurate, leading to much worse relocalisation performance.

Notably, our relocaliser's post-ICP performance drops sharply as $\sigma$ increases, whereas its pre-ICP performance degrades much more gracefully. There are two main reasons for this: (i) ICP is much more sensitive to outlying points than the RANSAC stage of our pipeline (which, like all RANSAC-based approaches, explicitly aims to exclude outliers from consideration), and (ii) as $\sigma$ increases, it becomes increasingly difficult for InfiniTAM to fuse a high-quality 3D model to which ICP can register the current depth image (see Figure~\ref{fig:noisydepth-examples}). To mitigate this latter problem, we used a larger voxel size of $2$cm for this experiment (without which, InfiniTAM is unable to fuse a reasonable model at all for high levels of depth noise), but even with this change, the model quality notably decreases for high values of $\sigma$. Based on these results, we thus recommend disabling ICP when the depth is anticipated to be particularly noisy.

\subsection{Usefulness of DA-RGB Features}
\label{subsec:rgbfeatures}

\begin{table}[!t]
	\centering
	\scriptsize
	\begin{tabular}{cccc}
		\toprule
		& \textbf{DA-RGB + Depth} & \textbf{DA-RGB Only} & \textbf{Depth Only} \\
		\midrule
		Chess   & 99.85\% & 99.65\% & 99.75\% \\
		Fire    & 98.50\% & 99.00\% & 97.15\% \\
		Heads   & 100\%   & 100\%   & 99.90\% \\
		Office  & 99.10\% & 98.95\% & 97.80\% \\
		Pumpkin & 89.50\% & 90.20\% & 80.90\% \\
		Kitchen & 90.32\% & 89.50\% & 85.26\% \\
		Stairs  & 77.80\% & 68.90\% & 68.70\% \\
		\midrule
		\textbf{Average} & 93.58\% & 92.31\% & 89.92\% \\
		\textbf{Avg.\ Median Error} & 0.013m/1.16$^\circ$ & 0.013m/1.17$^\circ$ & 0.013m/1.17$^\circ$ \\
		\bottomrule
	\end{tabular}
	\caption{Comparing the post-ICP performance on 7-Scenes \cite{Shotton2013} of three variants of our relocaliser based on randomly-generated forests (see \S\ref{subsec:nopretraining}) with different sets of features. `DA-RGB + Depth' is the same as `Ours (Random)', i.e.\ a randomly-generated forest that uses both Depth-Adaptive RGB features and Depth features. `DA-RGB Only' is a randomly-generated forest that uses only Depth-Adaptive RGB features. `Depth Only' is a randomly-generated forest that uses only Depth features.}
	\label{tbl:rgbfeatures}
\end{table}

To evaluate the usefulness of the Depth-Adaptive RGB (`DA-RGB') features we describe in \S\ref{subsubsec::forestpretraining}, we compared the post-ICP performance on 7-Scenes \cite{Shotton2013} of three variants of our relocaliser that are based on randomly-generated forests (see \S\ref{subsec:nopretraining}) with different sets of features. Specifically, we randomly generated three forests -- one (also shown as `Ours (Random)' in Tables~\ref{tbl:comparativeperformance7} and \ref{tbl:comparativeperformance12}) based on feature vectors containining $128$ DA-RGB features and $128$ Depth features, another based on feature vectors containing $256$ DA-RGB features, and a final one based on feature vectors containing $256$ Depth features. For the first forest, we randomly chose to split each branch node based on a Depth rather than a DA-RGB feature with probability $p = 0.4$, as per \S\ref{subsec:nopretraining}; for the other forests, $p$ was irrelevant, since all of the features were of one type or the other. We used a tree height of $14$ in each case.

The results are shown in Table~\ref{tbl:rgbfeatures}. For most sequences, we found that using DA-RGB features was superior to using Depth features alone. Moreover, we found that a combination of both DA-RGB and Depth features performed best overall, particularly on hard sequences like \emph{Stairs}. Based on these results, we recommend using a combination of both types of feature, rather than either of the two alone.

\subsection{Outdoor Relocalisation}
\label{subsec:outdoorrelocalisation}

\providecommand{\best}[1]{\textbf{\textcolor{red}{#1}}}
\providecommand{\secondbest}[1]{\textcolor{blue}{#1}}

\begin{table*}[!t]
	\centering
	\scriptsize
	\begin{tabular}{lllllll}
		\toprule
		& \textbf{Kings College}       & \textbf{Street} & \textbf{Old Hospital}        & \textbf{Shop Fa\c{c}ade}      & \textbf{St.\ Mary's Church}  & \textbf{Great Court} \\
		& 5600m$^2$                    & 50000m$^2$      & 2000m$^2$                    & 875m$^2$                      & 4800m$^2$                    & 8000m$^2$ \\
		\midrule
		Ours (Default)                               & \secondbest{0.07m}/\secondbest{0.24$^\circ$} (37.03\%) & -- & \secondbest{0.11m}/0.37$^\circ$ (35.17\%) & \secondbest{0.04m}/\secondbest{0.27$^\circ$} (60.19\%) & \secondbest{0.06m}/\secondbest{0.40$^\circ$} (42.45\%) & -- \\
		+ ICP                                        & \best{0.01m}/\best{0.06$^\circ$} (76.09\%) & -- & \best{0.01m}/\secondbest{0.06$^\circ$} (74.73\%) & \best{0.01m}/\best{0.04$^\circ$} (97.09\%) & \best{0.01m}/\best{0.06$^\circ$} (77.74\%) & -- \\
		+ Ranking                                    & \best{0.01m}/\best{0.06$^\circ$} (76.97\%) & -- & \best{0.01m}/\best{0.04$^\circ$} (82.97\%)  & \best{0.01m}/\best{0.04$^\circ$} (99.03\%) & \best{0.01m}/\best{0.06$^\circ$} (79.62\%) & -- \\
		\midrule
		PoseNet (Geom.\ Loss) \cite{Kendall2017}     & 0.99m/1.1$^\circ$ & \secondbest{20.7m}/\secondbest{25.7$^\circ$} & 2.17m/2.9$^\circ$ & 1.05m/4.0$^\circ$ & 1.49m/3.4$^\circ$  & 7.00m/3.7$^\circ$ \\
		Active Search (SIFT) \cite{Sattler2017}      & 0.42m/0.6$^\circ$ & \best{0.85m}/\best{0.8$^\circ$}  & 0.44m/1.0$^\circ$ & 0.12m/0.4$^\circ$ & 0.19m/0.5$^\circ$  & -- \\
		DSAC (RGB Training) \cite{Brachmann2017CVPR} & *0.30m/0.5$^\circ$ & -- & 0.33m/0.6$^\circ$ & 0.09m/0.4$^\circ$ & *0.55m/1.6$^\circ$ & \secondbest{*2.80m}/\secondbest{1.5$^\circ$} \\
		DSAC++ \cite{Brachmann2018CVPR}              & 0.18m/0.3$^\circ$  & -- & 0.20m/0.3$^\circ$ & 0.06m/\secondbest{0.3$^\circ$} & 0.13m/\secondbest{0.4$^\circ$}  & \best{0.40m}/\best{0.2$^\circ$} \\
		\bottomrule
	\end{tabular}
	\caption{Comparing the average median localisation errors (m/$^\circ$) of our \emph{adaptive} approach to those of state-of-the-art \emph{offline} methods on the Cambridge Landmarks dataset \cite{Kendall2015,Kendall2016,Kendall2017}. Since our approach requires depth, we only compare to methods that make use of the 3D models provided with the dataset, so as to facilitate a fairer comparison. As elsewhere in the paper, the percentages denote proportions of test frames with $\le 5$cm translation error and $\le 5^\circ$ angular error; red and blue colours denote respectively the best and second-best results in each column. Note that existing state-of-the-art methods do not report $5$cm/$5^\circ$ percentages on this dataset because their median translation errors are greater than $5$cm, but since our method achieves much lower errors for most scenes (with the exceptions of \emph{Street} and \emph{Great Court}), we are able to report these numbers as well. As elsewhere, `+ Ranking' means ranking the last $16$ candidates produced by RANSAC, as per \S\ref{subsubsec::hypothesisranking}. Note that the DSAC numbers marked with an asterisk were the result of end-to-end optimisation that did not converge. For all versions of our method, we report the results obtained by adapting a forest pre-trained on the \emph{Office} sequence from 7-Scenes \cite{Shotton2013}. For all scenes except \emph{Street} and \emph{Great Court}, we achieve results that are superior to the other methods, without needing to pre-train on the test scene. See \S\ref{subsec:outdoorrelocalisation} for further discussion.}
	\label{tbl:cambridgeresults}
\end{table*}

\begin{stusubfig}{!t}
	\begin{subfigure}{.47\linewidth}
		\centering
		\includegraphics[width=\linewidth]{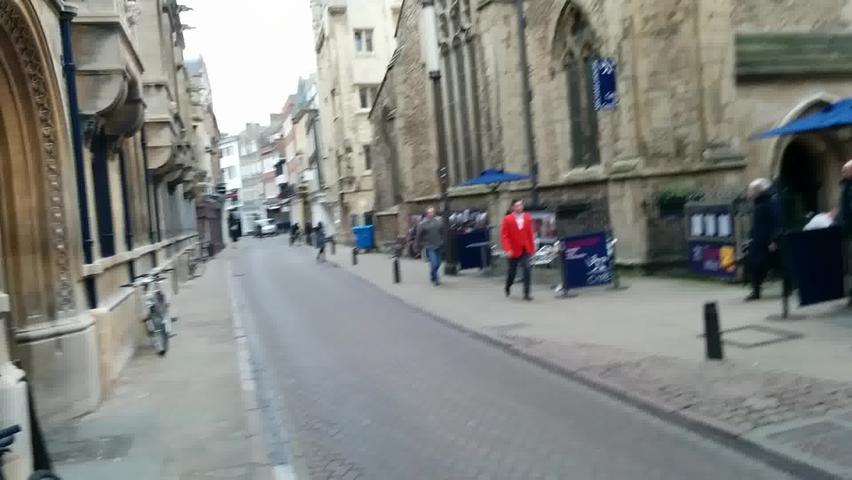}
		\caption{Street}
	\end{subfigure}%
	\hspace{4mm}%
	\begin{subfigure}{.47\linewidth}
		\centering
		\includegraphics[width=\linewidth]{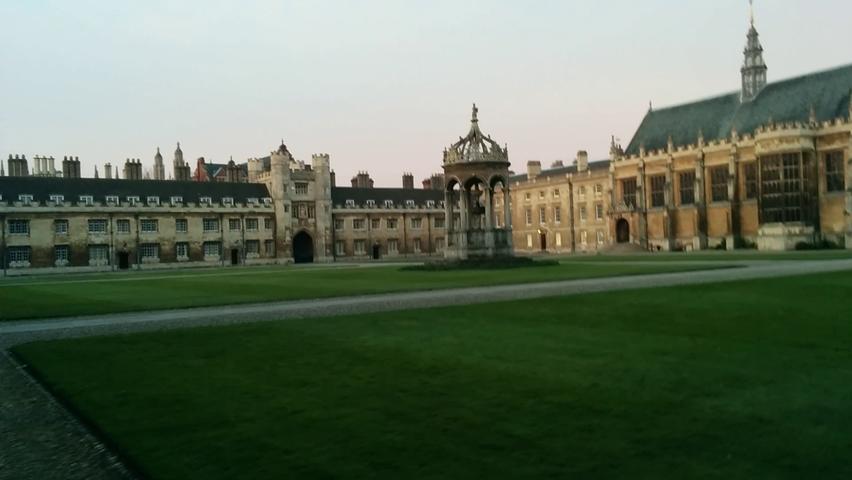}
		\caption{Great Court}
	\end{subfigure}%
	\caption{The \emph{Street} and \emph{Great Court} scenes from the Cambridge Landmarks dataset \cite{Kendall2015,Kendall2016,Kendall2017}.}
	\label{fig:cambridgescenes}
	\vspace{-\baselineskip}
\end{stusubfig}

Whilst our RGB-D relocaliser was primarily designed with indoor relocalisation in mind, it can also be used to relocalise outdoors. To show this, we evaluated its performance on the Cambridge Landmarks dataset \cite{Kendall2015,Kendall2016,Kendall2017}, which consists of a number of outdoor scenes at much larger scales\footnote{Note that for this reason, we used a voxel size of $4$cm in InfiniTAM for these models, and doubled the default size of the voxel hash table.} than those in either 7-Scenes \cite{Shotton2013} or Stanford 4 Scenes \cite{Valentin2016}. Whilst Cambridge Landmarks was originally designed for RGB-only relocalisation, depth images for each sequence can be constructed by rendering the 3D models provided with each scene as part of the dataset. For consistency with other works, we used the depth images rendered by Brachmann and Rother \cite{Brachmann2018CVPR} for this purpose.\footnote{We thank Eric Brachmann for providing us with these images.}

The results in Table~\ref{tbl:cambridgeresults} show how our approach compares to the best existing methods that also make use of the 3D models provided with Cambridge Landmarks. Encouragingly, we achieve state-of-art-results on four out of the six scenes on which we tested, showing that our approach has the potential to be effective for outdoor relocalisation. However, like some of the other methods in the table, our approach was unable to successfully relocalise in the remaining two scenes (\emph{Street} and \emph{Great Court}), owing to the significantly greater scales involved. As shown in Figure~\ref{fig:cambridgescenes}(a), \emph{Street} covers a $500$m $\times$ $100$m area \cite{Kendall2015}, which is an order of magnitude greater than \emph{Kings College} (the largest scene in which our method was able to successfully relocalise). To date, only Active Search \cite{Sattler2017}, which is based on SIFT, has been able to achieve reasonable relocalisation results in this scene. For \emph{Great Court}, the problem is not the overall scale of the scene, but that the camera sequences traverse the centre of a large quadrangle in Cambridge, such that most of the scene geometry is far away from the camera (see Figure~\ref{fig:cambridgescenes}(b)). Since our approach relies on reasonably accurate depth values (see \S\ref{subsubsec:noisydepth}), and depth inaccuracy tends to increase with distance, and since we also need there to be points in each leaf that are sufficiently close together to allow them to be clustered, and the points in an RGB-D image become sparser at greater distances due to perspective, it is unsurprising that our method struggles in this case. Methods like DSAC++ \cite{Brachmann2018CVPR} have a much better chance of working for scenes like \emph{Great Court}, because they train a network to predict the pose end-to-end on the target scene, and so are less dependent on the accuracy of the correspondences they predict as an intermediate step. However, their method, unlike ours, requires offline training on the target scene.

Overall, our initial results on Cambridge Landmarks suggest that our approach is already effective for moderately-sized outdoor scenes in which the average distance from the camera to the nearest scene geometry is not extreme. To make it also work effectively on city-scale scenes like \emph{Street}, we could in future consider initially using a coarse relocaliser to determine a particular area of the scene, and then using an instance of our relocaliser to yield an accurate pose within that area. However, extending our approach in this way is beyond the scope of this paper.

\section{Parameter Tuning}
\label{sec:parametertuning}

\begin{table*}[!t]
	\centering
	\begin{tabular}{cccccc}
		\toprule
		\textbf{Name} & \textbf{Category} & \textbf{$\theta^*$ (Default)} & \textbf{$\theta_1^*$ (Fast)} & \textbf{$\theta_2^*$ (Intermediate)} & \textbf{$\theta_3^*$ (Slow)} \\
		\midrule
		clustererSigma               & Forest & 0.1  & 0.1  & 0.1  & 0.1 \\
		clustererTau                 & Forest & 0.05 & 0.2  & 0.2  & 0.2 \\
		maxClusterCount ($M_{\max}$) & Forest & 50   & 50   & 50   & 50 \\
		minClusterSize               & Forest & 20   & 5    & 5    & 5 \\
		reservoirCapacity ($\kappa$) & Forest & 1024 & 2048 & 2048 & 2048 \\
		\midrule
		maxCandidateGenerationIterations                 & RANSAC & 6000 & 500   & 1000  & 250 \\
		maxPoseCandidates ($N_{\max}$)                   & RANSAC & 1024 & 2048  & 2048  & 2048 \\
		maxPoseCandidatesAfterCull ($N_{\textup{cull}}$) & RANSAC & 64   & 64    & 64    & 64 \\
		maxTranslationErrorForCorrectPose                & RANSAC & 0.05 & 0.05  & 0.1   & 0.1 \\
		minSquaredDistanceBetweenSampledModes            & RANSAC & 0.09 & 0     & 0.09  & 0.0225 \\
		poseUpdate                                       & RANSAC & True & False & True  & True \\
		ransacInliersPerIteration ($\eta$)               & RANSAC & 512  & 256   & 256   & 256 \\
		usePredictionCovarianceForPoseOptimization       & RANSAC & True & N/A   & False & False \\
		\midrule
		maxRelocalisationsToOutput                       & RANSAC & 16   & 1     & 1     & 16 \\
		\bottomrule
	\end{tabular}
	\caption{The parameters associated with our relocaliser, their values for our default relocaliser ($\theta^*$), and their tuned values for the individual relocalisers in our various cascades ($\theta_1^*,\theta_2^*,\theta_3^*$), as described in \S\ref{sec:cascadedesign}. Note that \texttt{maxCandidateGenerationIterations} is set to a seemingly low value for \emph{Slow}, even though the overall performance of \emph{Slow} is better than \emph{Fast} and \emph{Intermediate} overall -- as shown in Table~\ref{tbl:candidategenerationiterations}, this parameter makes little difference to the performance, and so can be tuned somewhat arbitrarily by the optimiser. In practice, \texttt{maxRelocalisationsToOutput}, which controls the number of hypotheses considered during model-based hypothesis ranking, has a much more significant effect on the relocalisation performance, which explains the better performance of \emph{Slow} overall.}
	\label{tbl:parameters}
\end{table*}

The parameters associated with our relocaliser are shown in Table~\ref{tbl:parameters}. Our goal when tuning is to find a set of values for them that let us accurately relocalise as many frames as possible, whilst staying within a fixed time bound to allow our relocaliser to be used in interactive contexts. To achieve this, we start by defining the following cost function, which computes a cost for running relocaliser $r$ on sequence $s$ in the context of a desired time bound $t_{\max}$:
\begin{equation}
\mathit{cost}(r,s,t_{\max}) = \begin{cases}
(1 - \mathit{score}(r,s))^2 & \mbox{if } \mathit{time}(r,s) \le t_{\max} \\
\infty & \mbox{otherwise}
\end{cases}
\end{equation}
In this, $\mathit{score}(r,s) \in [0,1]$ yields the fraction of the frames in $s$ that are correctly relocalised by $r$ to within $5$cm/$5^\circ$ of the ground truth, and $\mathit{time}(r,s)$ yields the average time that $r$ takes to run on a single frame of $s$.

\subsection{Tuning a Single Relocaliser}
\label{subsec:tuning-single}

To tune a single relocaliser, we choose a time bound $t_{\max}$ and then use the implementation of coordinate descent \cite{Wright2015} in the open-source SemanticPaint framework \cite{Golodetz2015SPTR} to find the parameters $\theta^*$ that minimise
\begin{equation}
\theta^* = \argmin_{\theta} \sum_{s \in \mathit{Sequences}} \mathit{cost}\left(\mathit{reloc}(\theta),s,t_{\max}\right),
\end{equation}
in which $\mathit{reloc}(\theta)$ denotes a variant of the relocaliser with parameters $\theta$. Any suitable set of sequences can be used for tuning; in our case, we took the training sequences from the well-known 7-Scenes dataset \cite{Shotton2013}, split them into training and validation subsets\footnote{To aid reproducibility, the splits are shared on our project page.}, and tuned on the validation subsets. (More precisely, for each scene, we first adapt the relocaliser by refilling its leaves with points from the training sequence for the scene, as per \S\ref{subsubsec::forestadaptation}, and then evaluate the cost on the validation sequence for the scene.)

\subsection{Tuning a Relocalisation Cascade}
\label{subsec:tuning-cascade}

Tuning a relocalisation cascade is more involved, since we need to tune not only the parameters for each individual relocaliser in the cascade, but also the depth-difference thresholds used to decide when to fall back from one relocaliser to the next in the sequence. Formally, let $\Theta = \{\theta_1,\ldots,\theta_N\}$ be the sets of parameters for the $N$ individual relocalisers in an $N$-stage relocalisation cascade, and let $\mathcal{T} = \{\tau_1,\ldots,\tau_{N-1}\}$ be the depth-difference thresholds used to decide when to fall back from one relocaliser to the next in the sequence. Then we could in principle cast the problem as
\begin{equation}
(\Theta^*,\mathcal{T}^*) = \argmin_{(\Theta,\mathcal{T})} \sum_{s \in \mathit{Sequences}} \mathit{cost}(\mathit{cascade}(\Theta,\mathcal{T}),s,t_{\max}),
\end{equation}
in which $\mathit{cascade}(\Theta,\mathcal{T})$ denotes a cascade with parameters $\Theta$ for the individual relocalisers and thresholds $\mathcal{T}$ to decide when to fall back from one relocaliser to the next. However, this has the disadvantage of treating the parameters for each relocaliser as completely independent of each other, whereas in reality we can significantly reduce the memory our approach needs by making all of the individual relocalisers in the cascade share the same regression forest.

To achieve this, we first observe that of the parameters in Table~\ref{tbl:parameters}, only those in the \emph{Forest} category are associated with the forest itself, whilst those in the \emph{RANSAC} category can be varied independently of the forest. We can therefore take the following approach to tuning a cascade:
\begin{enumerate}
	\item Choose $N$, the number of relocalisers to use in the cascade, and time bounds $t_{\max}^{(1)} > \ldots > t_{\max}^{(N)}$ for them. (See \S\ref{sec:cascadedesign} for how we chose these parameters.)
	\item For each individual relocaliser $i$, divide its parameters $\theta_i$ into shared ones associated with the forest ($\phi$) and independent ones associated with RANSAC ($\rho_i$):
	\begin{equation*}
	\theta_i = \phi \cup \rho_i
	\end{equation*}
	Then, tune all of the parameters of the fastest relocaliser to jointly find suitable parameters for the forest and RANSAC, before fixing the optimised forest parameters $\phi^*$ for all individual relocalisers in the cascade (we tune the forest parameters on the fastest relocaliser since it is impossible to get really fast relocalisation just by tuning the RANSAC parameters):
	
	\begin{equation}
	\theta_1^* = \argmin_{\theta} \sum_{s \in \mathit{Sequences}} \mathit{cost}\left(\mathit{reloc}(\theta),s,t_{\max}^{(1)}\right)
	\end{equation}
	
	\item Next, tune only the RANSAC parameters of all the other relocalisers in the cascade, i.e.\ for $i > 1$:
	
	\begin{equation}
	\rho_i^* = \argmin_{\rho} \sum_{s \in \mathit{Sequences}} \mathit{cost}\left(\mathit{reloc}(\phi^* \cup \rho),s,t_{\max}^{(i)}\right)
	\end{equation}
	
	\item Finally, having determined optimised parameters $\Theta^*$ for all the relocalisers in the cascade, tune the depth-difference thresholds between them by choosing a maximum average time bound $t_{\max}$ and minimising:
	
	\begin{equation}
	\mathcal{T}^* = \argmin_{\mathcal{T}} \sum_{s \in \mathit{Sequences}} \mathit{cost}\left(\mathit{cascade}(\Theta^*,\mathcal{T}),s,t_{\max}\right)
	\end{equation}
	
\end{enumerate}

\noindent The result of this process is a relocalisation cascade that takes no longer than $t_{\max}$ to relocalise on average. Provided we do not choose an overall time bound that is so low that it forces the cascade to always accept the results of the first relocaliser, this can also result in excellent average-case relocalisation performance (as we show in Table~\ref{tbl:comparativeperformance7}).

\section{Cascade Design}
\label{sec:cascadedesign}

\iftrue
\begin{table*}[!t]
	\centering
	\scriptsize
	\begin{tabular}{lcccccccccc}
		\toprule
		& \textbf{Chess} & \textbf{Fire} & \textbf{Heads} & \textbf{Office} & \textbf{Pumpkin} & \textbf{Kitchen} & \textbf{Stairs} & \textbf{Average} & \textbf{Avg.\ Median Error} & \textbf{Frame Time (ms)} \\
		\midrule
		Fast                 & 65.35\% & 47.55\% & 72.00\% & 56.10\% & 47.15\% & 47.60\% & 18.40\% & 50.59\% & 0.058m/2.44$^\circ$ & 25.06 \\
		Fast (w/ICP)         & 99.75\% & 97.10\% & 98.40\% & 99.55\% & 89.35\% & 89.26\% & 62.40\% & 90.83\% & 0.014m/1.17$^\circ$ & 29.91 \\
		\midrule
		Intermediate         & 86.60\% & 81.25\% & 69.00\% & 88.28\% & 67.20\% & 75.48\% & 53.20\% & 74.43\% & 0.033m/1.91$^\circ$ & 73.12 \\
		Intermediate (w/ICP) & 99.85\% & 98.90\% & 99.70\% & 99.73\% & 89.85\% & 90.34\% & 75.00\% & 93.34\% & 0.013m/1.17$^\circ$ & 77.87 \\
		\midrule
		Slow                 & 86.40\% & 81.25\% & 68.00\% & 88.70\% & 67.15\% & 75.42\% & 53.40\% & 74.33\% & 0.033m/1.95$^\circ$ & 77.78\\
		Slow (w/Ranking)     & 99.90\% &   100\% &   100\% & 99.55\% & 90.80\% & 89.38\% & 94.10\% & 96.25\% & 0.013m/1.17$^\circ$ & 203.64\\
		\midrule
		F$\stackrel{5\textup{cm}}{\rightarrow}$I & 99.85\% & 99.40\% & 99.70\% & 99.70\% & 90.65\% & 89.68\% & 76.00\% & 93.57\% & 0.013m/1.17$^\circ$ & 51.85 \\
		F$\stackrel{5\textup{cm}}{\rightarrow}$S & 99.90\% & 99.50\% & 99.80\% & 99.45\% & 90.50\% & 89.42\% & 91.80\% & 95.77\% & 0.013m/1.17$^\circ$ & 77.20 \\
		F$\stackrel{7.5\textup{cm}}{\rightarrow}$S & 99.90\% & 98.95\% & 99.90\% & 99.48\% & 90.95\% & 89.34\% & 86.10\% & 94.95\% & 0.013m/1.17$^\circ$ & 52.45 \\
		F$\stackrel{5\textup{cm}}{\rightarrow}$I$\stackrel{7.5\textup{cm}}{\rightarrow}$S & 99.85\% & 99.40\% & 99.90\% & 99.40\% & 90.85\% & 89.64\% & 89.80\% & 95.55\% & 0.013m/1.17$^\circ$ & 66.08 \\
		\bottomrule
	\end{tabular}
	\caption{The results of both the individual relocalisers that make up our cascades and the cascades themselves on 7-Scenes \cite{Shotton2013}. The percentages denote proportions of test frames with $\le 5$cm translation error and $\le 5^\circ$ angular error.}
	\label{tbl:cascadestages7}
\end{table*}
\fi

\iftrue
\begin{table*}[!t]
	\centering
	\scriptsize
	\begin{tabular}{lccccccc}
		\toprule
		\textbf{Sequence} & \textbf{Fast (w/ICP)} & \textbf{Intermediate (w/ICP)} & \textbf{Slow (w/Ranking)} & \textbf{F$\stackrel{5\textup{cm}}{\rightarrow}$I} & \textbf{F$\stackrel{5\textup{cm}}{\rightarrow}$S} & \textbf{F$\stackrel{7.5\textup{cm}}{\rightarrow}$S} & \textbf{F$\stackrel{5\textup{cm}}{\rightarrow}$I$\stackrel{7.5\textup{cm}}{\rightarrow}$S} \\
		\midrule
		Kitchen  &   100\% & 99.72\% &   100\% &   100\% &   100\% &   100\% &   100\% \\
		Living   & 99.80\% &   100\% &   100\% &   100\% &   100\% &   100\% &   100\% \\
		\midrule
		Bed      & 98.53\% & 98.53\% &   100\% & 99.51\% &   100\% &   100\% &   100\% \\
		Kitchen  &   100\% &   100\% &   100\% &   100\% &   100\% & 99.52\% &   100\% \\
		Living   &   100\% &   100\% &   100\% &   100\% &   100\% &   100\% &   100\% \\
		Luke     & 99.04\% & 99.20\% & 99.20\% & 99.20\% & 99.20\% & 99.20\% & 99.04\% \\
		\midrule
		Floor5a  & 97.99\% & 99.00\% & 98.99\% & 99.60\% & 99.80\% & 99.60\% &   100\% \\
		Floor5b  & 98.77\% & 98.77\% &   100\% & 99.01\% & 98.77\% & 99.01\% & 99.26\% \\
		\midrule
		Gates362 &   100\% &   100\% &   100\% &   100\% &   100\% &   100\% &   100\% \\
		Gates381 &   100\% &   100\% &   100\% &   100\% &   100\% & 99.91\% &   100\% \\
		Lounge   &   100\% &   100\% &   100\% &   100\% &   100\% &   100\% &   100\% \\
		Manolis  &   100\% &   100\% & 99.88\% &   100\% &   100\% &   100\% &   100\% \\
		\midrule
		\textbf{Average} & 99.51\% & 99.60\% & 99.84\% & 99.78\% & 99.81\% & 99.77\% & 99.86\% \\
		\textbf{Avg.\ Median Error} & 0.007m/0.26$^\circ$ & 0.007m/0.26$^\circ$ & 0.007m/0.26$^\circ$ & 0.007m/0.26$^\circ$ & 0.007m/0.26$^\circ$ & 0.007m/0.26$^\circ$ & 0.007m/0.26$^\circ$ \\
		\textbf{Frame Time (ms)} & 29.69 & 72.32 & 171.66 & 32.71 & 33.49 & 32.83 & 33.00 \\
		\bottomrule
	\end{tabular}
	\caption{The results of both the individual relocalisers that make up our cascades and the cascades themselves on Stanford 4 Scenes \cite{Valentin2016}. The percentages denote proportions of test frames with $\le 5$cm translation error and $\le 5^\circ$ angular error.}
	\label{tbl:cascadestages12}
\end{table*}
\fi

\iftrue
\begin{table}[!t]
	\centering
	\scriptsize
	\begin{tabular}{ccc}
		\toprule
		\textbf{Max.\ Candidate Generation Iterations} & \textbf{Average} & \textbf{Frame Time (ms)} \\
		\midrule
		250 & 85.47\% & 179.38 \\
		500 & 85.37\% & 182.54 \\
		1000 & 85.39\% & 184.48 \\
		\bottomrule
	\end{tabular}
	\caption{The average results of our \emph{Slow} relocaliser (with ranking enabled) on our validation subset of 7-Scenes \cite{Shotton2013}, for various different settings of the \texttt{maxCandidateGenerationIterations} parameter. In practice, we found that this parameter made very little difference to the results.}
	\label{tbl:candidategenerationiterations}
\end{table}
\fi

We had two goals when designing our cascades:
\begin{itemize}
	\item Obtain $\ge 85$\% accuracy for all sequences in both the 7-Scenes \cite{Shotton2013} and Stanford 4 Scenes \cite{Valentin2016} datasets.
	\item Relocalise in under $t_{\max} = 100$ms on average (for a frame rate of at least $10$ FPS), amortised across the entire dataset in each case.
\end{itemize}
Since the fastest relocalisers we were able to tune took around $30$ms to relocalise, there was a practical upper-bound of $N \le 3$ on the size of the cascades that we could use to meet the $100$ms time bound. We therefore chose to initially tune three individual relocalisers with $t_{\max}^{(1)} = 50\mbox{ms}$, $t_{\max}^{(2)} = 100\mbox{ms}$ and $t_{\max}^{(3)} = 200\mbox{ms}$, and then combine them into various possible cascades with $N = 2$ or $N = 3$ (tuning the thresholds separately in each case) to see if adding in a third relocaliser was worthwhile, or whether two relocalisers was enough.

The parameters that our tuning found for the individual relocalisers (which we call \emph{Fast}, \emph{Intermediate} and \emph{Slow}) are shown in Table~\ref{tbl:parameters}, and their results on 7-Scenes \cite{Shotton2013} and Stanford 4 Scenes \cite{Valentin2016} are shown in Tables~\ref{tbl:cascadestages7} and \ref{tbl:cascadestages12}.
Several interesting observations can be made from Table~\ref{tbl:parameters}. Firstly, it is noticeable that the \emph{Fast} relocaliser disables continuous pose optimisation, which is fairly costly in practice. In principle, we would expect this to have a significant negative effect on performance, and indeed (as can be seen in Table~\ref{tbl:cascadestages7}) this is the case prior to running ICP. However, the post-ICP results are actually relatively good, indicating that even without the pose optimisation, our approach is able to relocalise well enough to get into ICP's basin of convergence (intuitively, refining the final pose with ICP after the fact has a similar overall effect to optimising the pose hypotheses during RANSAC). Secondly, none of the tuned relocalisers make use of the covariance information in the leaves during pose optimisation, indicating that it may be possible to avoid storing it in practice. This is a potentially important observation for future work, since storing the covariance has a significant memory cost. Finally, it is noticeable that the \emph{Slow} relocaliser only attempts to generate a pose candidate on each thread at most $250$ times, whereas the other relocalisers all try much harder to generate the initial candidates. In principle, we might hope for the performance to be slightly better for higher \texttt{maxCandidateGenerationIterations} values, but in practice we found that for \emph{Slow}, it made little actual difference to our results (see Table~\ref{tbl:candidategenerationiterations}), indicating that most threads do not actually need more than $250$ iterations to generate a candidate. It is also worth mentioning that our optimiser was explicitly designed to find sets of parameters that perform well within a particular time bound, in order to relocalise quickly on average, and so even if the performance had been slightly better for higher values of this parameter, it would still have been naturally inclined to focus on parameters that make a significant difference to performance (e.g.\ \texttt{maxRelocalisationsToOutput}) at the expense of more minor parameters such as this one.

\begin{stusubfig*}{!t}
	\begin{subfigure}{.49\linewidth}
		\centering
		\includegraphics[height=5.5cm]{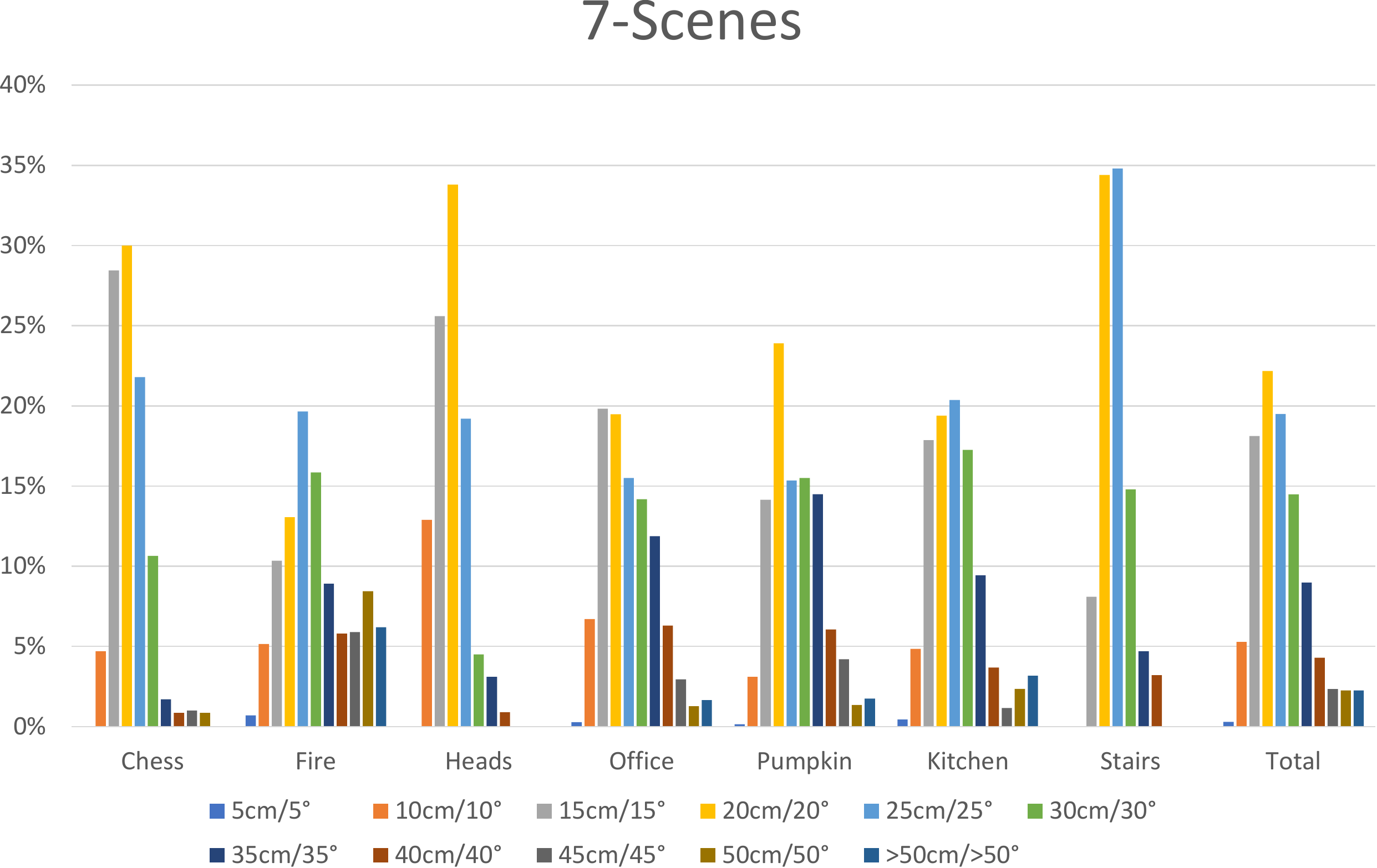}
	\end{subfigure}%
	\hspace{1mm}%
	\begin{subfigure}{.49\linewidth}
		\centering
		\includegraphics[height=5.5cm]{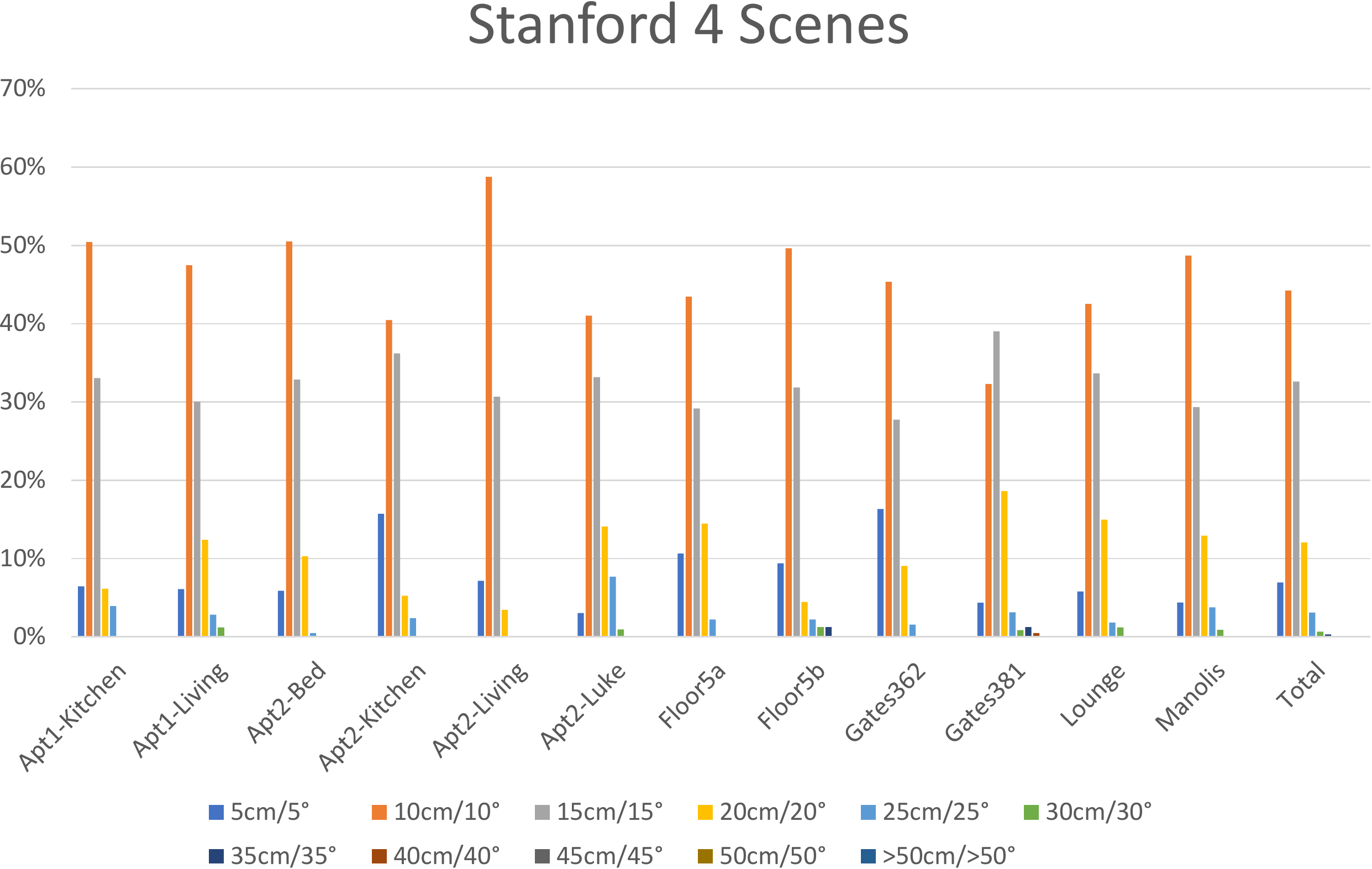}
	\end{subfigure}%
	\\[\baselineskip]
	\begin{subfigure}{.49\linewidth}
		\centering
		\includegraphics[height=5.5cm]{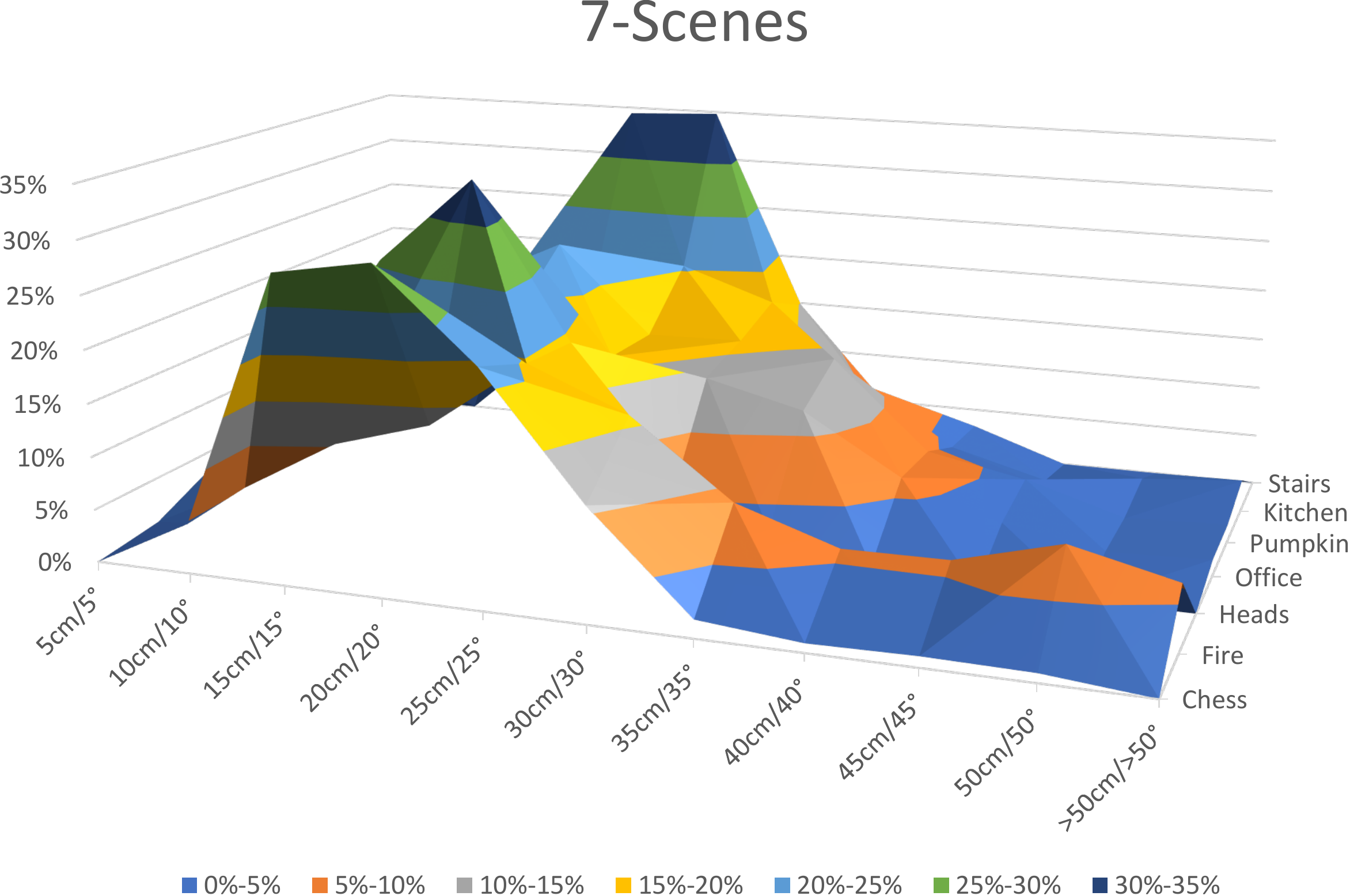}
	\end{subfigure}%
	\hspace{1mm}%
	\begin{subfigure}{.49\linewidth}
		\centering
		\includegraphics[height=5.5cm]{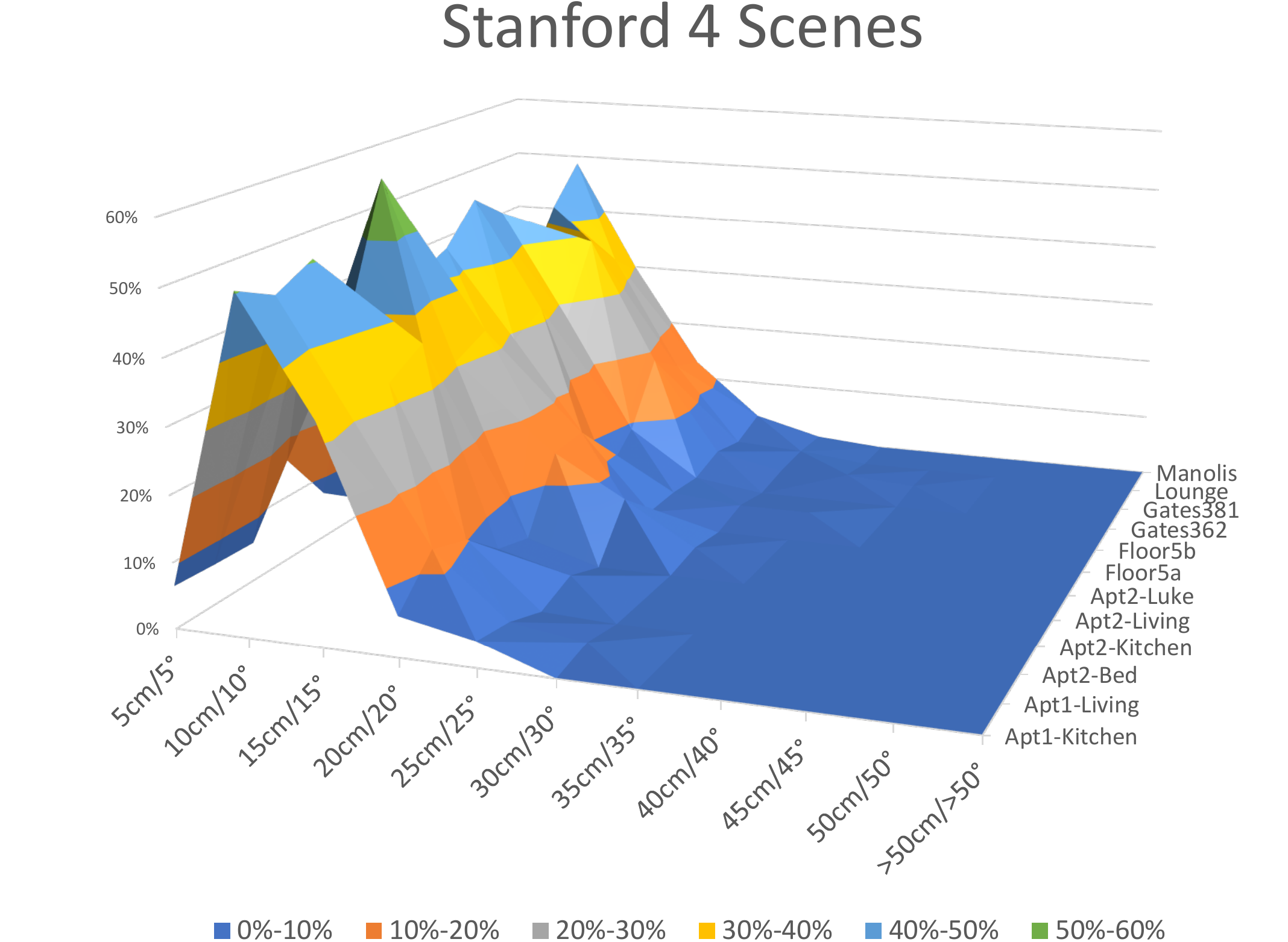}
	\end{subfigure}%
	\caption{The proportions of test frames from both 7-Scenes \cite{Shotton2013} and Stanford 4 Scenes \cite{Valentin2016} that are within certain distances of the training trajectories, visualised in two different ways. It is noticeable, particularly from the 3D surface plots, that in Stanford 4 Scenes the vast majority of the test frames fall within 30cm/30$^\circ$ of the training trajectory, whilst in 7-Scenes, far more of the test frames are at a much greater distance, particularly those from the \emph{Fire} sequence. This makes 7-Scenes a much harder dataset to saturate in practice.}
	\label{fig:datasetanalysis}
	\vspace{-\baselineskip}
\end{stusubfig*}

Having tuned the individual relocalisers, we then used them to construct and tune three different types of cascade: (i) \emph{Fast} $\rightarrow$ \emph{Intermediate}, (ii) \emph{Fast} $\rightarrow$ \emph{Slow}, and (iii) \emph{Fast} $\rightarrow$ \emph{Intermediate} $\rightarrow$ \emph{Slow}. We hypothesised that relocalisers of type (i) might be fast but have performance that was limited by that of the \emph{Intermediate} relocaliser, that those of type (ii) might achieve good results but be slow (since they might be forced to run the slow relocaliser on only moderately hard frames), and that those of type (iii) might be best overall.
For all types, our tuning process found that a depth-difference threshold of $5$cm was a good choice for falling back from \emph{Fast} to \emph{Intermediate}, and that $7.5$cm was a good choice for falling back from \emph{Intermediate} to \emph{Slow}. For falling back from \emph{Fast} to \emph{Slow}, the tuning proposed $5$cm as a good threshold, but we decided to also try $7.5$cm (the proposed \emph{Intermediate} to \emph{Slow} threshold) to see if an interesting alternative balance between accuracy and speed could be achieved.

We evaluated all of these cascades on the 7-Scenes \cite{Shotton2013} and Stanford 4 Scenes \cite{Valentin2016} datasets. On Stanford 4 Scenes (see Table~\ref{tbl:cascadestages12}), all four cascades we tested achieved almost perfect results, which we believe can be attributed to the relatively straightforward nature of the underlying dataset (see \S\ref{sec:datasetanalysis}), leading to all four cascades choosing to run the fast relocaliser on almost all frames. The results on 7-Scenes (see Table~\ref{tbl:cascadestages7}) are more interesting. In particular, we found that, as expected, our F$\stackrel{5\textup{cm}}{\rightarrow}$I cascade was relatively fast, but was unable to achieve high-quality results on all sequences. Also as expected, our 3-stage cascade (F$\stackrel{5\textup{cm}}{\rightarrow}$I$\stackrel{7.5\textup{cm}}{\rightarrow}$S) was generally preferable to F$\stackrel{5\textup{cm}}{\rightarrow}$S, with similar accuracy and a higher frame rate, as a result of its ability to use the \emph{Intermediate} relocaliser on only moderately difficult frames. Interestingly, our hand-tuned relocaliser (F$\stackrel{7.5\textup{cm}}{\rightarrow}$S) was also good, achieving acceptable results on all sequences whilst running at almost the speed of F$\stackrel{5\textup{cm}}{\rightarrow}$I.

Overall, we believe that the best cascade to choose most likely depends on the application at hand. All four cascades achieved accurate relocalisation at a high frame rate. The differences between them were most exposed by the \emph{Stairs} sequence from 7-Scenes \cite{Shotton2013}, which is a notoriously difficult sequence that most approaches have struggled to cope with. For more real-world use, the differences between our cascades are most likely small enough that they can be ignored in practice.

\section{Dataset Analysis}
\label{sec:datasetanalysis}

\begin{stusubfig*}{!t}
	\begin{subfigure}{\linewidth}
		\centering
		\includegraphics[width=.9\linewidth]{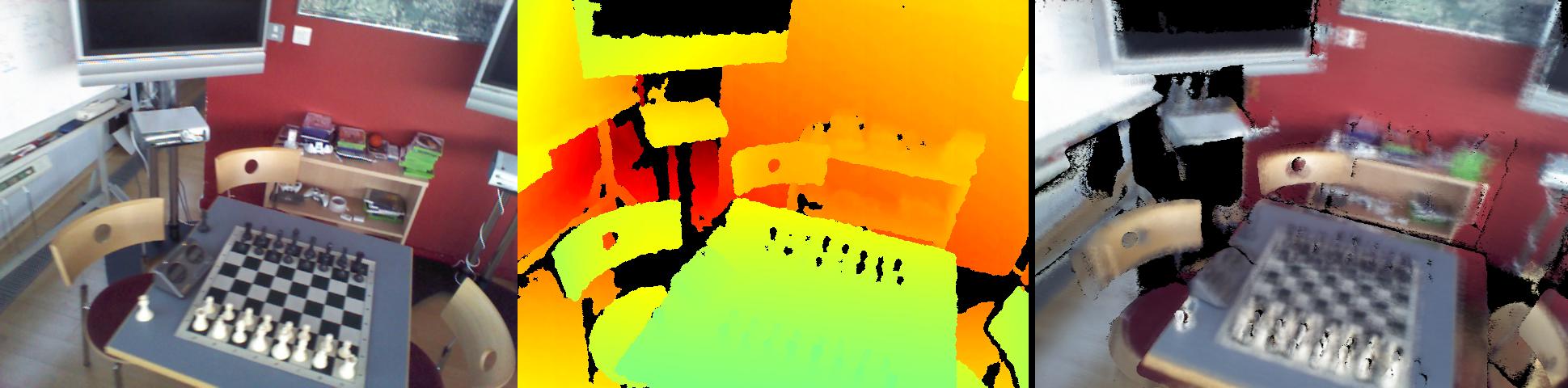}
	\end{subfigure}%
	\\[\baselineskip]
	\begin{subfigure}{\linewidth}
		\centering
		\includegraphics[width=.9\linewidth]{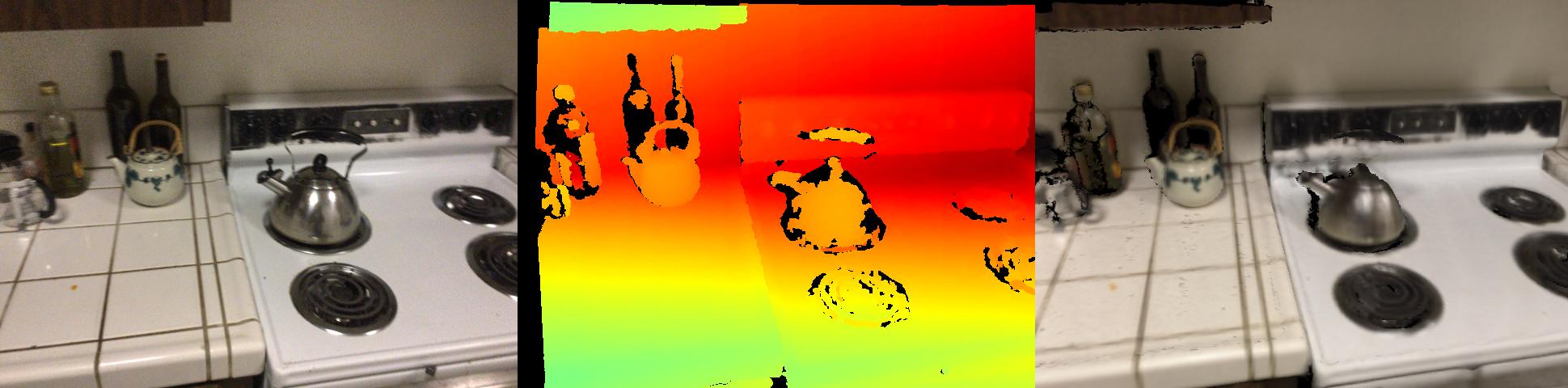}
	\end{subfigure}%
	\caption{The 7-Scenes dataset \cite{Shotton2013} (top row) was unfortunately captured with the internal calibration between the colour and depth cameras disabled, leading to poor alignment between the two. This is especially noticeable in the colour fusion results (see right-hand image). By contrast, the images in the Stanford 4 Scenes dataset \cite{Valentin2016} (bottom row) are accurately aligned, leading to much better results.}
	\label{fig:alignment}
	\vspace{-\baselineskip}
\end{stusubfig*}

To better understand the near-perfect results of all four of our cascades on the Stanford 4 Scenes \cite{Valentin2016} dataset (see \S\ref{sec:cascadedesign}), we analysed the proportions of test frames from both 7-Scenes \cite{Shotton2013} and Stanford 4 Scenes that are within certain distances of the training trajectories. The results, as shown in Figure~\ref{fig:datasetanalysis}, help explain why our approach invariably achieves such good results on Stanford 4 Scenes, whilst not yet fully saturating the more difficult 7-Scenes benchmark. In particular, it is noticeable that in Stanford 4 Scenes, the vast majority of the test frames fall within 30cm/30$^\circ$ of the training trajectory, whilst in 7-Scenes, far more of the test frames are at a much greater distance, particularly those from the \emph{Fire} sequence. This makes 7-Scenes a far harder benchmark in practice: test frames that are near the training trajectory are much easier to relocalise, since there is then less need to match keypoints across scale/viewpoint changes. In Stanford 4 Scenes, almost all of the sequences are dominated by test frames that are around 10-15cm/$^\circ$ from the training trajectory, making it an easy dataset to saturate.

Two other considerations make 7-Scenes difficult to fully saturate in practice. The more significant of the two is that the original dataset was captured with KinectFusion \cite{Newcombe2011}, which is prone to tracking drift, even at room scale, meaning that in practice the ground truth poses for some frames can be slightly inaccurate. This can cause at least two different types of problem: firstly, if the ground truth poses for the training sequence are slightly inaccurate, then InfiniTAM is liable to fuse an imperfect 3D model (e.g.\ see Figure~\ref{fig:redkitchen71-model}), which can affect the poses to which our ICP-based refinement process will converge at test time; secondly, if the ground truth poses for the testing sequence are slightly inaccurate, then relocalised poses that would have been within 5cm/5$^\circ$ of a `perfect' pose can be marked as incorrect, and other poses that would have been too far from the `perfect' pose but are within 5cm/5$^\circ$ of the ground truth pose given can be marked as correct. Dataset problems like this are unfortunately very difficult to mitigate at the level of an individual method such as ours -- whilst it might in principle be possible (if time-consuming) to bundle adjust all of the frames in each sequence to correct any inaccurate poses, any results obtained on the corrected sequences would then be incomparable with those obtained on the standard dataset. As such, we limit ourselves in this paper to simply noting that this problem is a noticeable failure mode of our approach on this dataset (see \S\ref{subsec:badgroundtruth}), and in common with other approaches, rely on the ground truth poses as given when computing our results.

A more minor issue is that the dataset was unfortunately captured with the internal calibration between the depth and colour cameras disabled, leading to poor alignment between the two (see Figure~\ref{fig:alignment}). In the context of our approach, this can lead to a slight offset between the locations at which the depth and RGB features for each pixel are computed, which can in principle cause the correspondences for some pixels (particularly those near the boundaries of objects) to be incorrectly predicted by the forest. However, in practice we found that we were fairly robust to this problem: most of the correspondences are still predicted correctly, and incorrect correspondences are in any case handled naturally by the RANSAC stage of our pipeline.

\section{Failure Case Analysis}
\label{sec:failurecaseanalysis}

As shown in both the main paper and this supplementary material, our approach is able to achieve highly-accurate online relocalisation in real time, from novel poses and without needing extensive offline training on the target scene. However, there are inevitably still situations in which it will fail. In this section, we analyse a few examples, and attempt to explain the underlying reasons in each case.

\subsection{Visual Ambiguities}

One of the most common reasons for our relocaliser to fail is the presence of repetitive structures and/or textures in the scene. Two examples of this are shown in the following subsections -- one showing a staircase, and the other showing a stretch of similar-looking red cupboards. Our relocaliser can sometimes struggle in situations like this because it relies on local features around each pixel to predict the pixel's world-space coordinates, and these local features do not always provide sufficient context for it to disambiguate between similar-looking points. We achieve significant robustness to this problem by allowing multiple correspondences to be predicted for each pixel (we use forests with multiple trees, and store multiple clusters of world-space points in each leaf), but for some inputs, our relocaliser can still fail.

\subsubsection{Stairs}

\stufig{width=\linewidth}{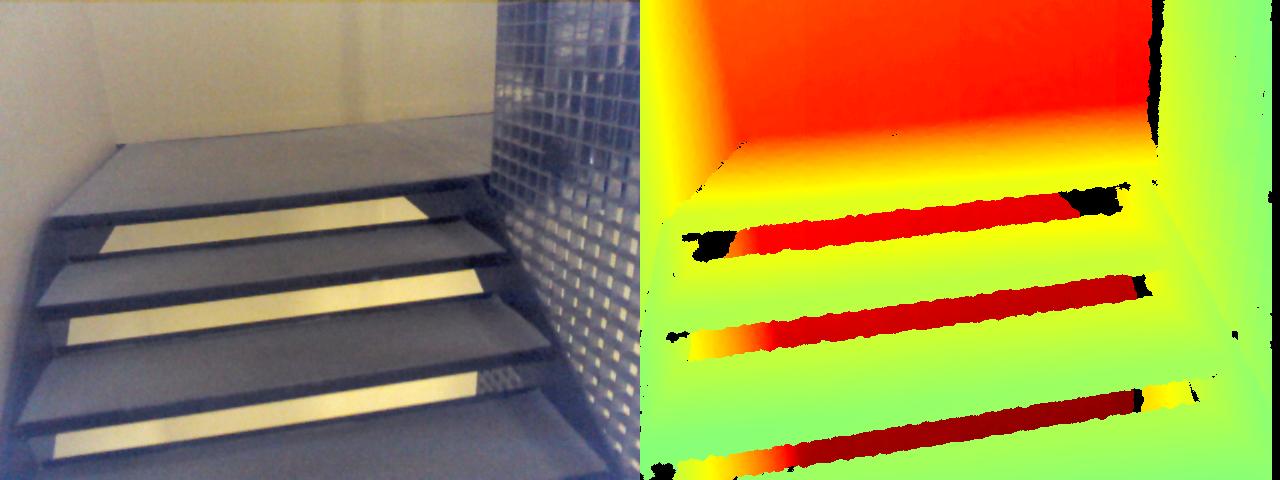}{The $470$th frame of the \emph{Stairs} sequence from 7-Scenes \cite{Shotton2013}. This is an example of an input that can confuse our relocaliser, owing to the presence of multiple visually-identical steps.}{fig:stairs470-input}{!t}

\stufig{width=\linewidth}{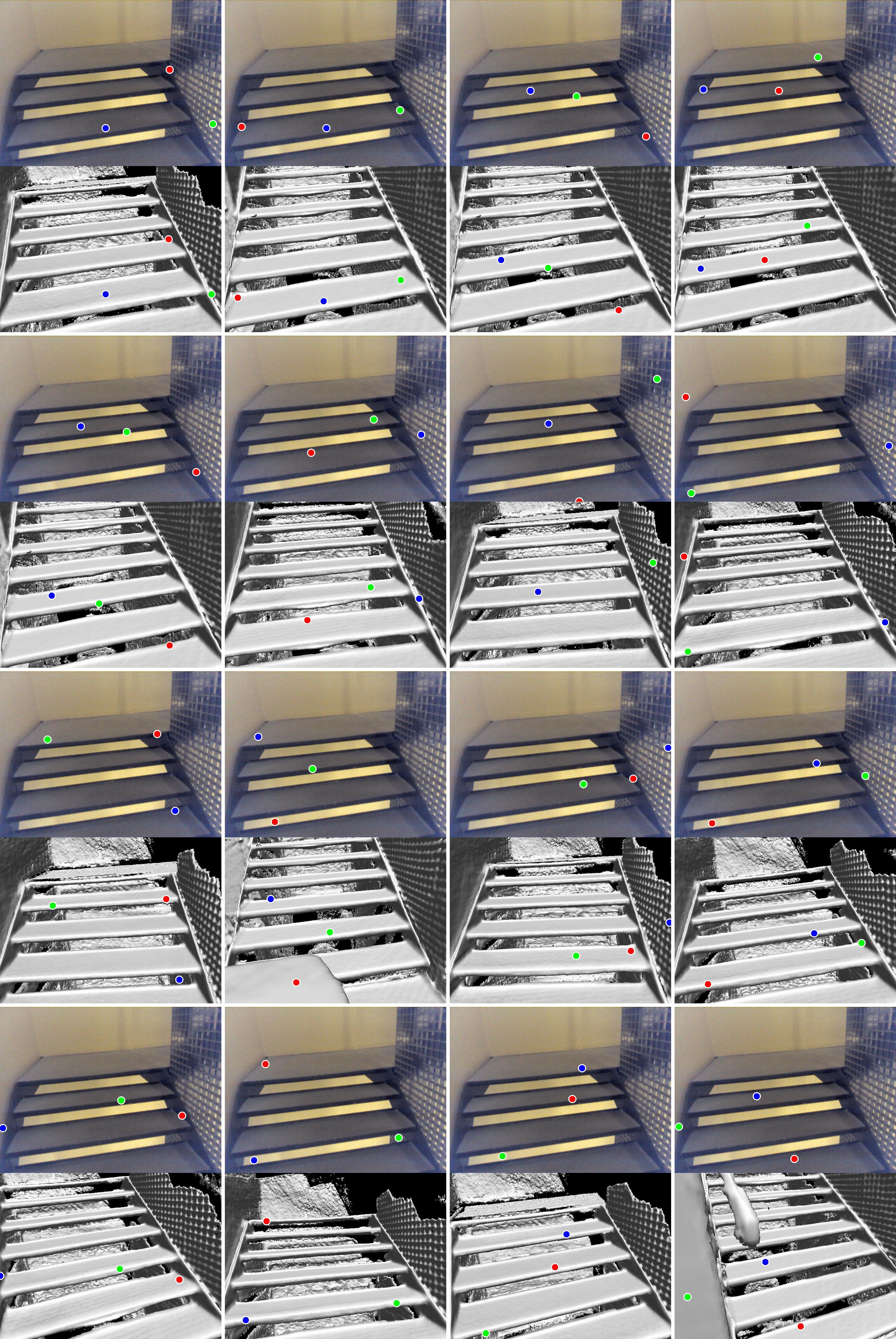}{The top $16$ pose candidates (left-to-right, top-to-bottom) corresponding to the failure case from the \emph{Stairs} scene shown in Figure~\ref{fig:stairs470-input}. The coloured points indicate the 2D-to-3D correspondences that are used to generate the initial pose hypotheses. Note that in this case, none of the candidates would relocalise the camera successfully. This is because the points at the same places on different stairs tend to end up in similar leaves, making the modes in the leaves less informative and significantly reducing the probability of generating good initial hypotheses.}{fig:stairs470-candidates}{!t}

The first example we consider is from the \emph{Stairs} scene in 7-Scenes \cite{Shotton2013}. This is a notoriously difficult scene containing a staircase that consists of numerous visually-identical steps. When viewing the scene from certain angles, the relocaliser is able to rely on points in the scene that can be identified unambiguously to correctly estimate the pose, but from viewpoints such as that in Figure~\ref{fig:stairs470-input}, it is forced to use more ambiguous points, e.g.\ those on the stairs themselves or the walls. When this happens, relocalisation is prone to fail, since the relocaliser finds it difficult to tell the difference between the different steps.

To illustrate this, we visualise the last $16$ surviving camera hypotheses for this instance in Figure~\ref{fig:stairs470-candidates}, in descending order (left-to-right, top-to-bottom). It is noticeable that in this case, none of the top $16$ hypotheses would have successfully relocalised the camera. As suggested by the points predicted in the 3D scene for each hypothesis (which are often in roughly the right place but on the wrong stairs), this is because the points at the same places on different stairs tend to end up in similar leaves, making the modes in the leaves less informative and significantly reducing the probability of generating good initial hypotheses.

\subsubsection{Pumpkin}

The second example we consider is from the \emph{Pumpkin} sequence (see Figure~\ref{fig:pumpkin920-input}). Here, the input image seems slightly easier to relocalise than in the previous example, but in practice, the repetitive red cupboards and dark grey ceiling panels (not to mention the reflections from the ceiling lights) provide many opportunities for a relocaliser such as ours to get confused (see Figure~\ref{fig:pumpkin920-candidates}). In this case, as with the \emph{Stairs} example, all of the individual matches seem individually reasonable (in the sense that the matched points genuinely do have a similar appearance), but the end result is nevertheless to relocalise in the wrong place.

\subsubsection{Analysis}

Notably, in both of these examples, there seem to be at least a few visually distinctive points that could have been chosen in order to successfully relocalise (e.g.\ for the \emph{Pumpkin} example, the top-right corner of the cupboards, the T-junction between the cupboards and the machine, and the top-left corner of the white notice all look useful), but no candidates based on these points were ever generated.
Ultimately, this is caused by the fact that RANSAC randomly samples only a subset of all of the possible candidates, and there is thus always a possibility of missing a candidate that might have worked.
One way of mitigating this problem might be to explicitly search for visually distinctive points in the image (i.e.\ reintroduce a keypoint detection stage into the pipeline) and generate additional candidates based on any points found. We do not currently implement this approach (and it would be expected to carry a speed cost), but it offers an interesting avenue for further work.

\subsection{Inaccurate Ground Truth Poses}
\label{subsec:badgroundtruth}

As mentioned in \S\ref{sec:datasetanalysis}, one of the other main types of failure we experienced was ultimately caused by slightly inaccurate ground truth poses in the 7-Scenes dataset \cite{Shotton2013}, which was captured using KinectFusion \cite{Newcombe2011}, a reconstruction method that is known to be prone to tracking drift. To demonstrate this, we consider an example from the \emph{Red Kitchen} scene (see Figure~\ref{fig:redkitchen71-input}).

\stufig{width=\linewidth}{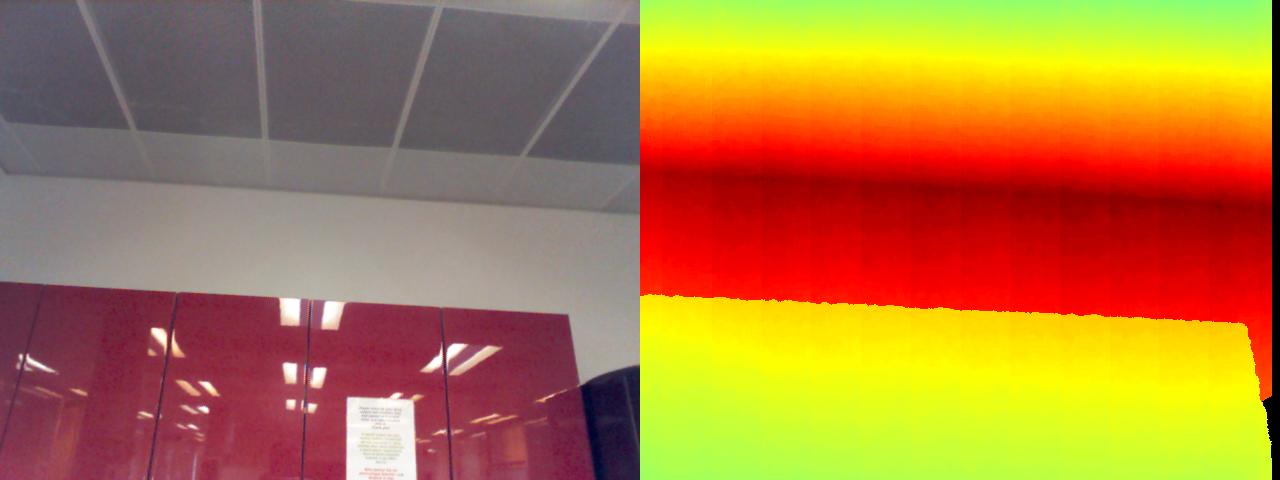}{The $920$th frame of the \emph{Pumpkin} sequence from 7-Scenes \cite{Shotton2013}. This is another example that can confuse our relocaliser, owing to the presence of the repetitive red cupboards and dark grey ceiling panels.}{fig:pumpkin920-input}{!t}

\stufig{width=\linewidth}{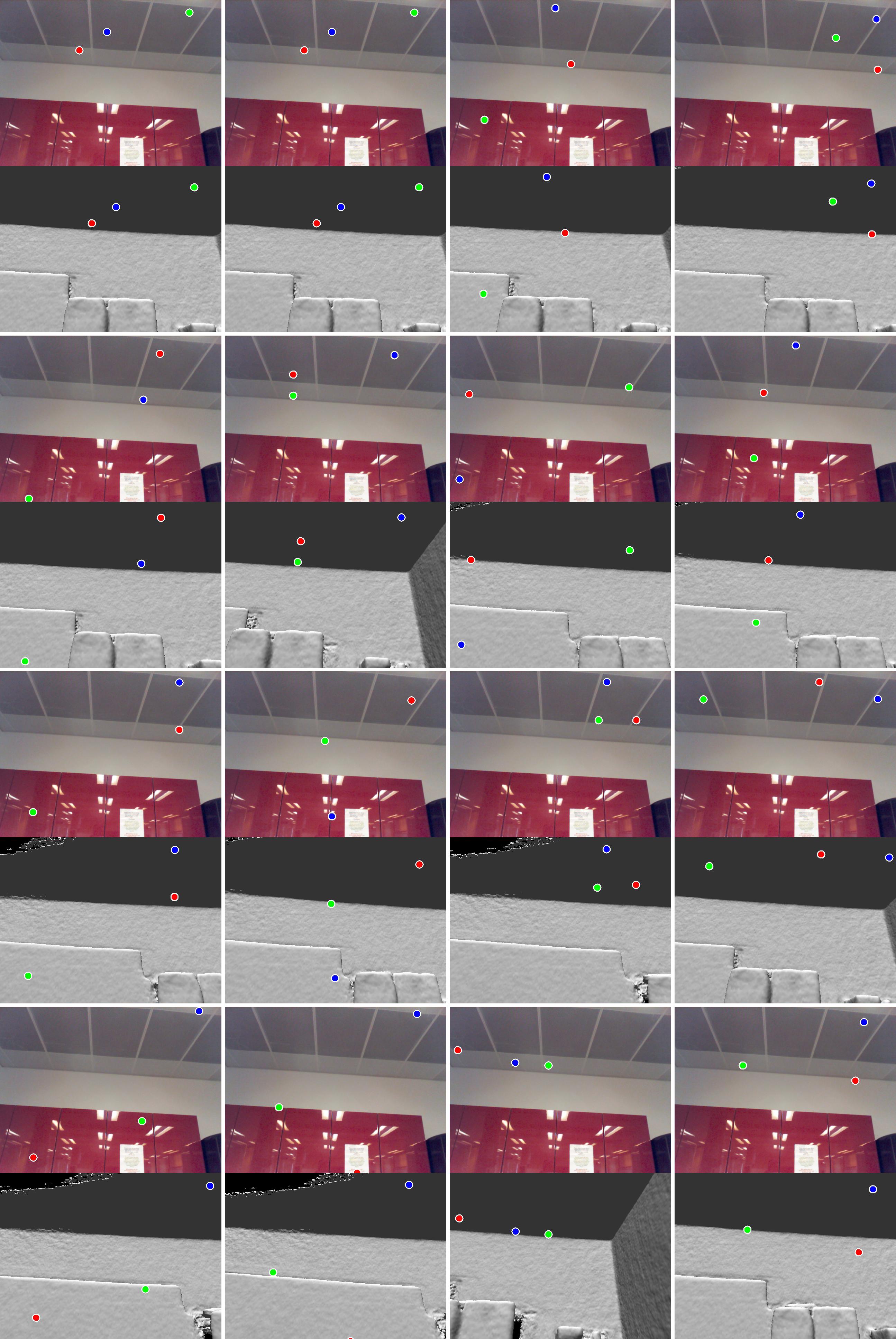}{The top $16$ pose candidates (left-to-right, top-to-bottom) corresponding to the failure case from the \emph{Pumpkin} scene shown in Figure~\ref{fig:pumpkin920-input}. The coloured points indicate the 2D-to-3D correspondences that are used to generate the initial pose hypotheses. Note that in this case, none of the candidates would relocalise the camera successfully, although several are close to being acceptable.}{fig:pumpkin920-candidates}{!t}

\stufig{width=\linewidth}{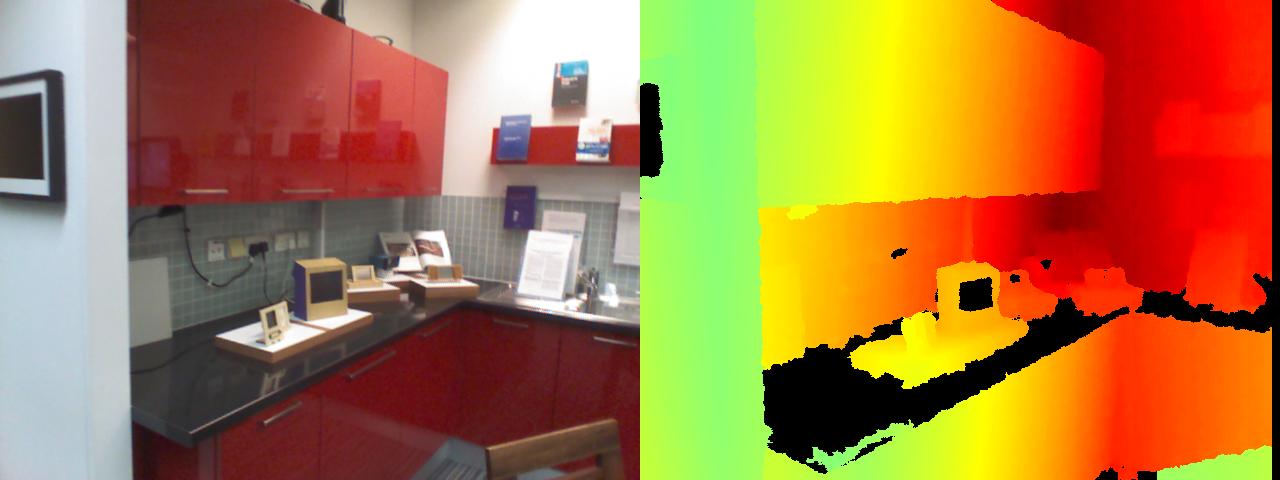}{The $71$st frame of the \emph{Red Kitchen} testing sequence from 7-Scenes \cite{Shotton2013}. Here, our relocaliser initially produces a pose that is within $5$cm/$5^\circ$ of the ground truth, but ICP against the fused 3D model then refines it to a pose that is further from the ground truth. The ultimate cause of this is slightly inaccurate ground truth poses in the training sequence, leading to InfiniTAM fusing an imperfect 3D model, which then affects the pose to which ICP converges.}{fig:redkitchen71-input}{!t}

\begin{stusubfig}{!t}
	\begin{subfigure}{.32\linewidth}
		\centering
		\includegraphics[width=\linewidth]{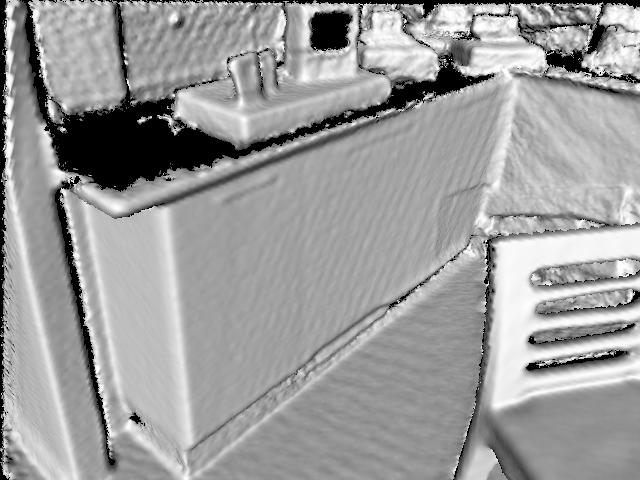}
	\end{subfigure}%
	\hspace{1mm}%
	\begin{subfigure}{.32\linewidth}
		\centering
		\includegraphics[width=\linewidth]{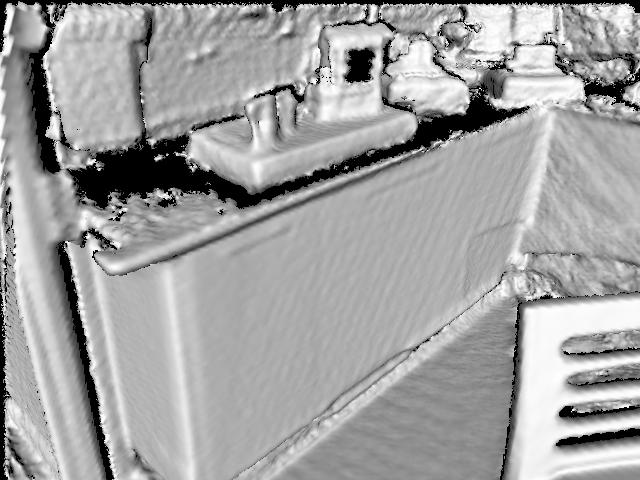}
	\end{subfigure}%
	\hspace{1mm}%
	\begin{subfigure}{.32\linewidth}
		\centering
		\includegraphics[width=\linewidth]{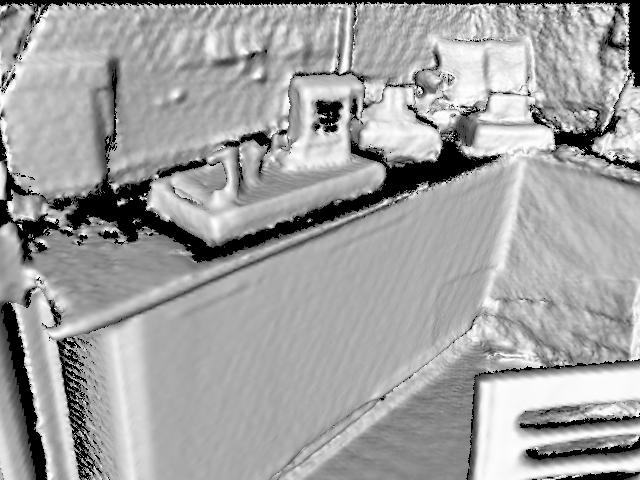}
	\end{subfigure}%
	\\[.25\baselineskip]
	\begin{subfigure}{.32\linewidth}
		\centering
		\includegraphics[width=\linewidth]{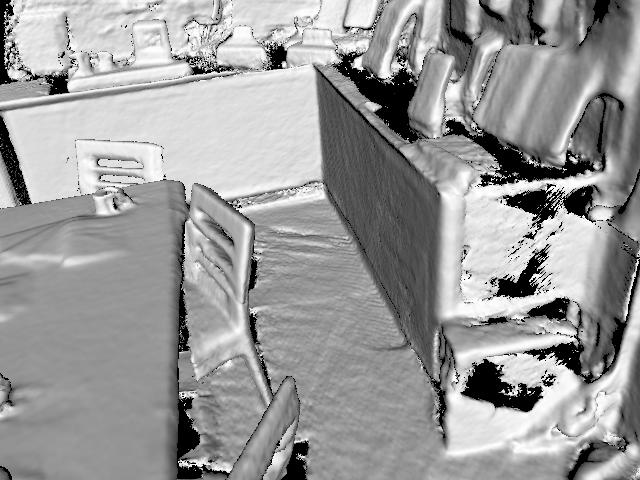}
	\end{subfigure}%
	\hspace{1mm}%
	\begin{subfigure}{.32\linewidth}
		\centering
		\includegraphics[width=\linewidth]{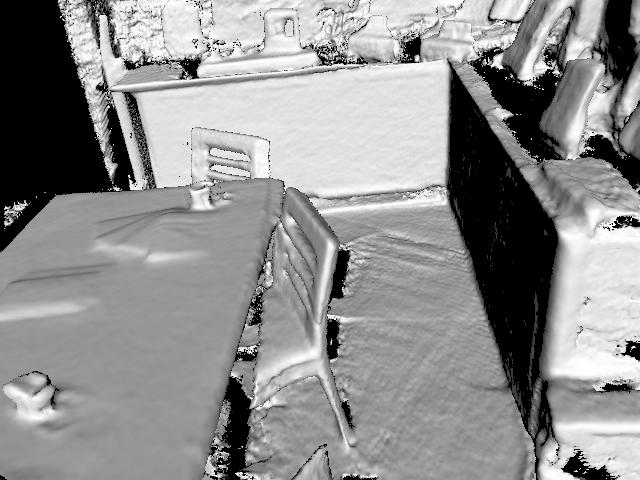}
	\end{subfigure}%
	\hspace{1mm}%
	\begin{subfigure}{.32\linewidth}
		\centering
		\includegraphics[width=\linewidth]{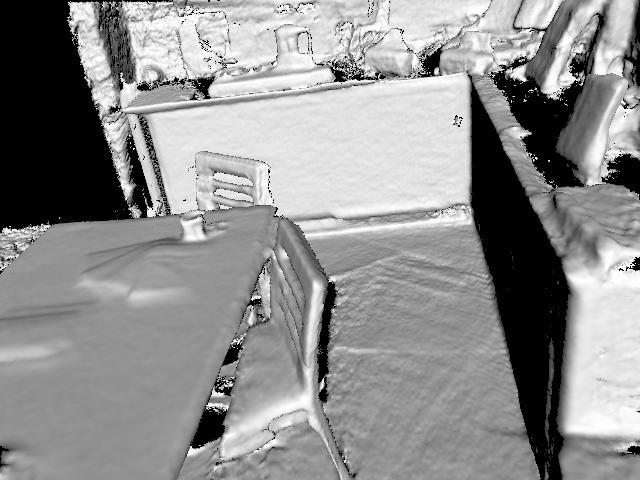}
	\end{subfigure}%
	\\[.25\baselineskip]
	\begin{subfigure}{.32\linewidth}
		\centering
		\includegraphics[width=\linewidth]{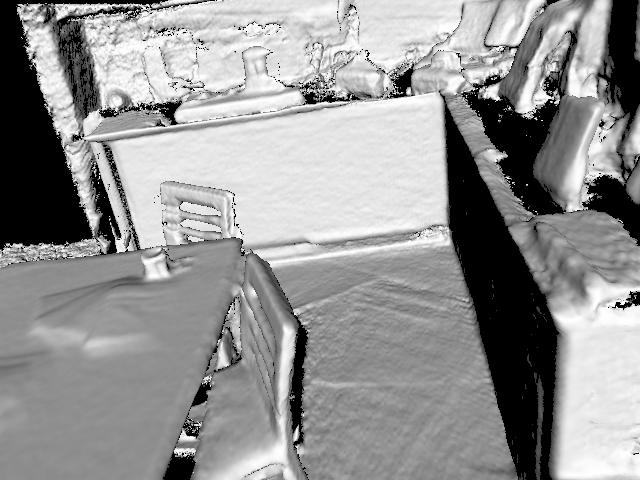}
	\end{subfigure}%
	\hspace{1mm}%
	\begin{subfigure}{.32\linewidth}
		\centering
		\includegraphics[width=\linewidth]{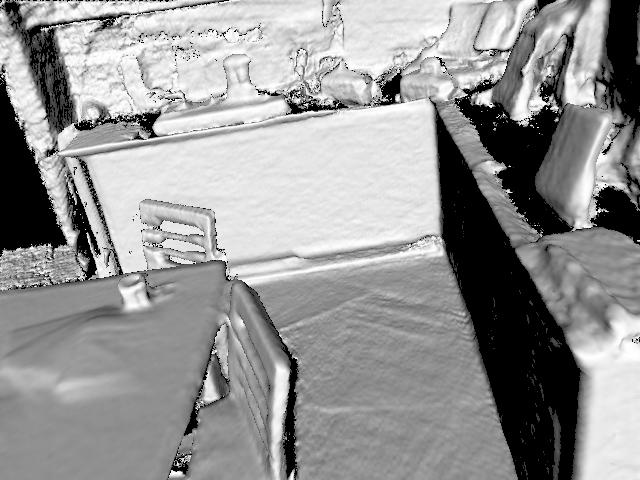}
	\end{subfigure}%
	\hspace{1mm}%
	\begin{subfigure}{.32\linewidth}
		\centering
		\includegraphics[width=\linewidth]{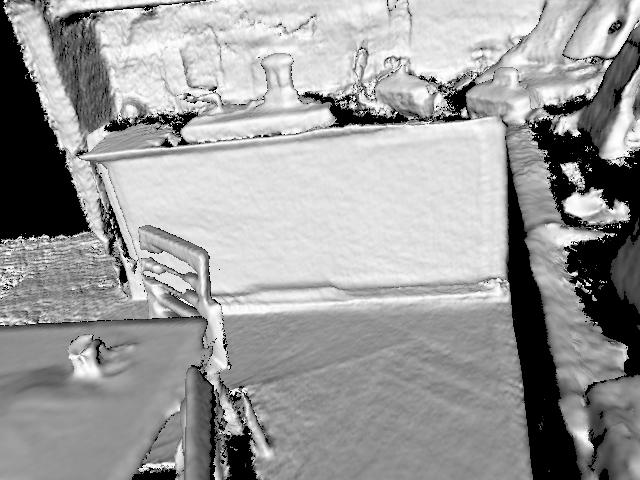}
	\end{subfigure}%
	\caption{The process of fusing the \emph{Red Kitchen} training sequence from 7-Scenes \cite{Shotton2013} using InfiniTAM (left-to-right, top-to-bottom). Initially (top-left), the brown box on the worktop is correctly reconstructed, but as the fusion process proceeds, it gradually becomes more and more eroded. This is ultimately caused by slightly inaccurate ground truth poses in the dataset, causing points that should be fused into the same place in the model to be fused at a slight offset from each other, with the results as shown (bottom-right).}
	\label{fig:redkitchen71-model}
	\vspace{-.5\baselineskip}
\end{stusubfig}

\stufig{width=\linewidth}{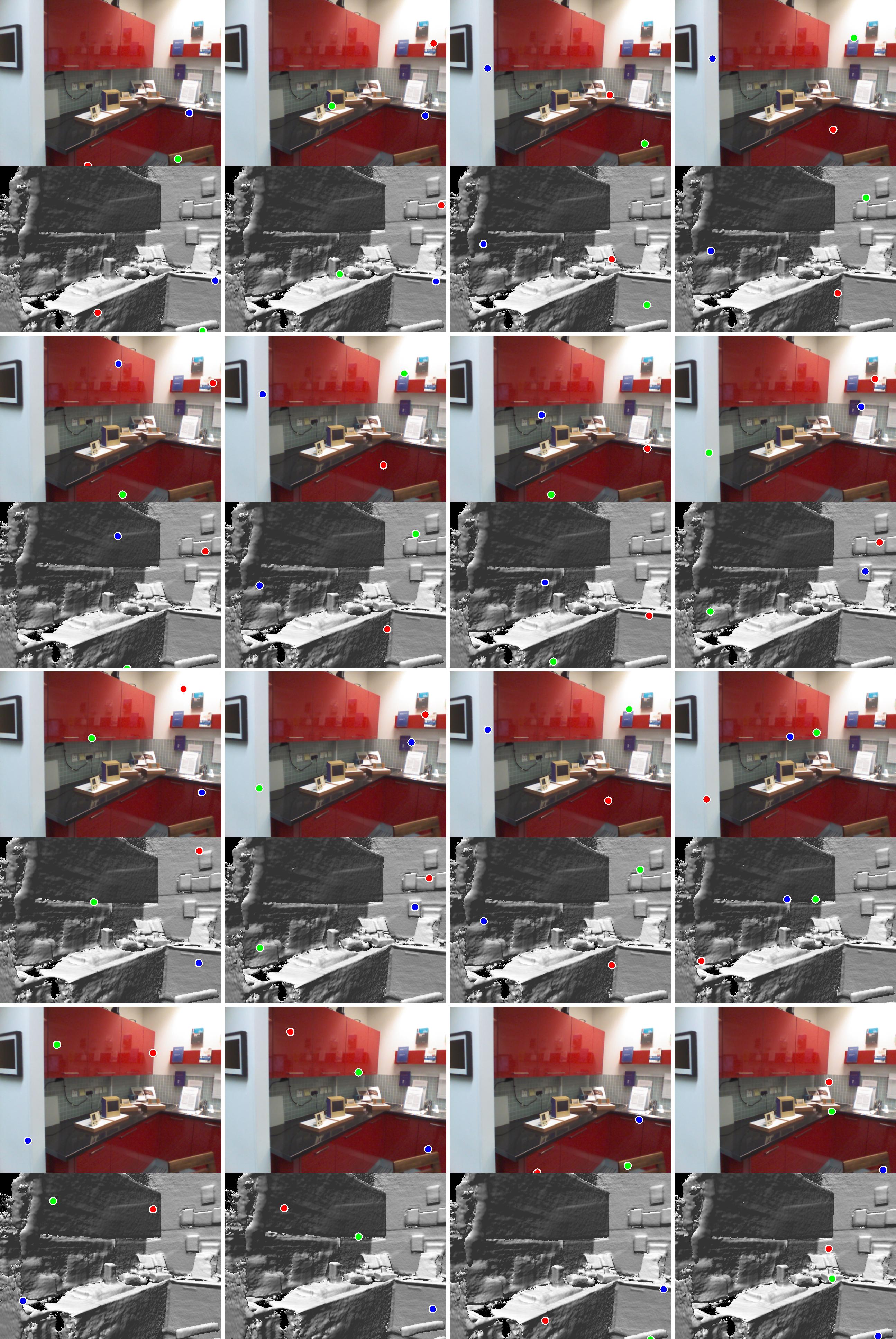}{The top $16$ pose candidates (left-to-right, top-to-bottom) corresponding to the failure case from the \emph{Red Kitchen} scene shown in Figure~\ref{fig:redkitchen71-input}. The coloured points indicate the 2D-to-3D correspondences that are used to generate the initial pose hypotheses. Note that in this case, the relocalised pose before ICP was within 5cm/5$^\circ$ of the ground truth, but that the 3D model seems somewhat eroded in comparison to the original input images. This is caused by slightly inaccurate ground truth poses in the training sequence, leading to InfiniTAM fusing an imperfect 3D model, which then affects the pose to which ICP converges.}{fig:redkitchen71-candidates}{!t}

In this sequence, we would hope that e.g.\ the brown box on the worktop would be perfectly reconstructed as InfiniTAM fuses frames from the training sequence into the 3D model, but as Figure~\ref{fig:redkitchen71-model} shows, this is not in fact the case. Initially, the box is indeed reconstructed as expected, but as the sequence proceeds, subtle inaccuracies in the ground truth poses cause parts of the box to be eroded away. When we later try to relocalise a frame from the testing sequence (see Figure~\ref{fig:redkitchen71-candidates}), the box has been significantly eroded, as has the wall at the left-hand side of the images. This is more than enough in practice to cause ICP against the corrupted model to converge to a pose that is more than $5$cm/$5^\circ$ from the ground truth, which will lead to the frame being recorded as having failed to relocalise after ICP. Moreover, since we cannot be sure that the ground truth pose for this specific test frame is not itself slightly inaccurate, an estimated pose that might have been within $5$cm/$5^\circ$ of the genuinely correct pose could in principle be marked as having failed, even though it would have succeeded if compared to a `perfect' ground truth pose.

\subsubsection{Analysis}

In practice, problems like these are almost impossible to mitigate, since they are ultimately caused by limitations of the 7-Scenes dataset itself, rather than those of our relocaliser: indeed, it could be argued that our relocaliser manages to achieve state-of-the-art results and a reasonably high degree of robustness even in the face of slightly inaccurate input data. Our results on the Stanford 4 Scenes \cite{Valentin2016} dataset, which was captured much more carefully (e.g.\ see Figure~\ref{fig:alignment}), and on the Cambridge Landmarks dataset (see \S\ref{subsec:outdoorrelocalisation}), which uses bundle-adjusted poses, also support this view.

\section*{Acknowledgements}

This work was supported by FiveAI Ltd., Innovate UK/CCAV project 103700 (StreetWise), the EPSRC, ERC grant ERC-2012-AdG 321162-HELIOS, EPSRC grant Seebibyte EP/M013774/1, EPSRC/MURI grant EP/N019474/1 and EC grant 732158 (MoveCare).

\ifCLASSOPTIONcaptionsoff
  \newpage
\fi



\bibliographystyle{IEEEtran}
\bibliography{grove}
%
%

\vspace{-2\baselineskip}

\begin{IEEEbiography}[{\vspace{-\baselineskip}\includegraphics[width=1in,height=1.25in,clip,keepaspectratio]{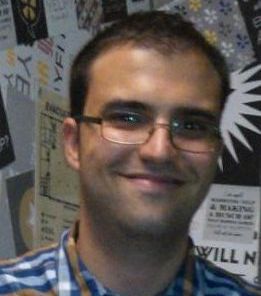}}]{Tommaso Cavallari}
received his PhD in Computer Science and Engineering from the University of Bologna in 2017. He then worked for a year as a postdoc in Prof. Torr's group at Oxford University, doing research on camera localisation and 3D reconstruction. He is now a research scientist in FiveAI's Oxford Research Group, which focuses on computer vision and machine learning for autonomous driving.
\end{IEEEbiography}

\vspace{-3.5\baselineskip}

\begin{IEEEbiography}[{\vspace{-\baselineskip}\includegraphics[width=1in,height=1.25in,clip,keepaspectratio]{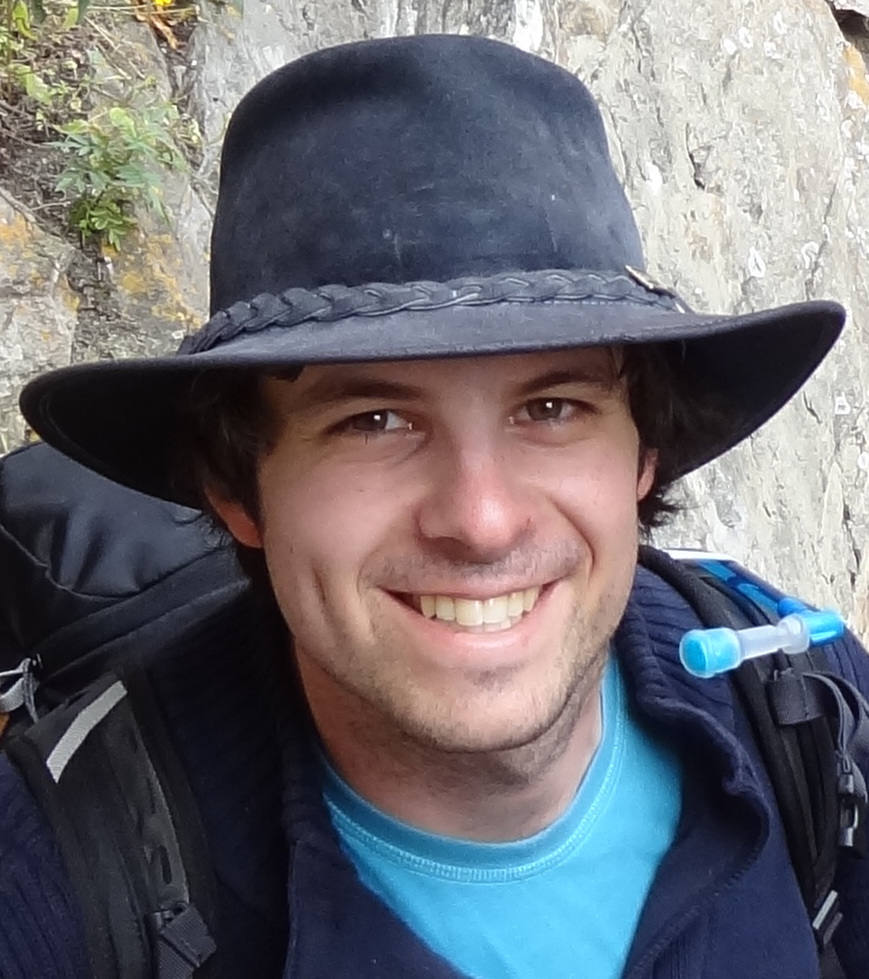}}]{Stuart Golodetz}
obtained his D.Phil. in Computer Science at the University of Oxford in 2011. After working for two years in industry, he returned to Oxford as a postdoc for four and a half years, working on 3D reconstruction, scene understanding and visual object tracking. He is currently the director of FiveAI's Oxford Research Group, which focuses on computer vision and machine learning for autonomous driving.
\end{IEEEbiography}

\vspace{-3.5\baselineskip}

\begin{IEEEbiography}[{\vspace{-\baselineskip}\includegraphics[width=1in,height=1.25in,clip,keepaspectratio]{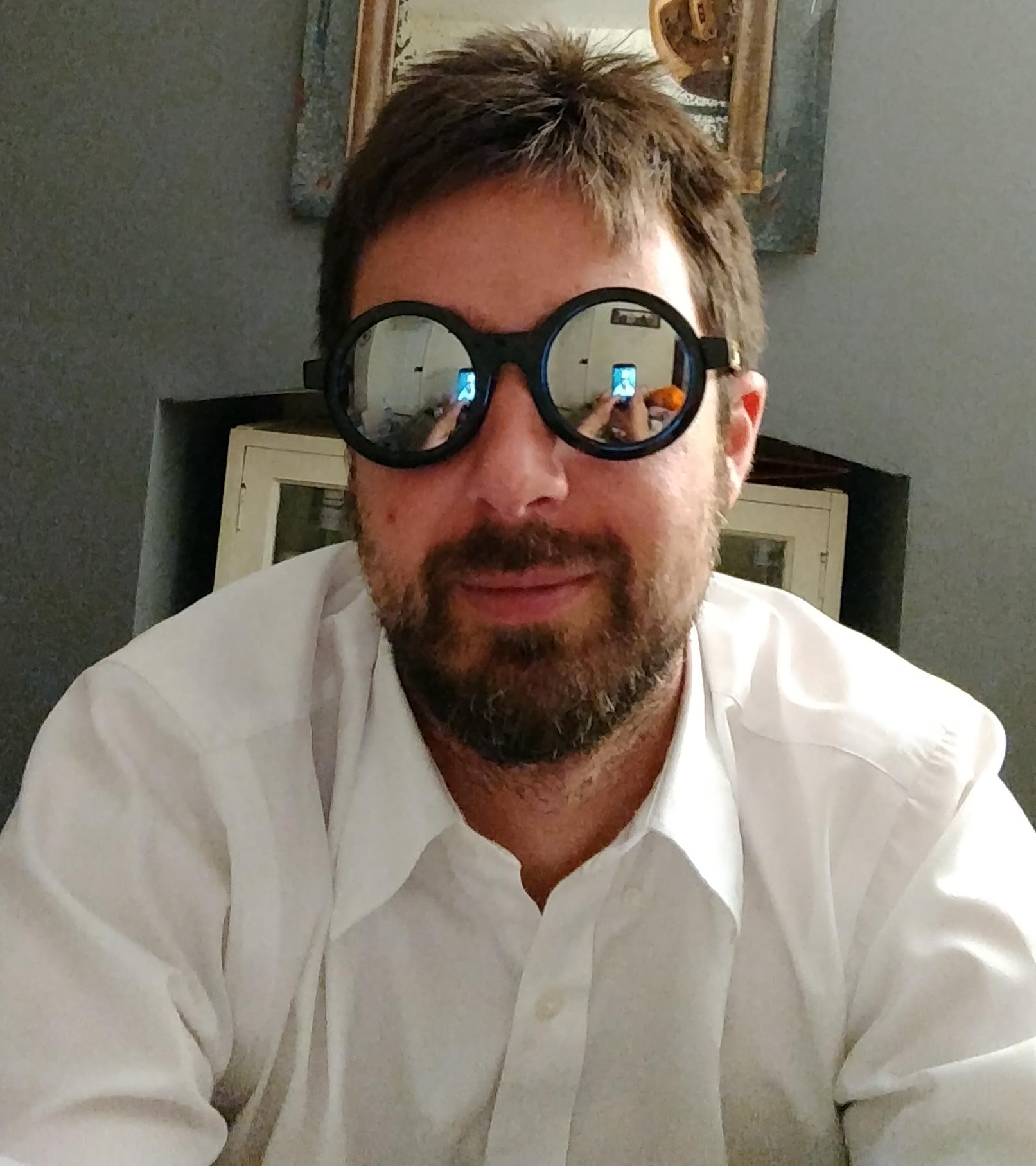}}]{Nicholas Lord}
received his Ph.D. in Computer Engineering from the University of Florida in 2007.
He spent several years in Sony's European Playstation division, including work on Wonderbook for the 2013 BAFTA Award-nominated Book of Spells.
As a postdoc at Oxford University, he worked on 6D relocalisation and analysis of the vulnerabilities of DNNs.
He is currently a research scientist at FiveAI, working on problems related to autonomous vehicles.
\end{IEEEbiography}

\vspace{-3.5\baselineskip}

\begin{IEEEbiography}[{\vspace{-2\baselineskip}\includegraphics[width=1in,height=1.25in,clip,keepaspectratio]{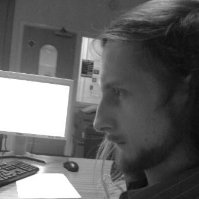}}]{Julien Valentin}
received his PhD from the University of Oxford, under the supervision of Philip H.\ S.\ Torr. He then became a founding member of PerceptiveIO, and is now working at Google on machine learning and optimization for real-time computer vision and graphics.
\end{IEEEbiography}

\vspace{-3.5\baselineskip}

\begin{IEEEbiography}[{\vspace{-2\baselineskip}\includegraphics[width=1in,height=1.25in,clip,keepaspectratio]{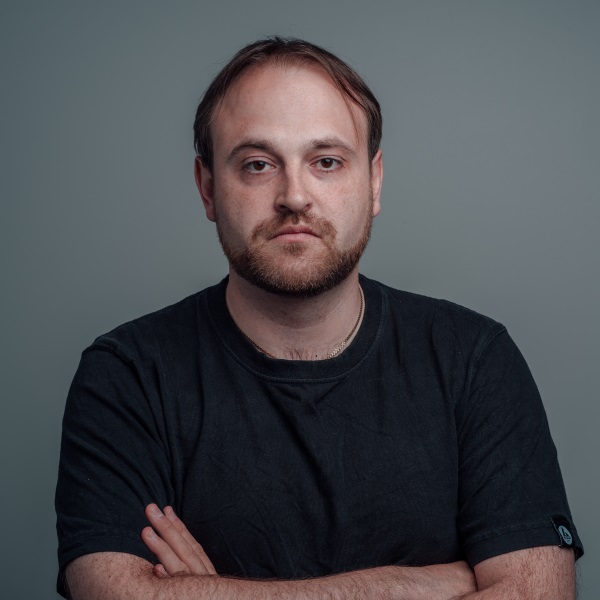}}]{Victor Prisacariu}
received the D.Phil. degree in
Engineering Science from the University of Oxford in 2012. He continued there first
as an EPSRC prize Postdoctoral Researcher and then as a Dyson Senior
Research Fellow, before being appointed an Associate Professor in 2017.
He is a Research Fellow with St Catherine's College, Oxford. His
research interests include semantic visual tracking, 3-D
reconstruction, and SLAM.
\end{IEEEbiography}

\vspace{-3.5\baselineskip}

\begin{IEEEbiography}[{\vspace{-2\baselineskip}\includegraphics[width=1in,height=1.25in,clip,keepaspectratio]{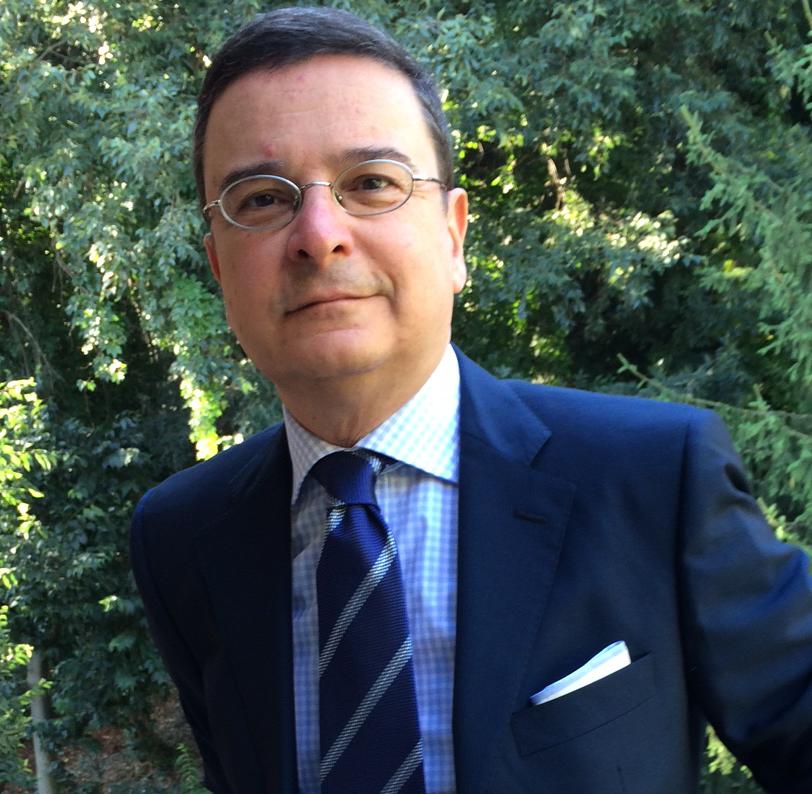}}]{Luigi Di Stefano}
received the PhD degree from the University of Bologna in 1994. He is now a full professor at the University of Bologna, where he founded and leads the Computer Vision Laboratory (CVLab).
He is the author of more than 150 papers and several patents. He has been a scientific consultant for major companies in computer vision/machine learning. He is a member of the IEEE Computer Society and the IAPR-IC.
\end{IEEEbiography}

\vspace{-3.5\baselineskip}

\begin{IEEEbiography}[{\includegraphics[width=1in,height=1.25in,clip,keepaspectratio]{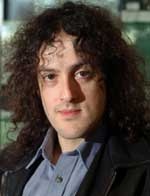}}]{Philip H.\ S.\ Torr}
received the PhD degree from Oxford University. After working for another three years at Oxford, he worked for six years for Microsoft Research, first in Redmond, then in Cambridge, founding the vision side of the Machine Learning and Perception Group. He is now a professor at Oxford University. He has won awards from top vision conferences, including ICCV, CVPR, ECCV, NIPS and BMVC. He is a senior member of the IEEE and a Royal Society Wolfson Research Merit Award holder.
\end{IEEEbiography}

\vfill

\end{document}